\documentclass{article} 
\usepackage{collas2022_conference,times}

\usepackage{amsmath,amsfonts,bm}

\def\eqref#1{equation~\ref{#1}}

\def\1{\bm{1}}

\DeclareMathAlphabet{\mathsfit}{\encodingdefault}{\sfdefault}{m}{sl}
\SetMathAlphabet{\mathsfit}{bold}{\encodingdefault}{\sfdefault}{bx}{n}

\DeclareMathOperator*{\argmax}{arg\,max}
\DeclareMathOperator*{\argmin}{arg\,min}

\usepackage{hyperref}
\hypersetup{
    colorlinks=true,
    linkcolor=red,
    filecolor=magenta,      
    urlcolor=blue,
    citecolor=purple,
    pdftitle={SANE},
    pdfpagemode=FullScreen,
    }

\usepackage{array}
\newcolumntype{H}{>{\setbox0=\hbox\bgroup}c<{\egroup}@{}}

\usepackage{xcolor}

\usepackage{graphicx}
\usepackage{subcaption}
\usepackage{float}
\usepackage[table]{colortbl} 
\usepackage{tikz}
\usepackage{booktabs}
\usepackage{algorithm, algpseudocode}

\newcommand{\tablemarginhack}{\centering\addtolength{\leftskip}{-2cm}\addtolength{\rightskip}{-2cm}}

\newcommand{\brackets}[1]{\langle #1 \rangle}

\renewcommand*\cite[1]{\citep{#1}}
\newcommand{\abs}[1]{\lvert #1 \rvert}

\title{Self-Activating Neural Ensembles for Continual Reinforcement Learning}

\author{Sam Powers \\
Carnegie Mellon University\\
\texttt{snpowers@cs.cmu.edu}
\And 
Eliot Xing \\
Georgia Institute of Technology \\
\texttt{exing@gatech.edu} \\
\And 
Abhinav Gupta \\
Carnegie Mellon University\\
\texttt{gabhinav@cs.cmu.edu}
}

\newcommand{\github}{\url{https://github.com/AGI-Labs/continual_rl}
}

\collasfinalcopy 

\begin{document}

\maketitle

\begin{abstract}
The ability for an agent to continuously learn new skills without catastrophically forgetting existing knowledge is of critical importance for the development of generally intelligent agents. Most methods devised to address this problem depend heavily on well-defined task boundaries, and thus depend on human supervision. Our task-agnostic method, Self-Activating Neural Ensembles (SANE), uses a modular architecture designed to avoid catastrophic forgetting without making any such assumptions. At the beginning of each trajectory, a module in the SANE ensemble is activated to determine the agent’s next policy. During training, new modules are created as needed and only activated modules are updated to ensure that unused modules remain unchanged. This system enables our method to retain and leverage old skills, while growing and learning new ones. We demonstrate our approach on visually rich procedurally generated environments. 


\end{abstract}

\section{Introduction}

Lifelong learning~\cite{thrun1995lifelong} is of critical importance for the field of robotics. An agent that interacts with the world should continuously learn from it and act intelligently in a wide variety of situations. In contrast to this ideal, most standard deep reinforcement learning methods are centered around a single task. First, a task is defined, then a policy is learned to maximize the rewards the agent receives in that setting. If the task is changed, a new model is learned from scratch, discarding the previous model and previous interactions. Task specification thus plays a central role in current end-to-end deep reinforcement learning frameworks. 

In contrast, humans do not require concrete task boundaries to be able to effectively learn separate tasks---instead, we perform continual (lifelong) learning. We learn new skills efficiently by leveraging prior knowledge, without forgetting old behaviors. However, when placed into continual learning settings, current deep reinforcement learning approaches do neither: the forward transfer properties of these systems are negligible, and they suffer from catastrophic forgetting~\cite{mccloskey1989catastrophic, french1999catastrophic}. 

The core issue of catastrophic forgetting is that a neural network trained on one task starts to forget what it knows when trained on a second task, and this issue only becomes exacerbated as more tasks are added. The problem ultimately stems from sequentially training a single network in an end-to-end manner. The shared nature of the weights and the use of backpropagation to update them mean that later tasks overwrite earlier ones~\cite{mccloskey1989catastrophic, ratcliff1990connectionist}. 

To handle this, past approaches have proposed a wide variety of ideas: from task-based regularization~\cite{kirkpatrick2017ewc}, to learning different sub-modules for different tasks~\cite{rusu2016progressive}, and dual-system slow/fast learners inspired by the human hippocampus~\cite{schwarz2018progress}. The fundamental problem of continual learning, which few methods address, is that the agent should autonomously determine how and when to adapt to changing environments, or stabilize existing knowledge, without explicit task specification. It is infeasible for a human to indefinitely provide agents with task-boundary supervision, and doing so side-steps the core problem.  

There are a few existing task-agnostic~\cite{zeno2018task} methods, though most have only been demonstrated on classification or behavior cloning: for example~\citet{aljundi2019task} addresses the problem by detecting plateaus and using those as boundaries,~\citet{lee2019neural} adaptively creates new clusters using Dirichlet processes, and~\citet{veness2021gated} replaces backpropagation completely. Methods that have been demonstrated on reinforcement learning are rarer; exceptions include~\citet{rolnick2019experience}, which utilizes a large replay buffer, and~\citet{lomonaco2020continual} which uses the error in the value estimate to determine when to consolidate modules.

\begin{figure}[h]
    \centering
    \includegraphics[trim=1em 1em 1em 1em, clip, width=0.35 \columnwidth]{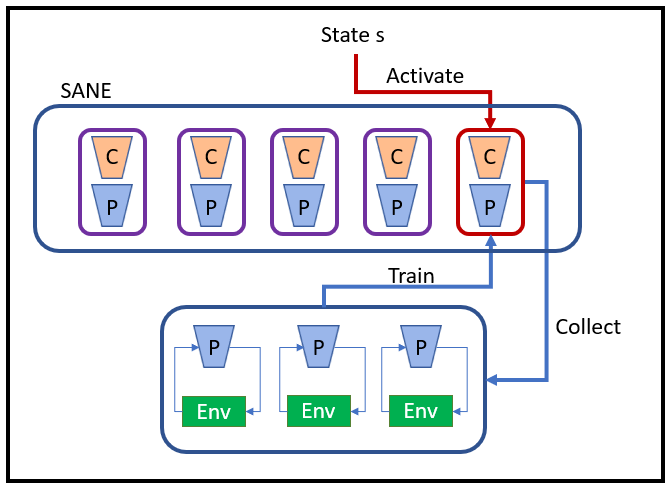}
    \caption{The overall structure of the SANE system. Each module contains an actor and a critic. Upon activation, collection occurs from several environments in parallel.}
    \label{fig:overall_structure}
\end{figure}

We approach the problem by introducing a system that continuously, dynamically adapts to changing environments. Our ensemble-based method, Self-Activating Neural Ensembles\footnote{Code available: \github} (SANE), depicted in Figure \ref{fig:overall_structure}, is the core of our proposal. Each module in the ensemble is a separate, task-agnostic network. Periodically, a single module from the ensemble is activated to determine which policy to use. Only activated modules are updated, leaving unused modules unchanged and therefore protected from catastrophic forgetting. Crucially, our ensemble is dynamic: new modules are created when existing modules are found to be insufficient. In this way, modules are created when novel scenarios are encountered, preventing destructive updates to other modules. Additionally, SANE is simple; modules control their own relevance, activating when the situation to which they are specialized is encountered, and remaining untouched the rest of the time. SANE provides the following desirable properties for continual reinforcement learning: (a) It mitigates catastrophic forgetting by only updating relevant modules; (b) Because of its task-agnostic nature, unlike previous approaches, it does not require explicit supervision with task IDs; (c) It achieves these targets with bounded resources and computation. We demonstrate SANE on three visually rich, challenging level sequences based on Procgen~\cite{cobbe2020leveraging} environments. Additionally, we analyze the behavior of SANE at a more fine-grained level on 2 individual runs, to gain more understanding of the dynamics of training SANE.

\section{Related Work}
\label{section:related_works}

\textbf{Continual learning} Any continual learning system must balance \textit{stability} (the extent to which existing knowledge is retained) and \textit{plasticity} (how readily new knowledge is acquired)~\cite{grossberg1982does, abraham2005memory, mermillod2013stability}. Stability has posed a substantial challenge due to \textit{catastrophic forgetting}, by which neural networks trained by backpropagation abruptly forget learned behavior for solving old tasks when presented with new tasks~\citep{kemker2018measuring, mccloskey1989catastrophic, ratcliff1990connectionist, lewandowsky1995catastrophic, french1999catastrophic}. Broadly, methods for continual learning can be categorized under Regularization, Rehearsal, or Architectural approaches, as well as combinations of them. We refer the reader to the survey papers by~\citet{parisi2019continual, lesort2020continual, mundt2020wholistic} for general discussion. Here we review methods for continual learning relevant to our approach.

Recent strategies for mitigating catastrophic forgetting such as Elastic Weight Consolidation (EWC)~\cite{kirkpatrick2017ewc}, among other Regularization approaches~\cite{lee2017overcoming, li2017learning, zenke2017continual, ritter2018online, chaudhry2018riemannian, jaeger2017using, he2018overcoming, serra2018overcoming, aljundi2018memory, aljundi2019task, park2019continual}, constrain updates to network parameters important for past tasks when learning new tasks. 
However, these methods fundamentally run into the stability-plasticity dilemma, as over-constraining updates can hinder the learning of new tasks. To improve plasticity, dynamic architectures~\cite{ring1998child, zhou2012online, terekhov2015knowledge, rusu2016progressive} incorporate additional network parameters to help learn new tasks. Furthermore, to prevent model size from growing unbounded, such approaches~\cite{xiao2014error, cortes2017adanet, yoon2018lifelong, mallya2018piggyback, mallya2018packnet, xu2018reinforced, schwarz2018progress, kaplanis2019policy, traore2019continual} use distillation~\cite{bucilua2006model, hinton2015distilling, rusu2015policy, teh2017distral}, pruning, and related techniques to consolidating learned behavior while reducing parameter count. Similarly, Rehearsal and (generative) memory-based approaches~\cite{robins1995catastrophic, french1997pseudo, gepperth2016geppnet, furlanello2016active, rebuffi2017icarl, shin2017continual, draelos2017neurogenesis, kamra2017deep, lopez2017gradient, chaudhry2018efficient, wu2018memory, isele2018selective, parisi2018lifelong, riemer2018learning, soltoggio2018born, kemker2018fearnet, van2018generative, xiang2019incremental, aljundi2019online, lesort2019generative, caccia2020continual_compression, caselles2021s} must also balance data storage and memory network constraints when determining which examples are needed to preserve previously learned behavior. We build our ensemble approach off of CLEAR~\cite{rolnick2019experience}, a state-of-the-art asynchronous continual RL method which uses Rehearsal, by maintaining a replay buffer that uniformly preserves past experience via reservoir sampling~\cite{isele2018selective}, along with Regularization, via behavioral cloning and a KL penalty to preserve prior learned behavior. 


\textbf{Ensemble methods} Falling under Architectural approaches, aggregation ensembles~\cite{cheung2019superposition, wen2020batchensemble, veness2021gated} combine predictions from multiple models to produce a final output. These types of ensembles are also commonly used for uncertainty estimation~\cite{lakshminarayanan2017simple}, exploration~\cite{pathak2019self}, or reducing overestimation bias such as in double Q-learning~\cite{van2016deep, fujimoto2018td3}. In contrast, modular ensembles~\cite{aljundi2017expertgate, fernando2017pathnet, lee2019neural, parascandolo2018learning, kessler2021same} use a subset of the entire ensemble's parameters to select an appropriate expert model for the task presented. Selectively updating a subset of parameters or specific modules instead of the entire ensemble can circumvent catastrophic forgetting while bounding compute costs; this is a feature we utilize in SANE, which is a type of modular ensemble rather than the former, aggregation ensemble. Our method is similar to Multiple Choice Learning~\cite{guzman2012multiple, lee2016stochastic, lee2017confident, seo2020trajectory}, which chooses and updates only the best expert from an ensemble, encouraging specialization. However, Multiple Choice Learning uses fixed-size static ensembles, while SANE is a dynamic ensemble that merges similar modules and works with a given resource budget. For supervised continual learning, LMC~\cite{ostapenko2021continual} also proposes a modular ensemble approach, although LMC assumes access to task IDs at training time and can only add modules, meaning that its computational footprint is linear relative to the number of tasks learned. In contrast, SANE is completely task-agnostic at train and test time, while also creating and merging modules to meet a given compute budget.




Hierarchical RL can be seen as a hierarchy of meta-policies that control access to an ensemble of (often hand-designed) sub-policies that act at differing temporal resolutions~\cite{sutton1999_options, brunskill2014_pac, tessler2017_deep_hierarchical, zhang2021_intrinsicoptions}. Analogous to our own value-based activation score, some hierarchical RL methods use predicted Q-values to select amongst their ensemble, as in~\cite{dayan1992_feudalrl, dietterich1999_maxq}.~\citet{goyal2019_competitive_ensembles} demonstrates the utility of avoiding meta-policies, instead relying on primitives that independently determine their own relevance, similar to self-activation in our approach. However, their primitives distinguish themselves by factorizing a state space, placing strong assumptions on the learnable policies. Additionally, their primitives are not created over time, so the method relies on regularization to ensure primitives in their ensemble are used.

\section{Background}
We review background on the continual RL setting we study in Appendix~\ref{appendix:extended_background}. Traditional neural networks suffer from catastrophic forgetting because weights in the network are changed by backpropagation every update~\cite{mccloskey1989catastrophic}, causing information learned in a new scenario to overwrite prior behavior. 
Instead of learning and updating a single neural network for policy $\pi$ across multiple tasks, we propose using an dynamic ensemble of \textit{self-activating modules}. Our approach partitions, allocates, and manages parameters for separate modules, so that each module may handle different situations without interfering with others.

If a module is relevant to the current situation, it activates during inference and is updated during training. If a module is irrelevant, it is unused and remains unchanged. One way of viewing these modules is as latent behaviors, each specialized to a particular circumstance. For example, if in one context an agent must carefully wait to allow an enemy to pass, we don't want this to disrupt a behavior where moving quickly to dodge an enemy is the best action. 

How may we know when to use which module, when task boundaries are ambiguous and not given by human supervision? Each module in our ensemble predicts an \textit{activation score}, which estimates the relevancy of a given module's behavior to the current situation, and the module with the highest activation score is selected from the ensemble. An appropriate activation score will protect modules against catastrophic forgetting, and can also enable forward transfer, by activating modules with prior learned behaviors that are advantageous in new settings.


How should such an ensemble be structured? Pre-defining a static fixed-size ensemble is ineffective for module-based behavior specialization. In such a static ensemble, one module will tend to perform well at a task, leading to that module being chosen as the starting point for future tasks which results in catastrophic forgetting. Regularizing with additional losses would be necessary to distribute activation across the ensemble's experts, as in~\cite{jacobs1991adaptive, shazeer_2018_gated_mixture_of_experts, goyal2019_competitive_ensembles}. Instead, we design SANE as a \textit{dynamic ensemble}, in which modules are created and merged together as necessary. Intuitively, modules are created when existing latent behaviors fail to perform as expected, and the ensemble determines that a new latent behavior is needed. Modules may also be merged to conserve resource consumption and meet a given compute budget. 


Bringing self-activating modules and a dynamic ensemble together, we present Self-Activating Neural Ensembles (SANE). To summarize, our approach differs from traditionally-used ensembles in two ways:
(i) We do not aggregate results across modules, in order to keep modules isolated from one another. This circumvents catastrophic forgetting, by not backpropogating through the entire ensemble.
(ii). The ensemble itself is dynamic, in that modules are being created and merged throughout training.


\section{Self-Activating Neural Ensembles for Continual RL}
We now proceed to formally describe SANE in full detail. SANE is a dynamic collection of modules $\{M_1, \ldots, M_k\}$ where, based on the context, one module $M_t$ activates and is used for inference. Subsequently, given transitions from collected episodes, only the selected module $\mathcal{M}_t$ is updated. We describe an individual SANE module in Section~\ref{section:primitive}, including how activation scores are computed to determine which module to use. 
We present the learning process to manage a dynamic ensemble in Section~\ref{section:ens_structure_updates}. Pseudocode is provided in Appendix \ref{sec:pseudocode}.

\subsection{Self-Activating Module}
\label{section:primitive}


Every module $\mathcal{M}_i$ is an actor-critic algorithm represented by: a policy $\pi_i(a | s)$, a critic $V_i(v, u | s)$, as well as a replay buffer $\mathcal{B}_i$ that holds experience transitions. We modify the critic $V_i$ from the standard formulation in the following way. Given a state $s$ at timestep $t$, the critic $V_i$ predicts two scalars: $v_i(s)$, the value estimate of the return $R_t$ received if module $\mathcal{M}_i$ is activated, and $u_i(s)$, an uncertainty estimate of the absolute error:
\begin{equation}
\label{eqn:uncertainty}
u_i(s) \approx \abs{R_t - v_i(s)}
\end{equation}

We proceed by defining an optimistic estimate $v^{UCB}_i$ (upper confidence bound) and a pessimistic estimate $v^{LCB}_i$ (lower confidence bound) for the return that the module $\mathcal{M_i}$ can achieve from state $s$:
\begin{equation}
\label{eqn:ucb}
v^{UCB}_{i}(s) = v_i(s)  + \alpha_u * u_i(s) 
\end{equation}
\begin{equation}
\label{eqn:lcb}
v^{LCB}_i(s) = v_i(s)  - \alpha_l * u_i(s)
\end{equation}

where $\alpha_u, \alpha_l>0$ are hyperparameters which represent how wide a margin around the expected value to allow. We use these margins to: (i) choose which module to activate during inference; (ii) decide when to create a new module during Structure Update (Section~\ref{section:ens_structure_updates}).

In all, each module $\mathcal{M}_i$ can be considered to be a tuple $\brackets{\pi_i, V_i, \mathcal{B}_i, \overline{V}_i, A_i}$, where $\overline{V}_i$ and $A_i$ are two other versions of the critic $V_i$, which we proceed to describe.

The target network $\overline{V}_i$ is used for the confidence bounds estimates (Equation~\ref{eqn:ucb} and~\ref{eqn:lcb}) instead of $V_i$. Target networks are commonly used in Q-learning~\cite{mnih2015human, lillicrap2016continuous, anschel2017averaged, fujimoto2018td3} to stabilize training by reducing variance from approximation error. Similarly, we update $\overline{V}_i$ with an exponential moving average~\cite{ruppert1988efficient, polyak1992acceleration, izmailov2018averaging}. We denote $V_i$'s parameters by $\theta_i$, $\overline{V}_i$'s parameters by $\theta_i'$, and the update rate by $\tau_V$; we use the update: $\theta_i' \leftarrow \tau_V \theta_i + (1-\tau_V) \theta_i'$. 

The anchor $A_i$ is a frozen instance of the critic $V_i$ from when the module $\mathcal{M}_i$ was created. We describe how we use the anchor $A_i$ to measure drift in Section~\ref{sec:measuring_drift}.

\textbf{Module update} SANE can be applied to any actor-critic algorithm; we describe the specifics of our implementation in Section 4.4. Let $L_{\mathcal{M}_i}$ denote the loss function of the active SANE module and $L_{rl}$ be the loss of the actor-critic RL algorithm, with components associated with module $\mathcal{M}_i$. We perform a module update by optimizing $L_{\mathcal{M}_i}=L_{rl} + \mu L_{ue}$, where $L_{ue}$ is MSE loss to estimate uncertainty from Equation~\ref{eqn:uncertainty}.


\textbf{Inference (Self-Activation)} SANE consists of several modules where each module represents the behavior for a particular situation. Activating the right module for the right situation is key to the success of the SANE method. In an RL setting, the critic predicts a value estimate, which can serve as an effective proxy for how successful a module $\mathcal{M}_i$ may be in obtaining high return. At the beginning of the episode, we compute $v^{UCB}_i$ for each module in the ensemble $\{\mathcal{M}_1, \ldots, \mathcal{M}_k\}$ using the target network critic $\overline{V}_i$. Then, we greedily select the module whose critic predicts the highest such value, and use that module for the whole episode.

\subsection{Dynamic Ensemble}
\label{section:ens_structure_updates}

We propose a process to dynamically update the structure of the ensemble in SANE. If the current set of modules behave in an expected manner (returns are within the expected range) then the current set is sufficient. However at some point in training, if the returns are outside the expected range, then we know the current set of modules is insufficient. We create new modules to handle the new situation, and merge modules together to stay within a given compute budget. 


\subsubsection{Measuring drift}
\label{sec:measuring_drift}

The key to successfully updating the SANE structure lies in our ability to detect that we have moved outside this expected range. Our main assumption here is that the change in rewards received is sufficient for distinguishing relevant changes in setting. Therefore, we detect change in setting by measuring drift in rewards. Drift describes when an environment is non-stationary, e.g. when the reward distribution or the state transition distribution is changing over time. Often where drift occurs, catastrophic forgetting follows because networks update to the new setting, forgetting the old. 

To recognize drift with SANE, at the time of their creation modules have their critic cloned and frozen, creating a static critic called the \emph{anchor}. We compare the prediction of a module's critic to the prediction of its anchor. We say that sufficient change has occurred when the bounds of the expected return, as defined in Section \ref{section:primitive}, predicted by a module's critic do not include the value predicted by its anchor, which serves as a static baseline. 

Let $v_{A_i}$ denote the value estimate of the anchor $A_i$. Formally, we say that sufficient change has occurred when for a given state $s$, critic $V_i$, and anchor $A_i$, either of the following inequalities hold:
\begin{align}
    v^{UCB}_i(s) &< v_{A_i}(s) \\
    v^{LCB}_i(s) &> v_{A_i}(s)
\end{align}

In practice, we use the target network critic $\overline{V}_i$ to predict $v^{UCB}_i(s)$ and $v^{LCB}_i(s)$, instead of $V_i$. 

\subsubsection{Creating a new module}

When drift occurs such that the returns are better than anticipated, we expect that this corresponds to the case that the policy has simply improved in the current setting, as intended by standard module policy training. In this setup, we just update the anchor to improve expectation. However, the case of negative drift, where the UCB falls below the estimate of the anchor, requires a different strategy. This situation occurs when the behavior (policy) starts under-performing expectation, which can occur when the task has been changed and the old policy is no longer as effective as it had been. What we do in this case is create a new module that is a clone of the one that was activated. We empty the replay buffer and update the anchor at the time of creation of the new module. The goal is for the new module to be activated by the new setting while the old one continues to be activated by the old setting, splitting the input space to more effectively handle the two desired behaviors that are not well handled by a single policy.

\subsubsection{Merging modules}

To prevent unlimited memory consumption, we limit the the total number of modules in our ensemble by merging modules. To execute a merge, we start by finding the two modules in the ensemble that are closest by the L2 distance between frames averaged from a sample of trajectories from the replay buffers. We then keep the more frequently used module and drop the less frequent module from the ensemble. Before dropping, we combine the replay buffer of the two modules and run a module update (Section \ref{section:primitive}) on the combined module. 

Note that in combining the replay buffers of the two modules we use the reservoir sampling technique from CLEAR~\cite{rolnick2019experience}. We maintain a reservoir value for each trajectory, defined as a random value between 0 and 1, that allows every trajectory to have an equal chance of being stored in the buffer, regardless of when it was collected. The trajectories from the module being dropped are added to the replay buffer of the module being kept using the reservoir values that were originally generated.

\subsection{Implementation Details}
\label{section:implementation}

\noindent {\bf Leveraging CLEAR and IMPALA} We have chosen to base our modules on IMPALA-based CLEAR as implemented by CORA~\cite{powers2021cora}, as it allows us to get several useful features for free: a. Learning is done efficiently, in a highly parallel manner. b. The expected return, used for training both the critic and the policy (for SANE as well as all baselines), is computed using vtrace, an effective credit assignment method, as described in~\cite{espeholt2018impala}. c. The CLEAR replay buffers are maintained using reservoir sampling. d. CLEAR provides auxiliary losses that maintain consistency of both the policy and critic with the replay buffer.


\noindent {\bf Model architecture} The base implementation uses the Nature CNN model from~\citet{mnih2015human}. We augment the baseline network with 2 hidden linear layers of dimension 32 with ReLU nonlinearities to increase its representational capacity. All other hyperparameters for the experiments are provided in Appendix \ref{section:hyperparams}. Our code is provided in additional materials, and will be open sourced upon publication. 

\noindent {\bf Parallelism} By collecting data from multiple environments in parallel, training is considerably faster, but it requires us to make one key assumption: the activated module must be guaranteed to be applicable to all actors being run at the same time. This requires that all actors be running the same task. 

\section{Experiments}
\label{section:experiments}

\begin{figure}[H]
    \centering
    \captionsetup[subfigure]{justification=centering}
    \begin{subfigure}{0.33\textwidth}
        \includegraphics[width=0.18 \textwidth]{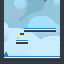}
        \includegraphics[width=0.18 \textwidth]{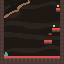}
        \includegraphics[width=0.18 \textwidth]{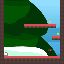}
        \includegraphics[width=0.18 \textwidth]{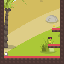}
        \caption{Climber Levels}
    \end{subfigure}
    \begin{subfigure}{0.33\textwidth}
        \includegraphics[width=0.18 \textwidth]{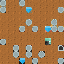}
        \includegraphics[width=0.18 \textwidth]{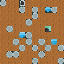}
        \includegraphics[width=0.18 \textwidth]{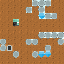}
        \includegraphics[width=0.18 \textwidth]{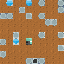}
        \caption{Miner Levels}
    \end{subfigure}
    \begin{subfigure}{0.33\textwidth}
        \includegraphics[width=0.18 \textwidth]{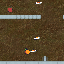}
        \includegraphics[width=0.18 \textwidth]{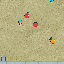}
        \includegraphics[width=0.18 \textwidth]{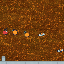}
        \includegraphics[width=0.18 \textwidth]{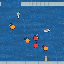}
        \includegraphics[width=0.18 \textwidth]{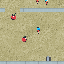}
        \caption{Fruitbot Levels}
    \end{subfigure}
    \caption{The first frame of each sequence of levels used in our experiments.}
    \label{fig:climber_levels}
\end{figure}

\textbf{Task sequences} We choose three procedurally-generated game environments (Climber, Miner, Fruitbot) from Procgen~\cite{cobbe2020leveraging}. We construct three task sequences using each of these game environments, by isolating sequences of levels that are likely to cause catastrophic forgetting and where approaches like CLEAR would perform poorly. We selected four levels for Climber and Miner. For Fruitbot, we added an easier fifth level at the start as a simple curriculum. We run each set of levels for three cycles, to see how learning evolves as the levels are seen again. The first frame of the selected levels are visualized in Figure~\ref{fig:climber_levels}.

\noindent {\bf Baselines} We compare our approach to three baselines. We also perform an ablation showing the importance of the dynamic ensemble compared to a static set. The baselines we selected are:

\begin{itemize}

\item {\bf CLEAR}
 We compare to CLEAR \citep{rolnick2019experience}, a state-of-the-art continual RL method~\citep{powers2021cora}.
 In addition to comparing to CLEAR with the default number of parameters (the same as each module in the SANE ensemble), we also compare to a version of CLEAR with as many total parameters as we use in our SANE ensemble. We refer to this as ``CLEAR 8x''. 
 
 Note that while SANE takes around 14 hours to run our Fruitbot sequence and standard CLEAR takes around 10 hours, these larger models take longer to run: CLEAR 8x took 4.5 days. We would have liked to compare to a CLEAR 32x as well, but such an experiment was on track to take more than 2 weeks. This exemplifies another benefit of SANE: the effective usage of more parameters without such a dramatic increase in runtime.
 
 \item {\bf Elastic Weight Consolidation (EWC)} We compare to EWC \citep{kirkpatrick2017ewc}, which uses the diagonal of the Fisher matrix to estimate the importance of parameters for past tasks, and slows updates to those parameters when learning new tasks. 
 
 \item {\bf Progress \& Compress (P\&C)}
 We additionally compare to P\&C \citep{schwarz2018progress}, which uses an online variant of EWC to consolidate learned behavior between dual networks, after each task is learned.
 
 \item {\bf Static SANE Ensemble} To validate the utility of our dynamic SANE ensemble, we compare to a SANE ensemble that is static: all modules are initialized upfront, and no creation or merging occur.
 
 \item {\bf SANE Oracle} We also compare against an Oracle version of SANE, where each task has its own pre-specified module, which is looked up by task ID.
 
 \end{itemize}
 
 \noindent {\bf Experimental setup \& metrics} All hyperparameters for the methods used are given in the Appendix~\ref{section:hyperparams}. For fairness of comparison we hold constant the number of replay frames each method has access to in total, at 400k frames. 

All implementations for baselines are based on those provided by~\cite{powers2021cora}. We use Continual Evaluation to generate plots for each task in the task sequence, which show how well each task was learned and how well each task was remembered. Every method was run for 5 seeds, and the mean and standard error of the mean are shown in the graphs. Gray shaded rectangles show when the agent trains on each task. We also report the Forgetting metric introduced in~\cite{powers2021cora}. We reproduce the definition of their Forgetting metric here.
\begin{align}
r_{i, j, end} & \text{\hspace{1em}expected return achieved on task $i$ after training on task $j$} \\
r_{i, all, max} & \text{\hspace{1em}maximum expected return achieved on task $i$ after training on all tasks}
\end{align}
Forgetting compares the maximum final expected return achieved for a task $i$ at any prior point to the expected return while training on task $j$, where $j > i$:
$$\mathcal{F}_{i, j} = \frac{10}{s} \sum_s \left( \frac{r_{i, j, end} - r_{i, j-1, end}}{|r_{i, all, max}|}  \right)$$

We compute the Forgetting statistic for only the first cycle for each seed and take the average across tasks. We report the average and standard error of the mean across seeds for the Forgetting summary statistic.

\subsection{Results}
 We present the Forgetting summary statistics \citep{powers2021cora} for all methods in Table \ref{table:forgetting_stats} and the Continual Evaluation graphs, which present the average and standard error of the mean of the returns received from the environment versus steps taken in the environment, in each section. We also present the final average performance and standard error of the mean for all benchmarks in the Appendix, Tables \ref{tab:climber_final}-\ref{tab:fruitbot_final}.
 
 \noindent{\bf Climber:} First we demonstrate SANE on Climber, a side-view task where the agent must ascend a series of platforms while collecting coins and dodging bats. The selected levels are particularly challenging because avoiding the bats requires relatively precise timing; a slight decay in policy performance results in significantly reduced reward. 
 
\begin{figure}[H]
    \centering
    \includegraphics[trim=0 0em 22em 0, clip,width=0.18 \columnwidth]{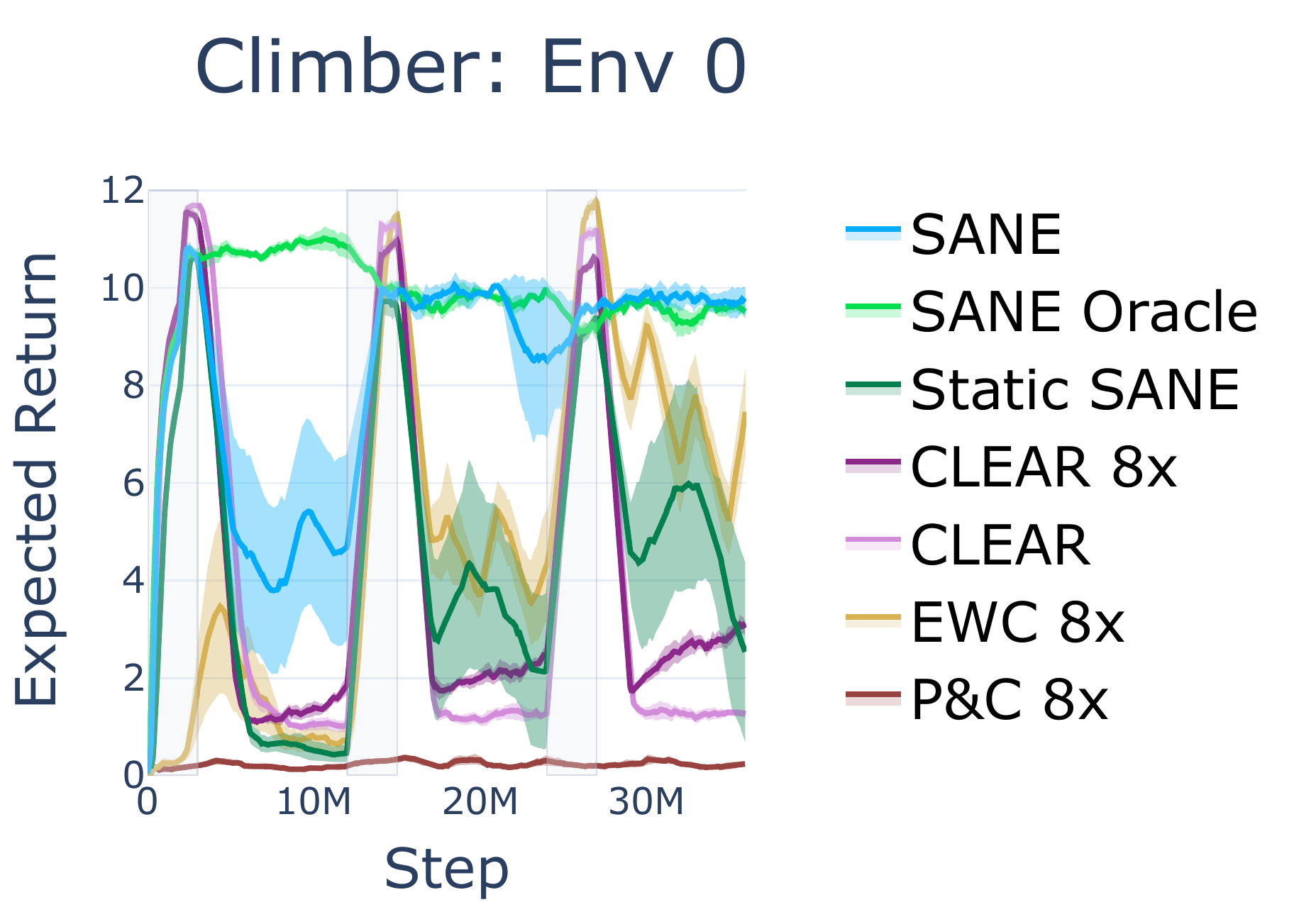}
    \includegraphics[trim=0 0em 22em 0, clip,width=0.18 \columnwidth]{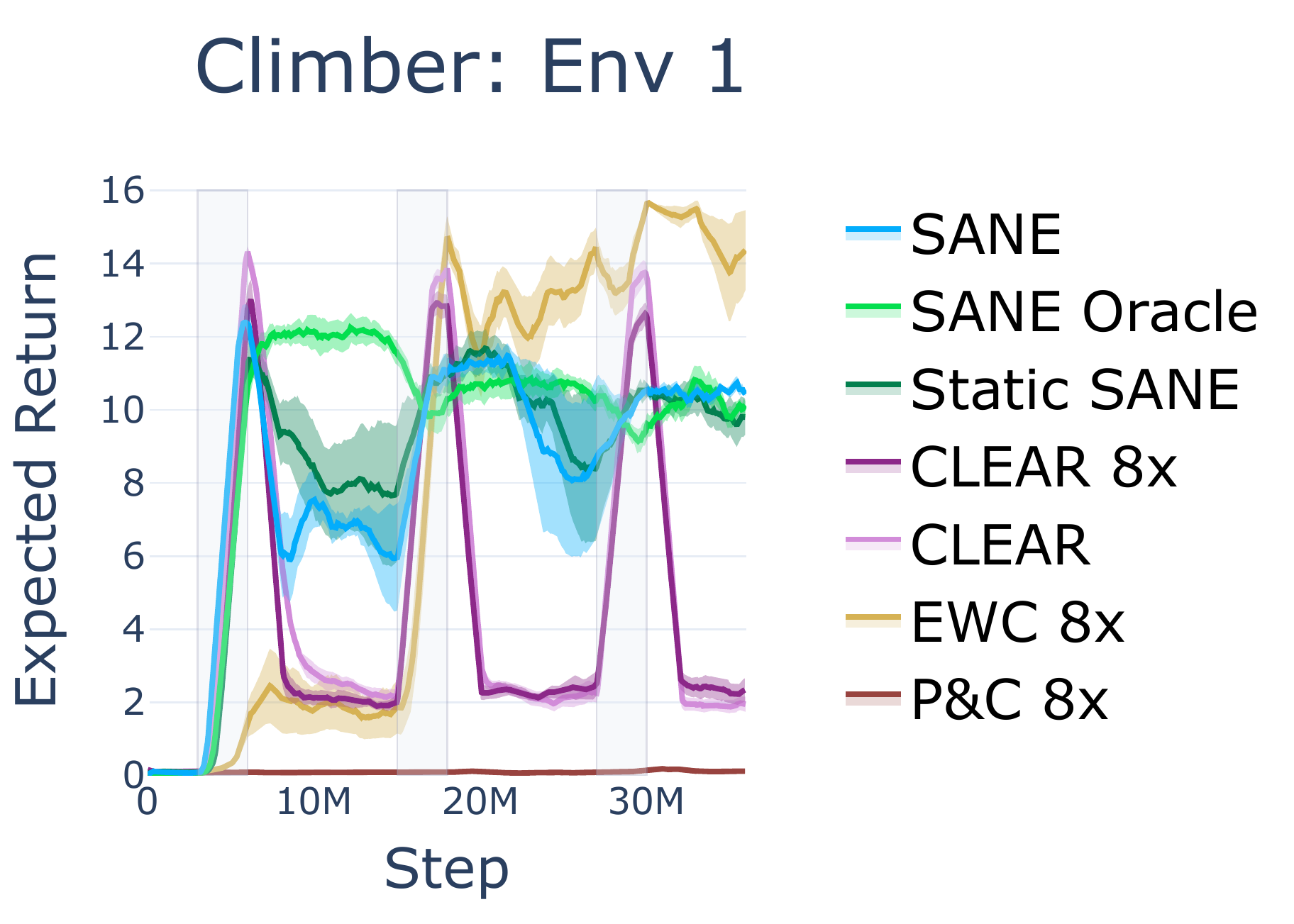}
    \includegraphics[trim=0 0em 22em 0, clip,width=0.18 \columnwidth]{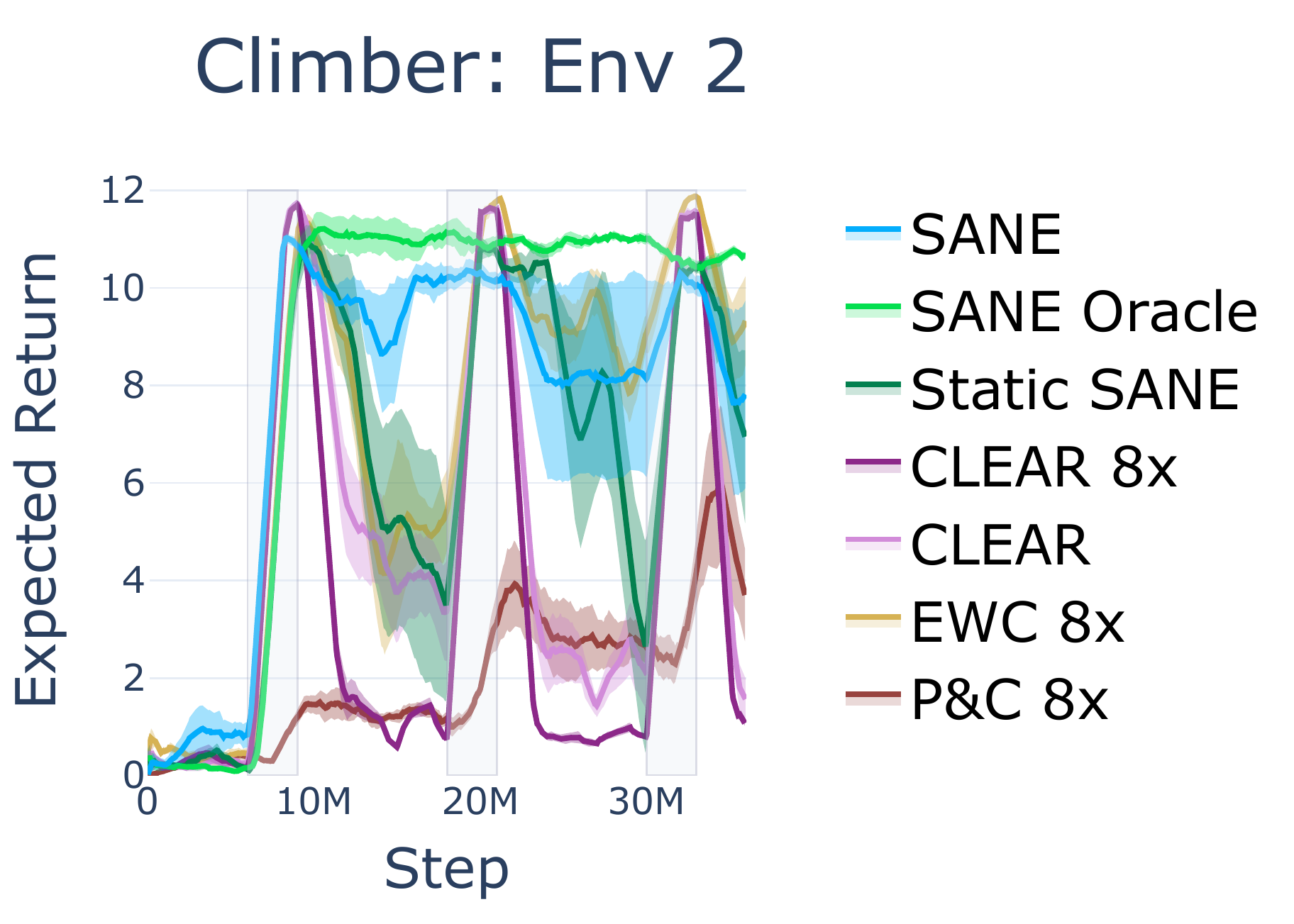}
    \includegraphics[width=0.31 \columnwidth]{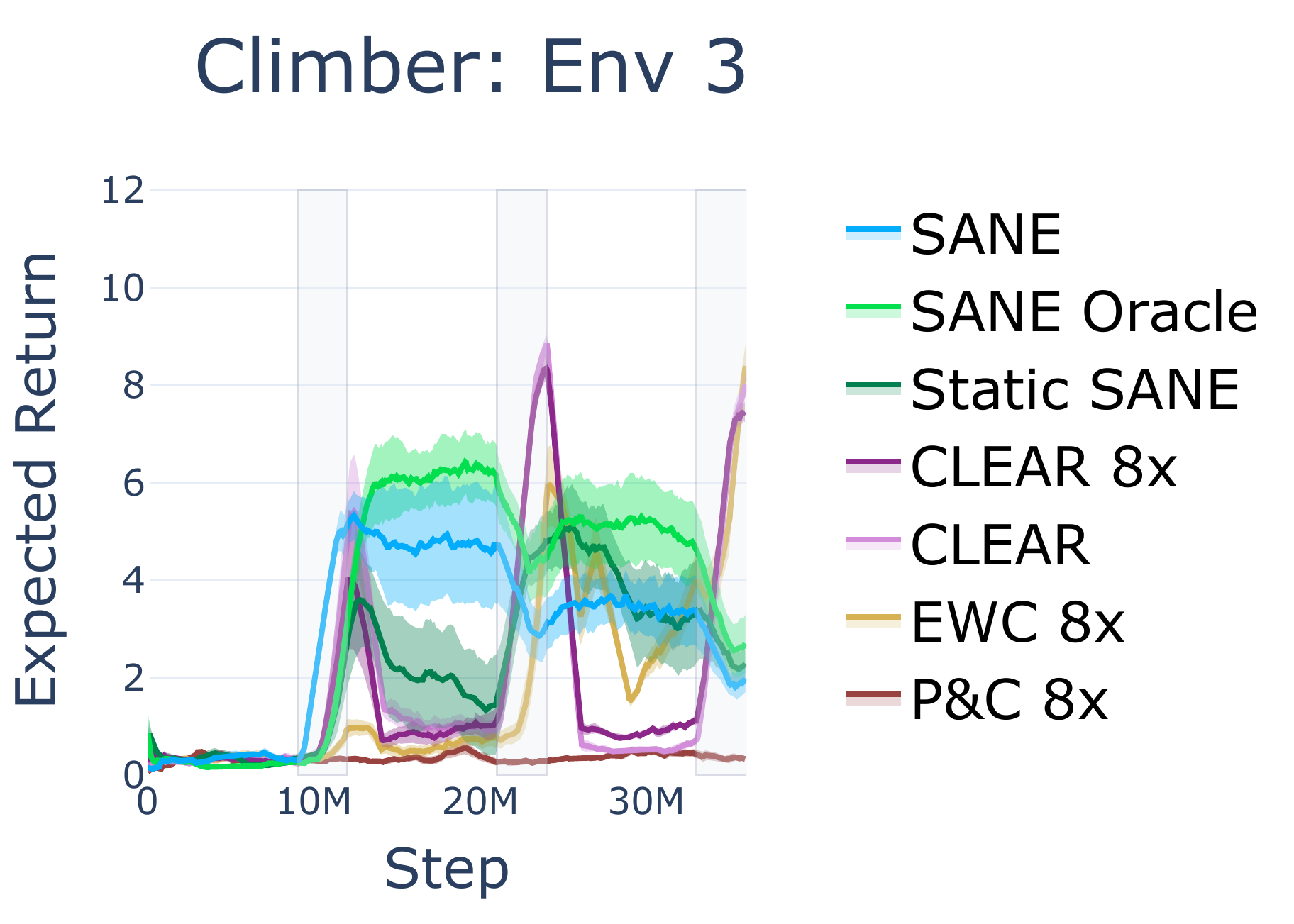}
    \caption{Results for Continual Evaluation on the Climber sequence of tasks. We observe that SANE consistently learns and recalls the tasks. Gray shaded rectangles show when the agent trains on each task.}
    \label{fig:climber_results}
\end{figure}

 We start by analyzing the Continual Evaluation results in Figure \ref{fig:climber_results}. SANE and Static SANE both learn the tasks, but we can see that our dynamic model consistently learns and remembers, while Static SANE overall shows more inconsistent performance, doing particularly poorly on Envs 0 and 2. Both versions of CLEAR learn the tasks but readily forget them, indicating that SANE is not improving by merely adding more parameters. EWC has mixed results; it does worse than SANE uniformly on all cycles of Env 0 and the first cycle of the other Envs, but approximately ties it on the other cycles of Envs 2 and 3, and exceeds it on the other cycles of Env 1. P\&C largely fails to learn the tasks at all, with some exception on Env 2. 
 
 These results are further validated by looking at the Forgetting summary statistics presented in Table \ref{table:forgetting_stats}. By this metric EWC does the best, likely aided by poorer learning during the first cycle of Envs 0 and 1 and the particularly good later performance on Env 1. Of the four methods that learned all tasks immediately (SANE, Static SANE, CLEAR, and CLEAR 8x), SANE exhibits the least forgetting.
 
\begin{table}
\centering
\begin{tabular}{lccccccc}
    \toprule
     & SANE & Static SANE & CLEAR & CLEAR 8x & EWC 8x & P\&C 8x & SANE Oracle\\
    \midrule
    Climber & \cellcolor{red!17} 4.4 ± 1.0  & \cellcolor{red!21} 5.3 ± 0.5 & \cellcolor{red!30} 7.7 ± 0.1 & \cellcolor{red!32} 8.1 ± 0.2 & \cellcolor{red!1} 0.4 ± 0.3 & \cellcolor{red!6} 1.5 ± 0.8 & \cellcolor{green!2} -0.6 ± 0.1\\
    Miner & \cellcolor{green!0} -0.2 ± 0.2 & \cellcolor{red!5} 1.4 ± 1.1 & \cellcolor{red!25} 6.3 ± 0.3 & \cellcolor{red!25} 6.3 ± 0.2 & \cellcolor{red!16} 4.1 ± 0.8 & \cellcolor{red!0} 0.1 ± 0.1 & \cellcolor{green!3} -0.9 ± 0.4\\
    Fruitbot  & \cellcolor{red!22} 5.5 ± 0.7 & \cellcolor{red!23} 5.8 ± 0.3 & \cellcolor{red!30} 7.6 ± 0.4 & \cellcolor{red!25} 6.3 ± 0.6 & \cellcolor{red!14} 3.7 ± 0.7 & \cellcolor{red!6} 1.6 ± 0.1 & \cellcolor{red!1} 0.3 ± 0.1\\
    \bottomrule
\end{tabular}
\caption{Forgetting $(\mathcal{F})$ summary statistics for all experiments. EWC and P\&C exhibit little forgetting because they also exhibit little learning. Of the methods that learned the tasks, we see SANE performs best.}
\label{table:forgetting_stats}
\end{table}


\noindent {\bf Miner:} We additionally demonstrate SANE on Miner, a task where the agent must dig through dirt in two dimensions, collecting diamonds and going to a specified end-goal without getting crushed by rocks. 

\begin{figure}[h]
    \centering
    \includegraphics[trim=0 0em 22em 0, clip,width=0.18 \columnwidth]{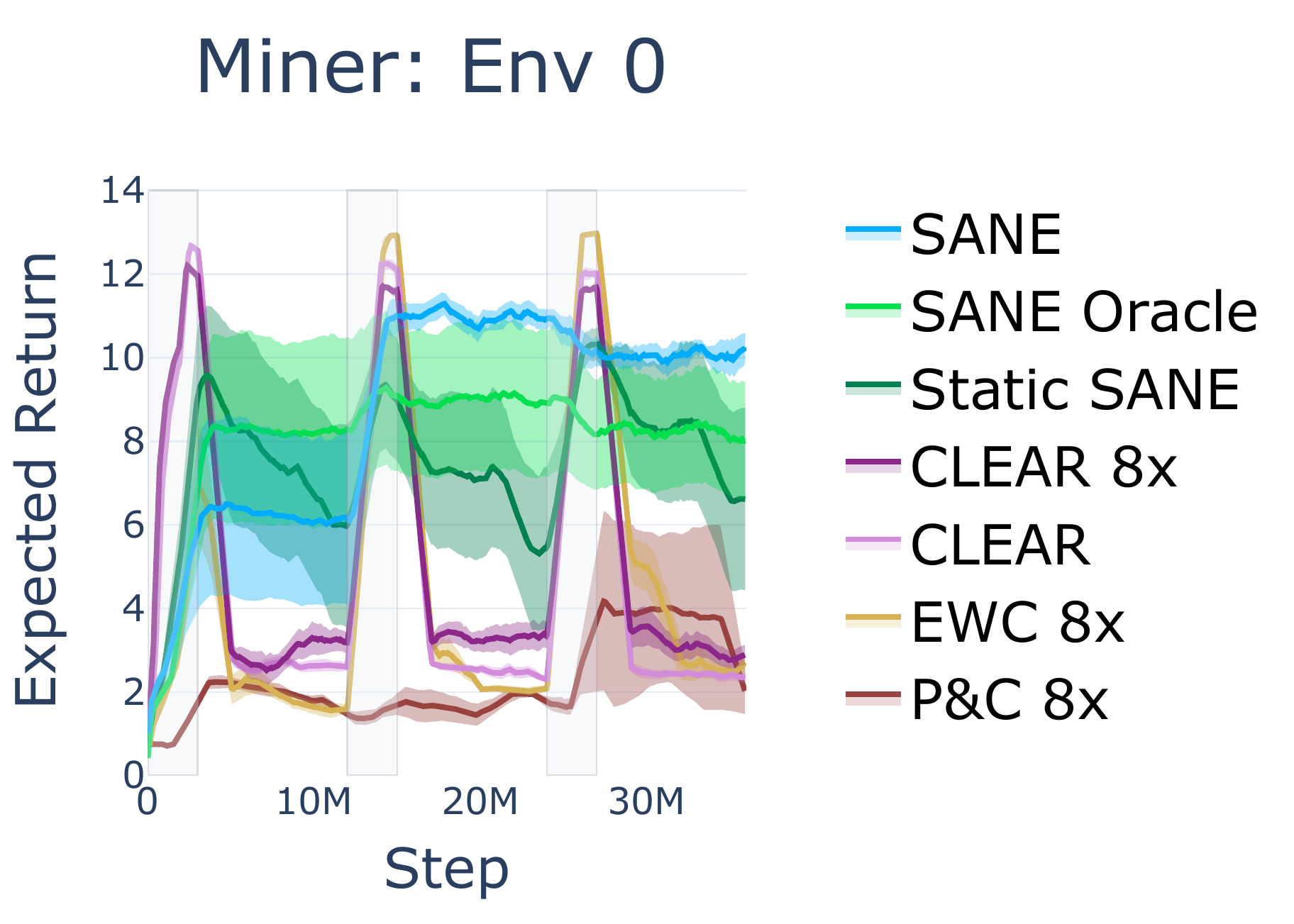}\includegraphics[trim=0 0em 22em 0, clip,width=0.18 \columnwidth]{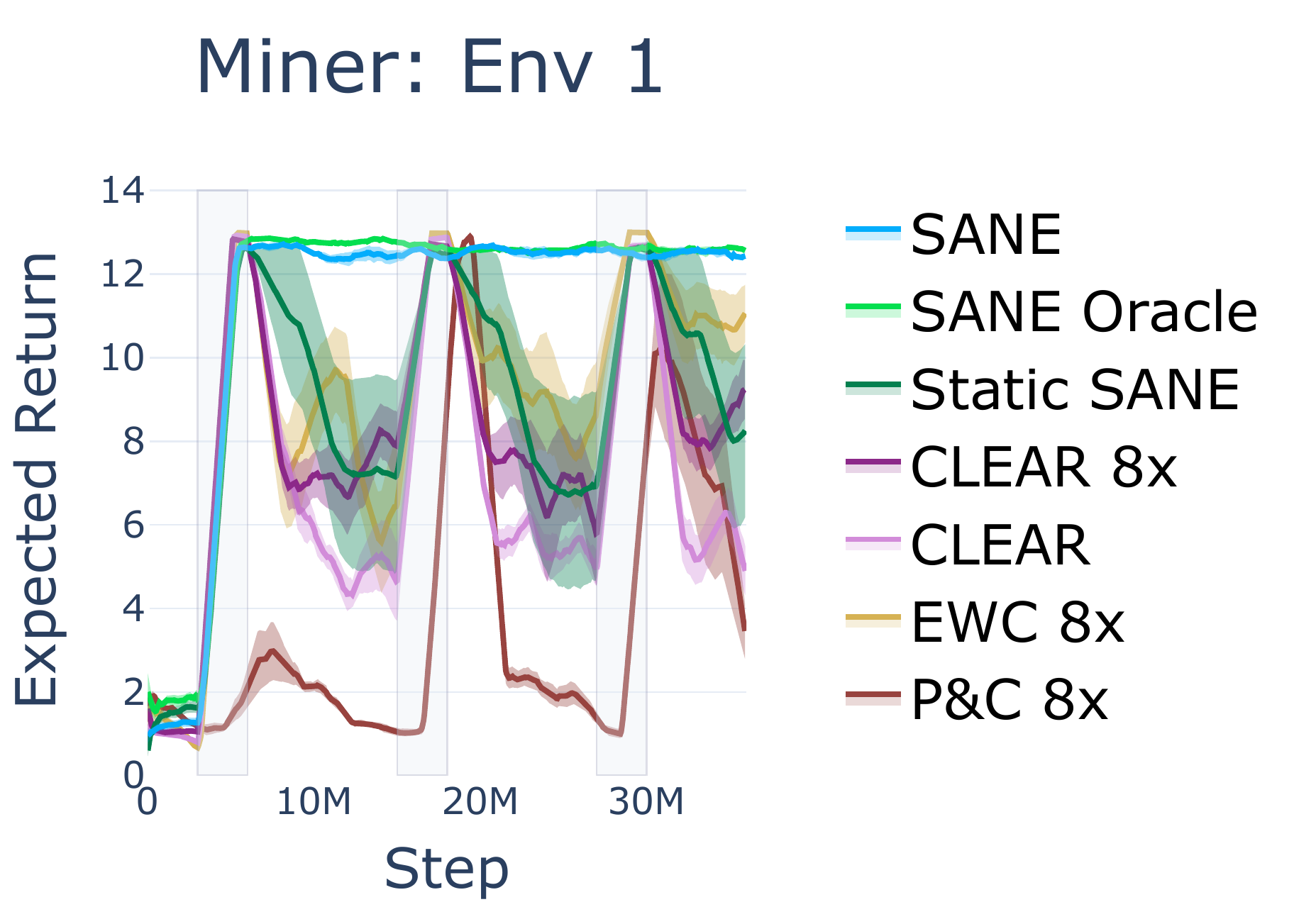}
    \includegraphics[trim=0 0em 22em 0, clip,width=0.18 \columnwidth]{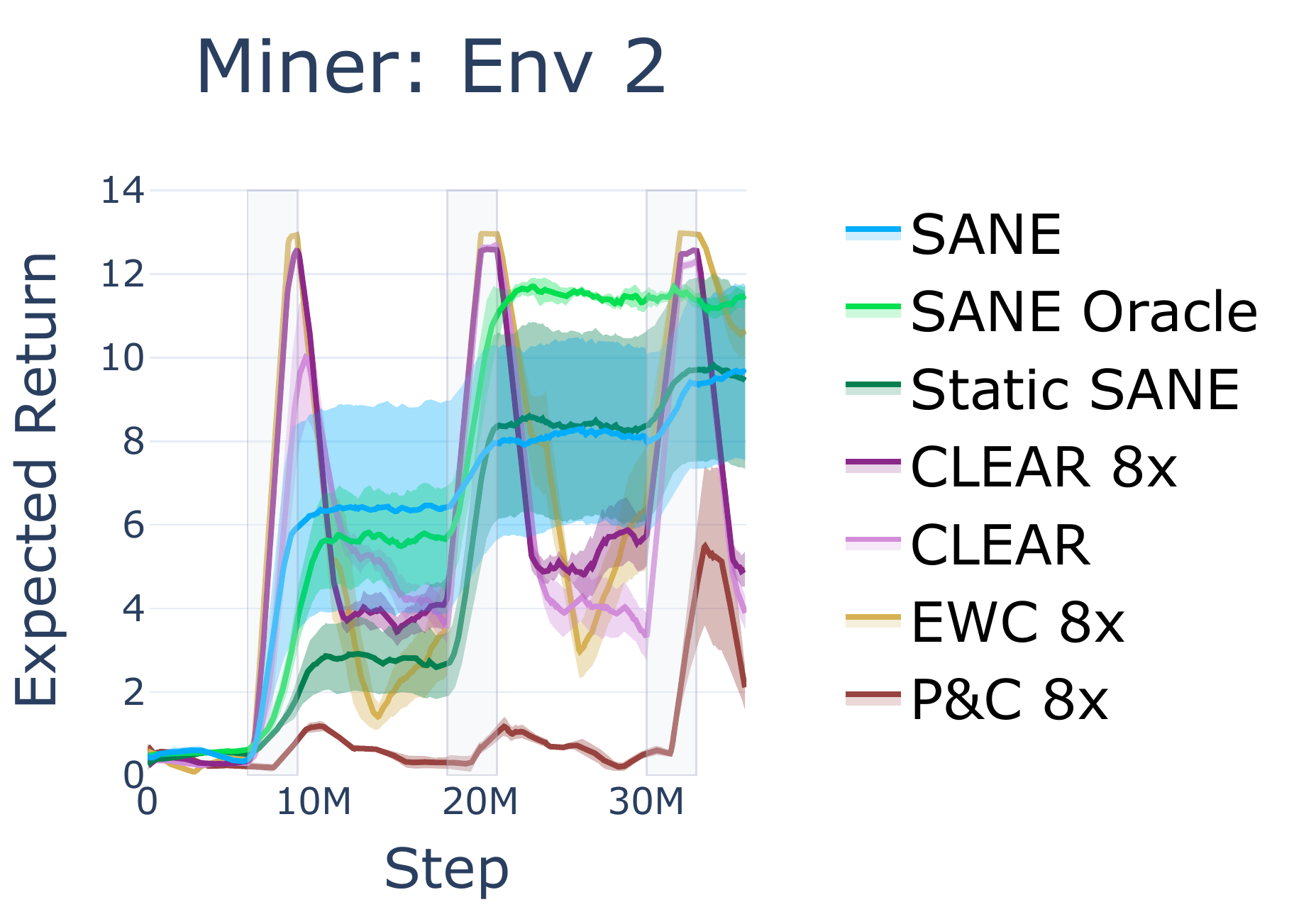}
    \includegraphics[width=0.31 \columnwidth]{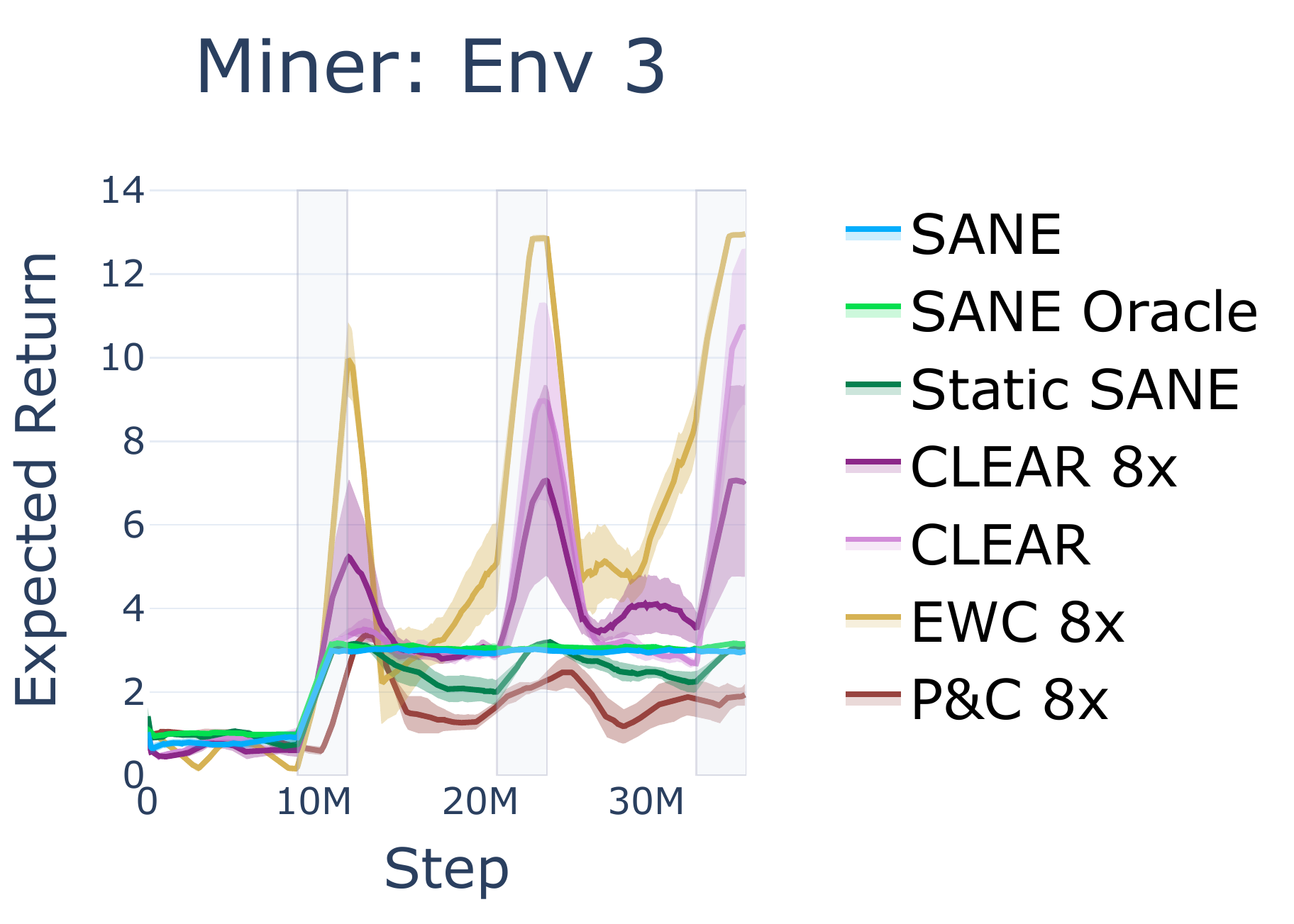}
    \caption{Results for Continual Evaluation on the Miner task sequence. We again observe that SANE improves on the baselines at recall across the tasks. Gray shaded rectangles show when the agent trains on each task.}
    \label{fig:miner_results}
\end{figure}

Continual Evaluation results are shown in Figure \ref{fig:miner_results}. SANE overall outperforms the baselines on the first three environments; we see CLEAR and EWC learning and forgetting, static SANE showing more recall than CLEAR but less than SANE, and largely little learning from P\&C with the exception of Env 1. However, on Env 2 one of SANE's seeds fails to learn the task, and on Env 3 all seeds did. Perhaps CLEAR's larger buffer effectively provides more exploration, as there is more randomness amongst the batches selected to be trained upon.

Table \ref{table:forgetting_stats} again demonstrates numerically these qualitative results. We see that of the methods that learned the tasks, SANE not only did the best, it also exhibited some backwards transfer (negative forgetting). 

\begin{figure}[t]
    \centering
    \includegraphics[trim=0 0em 20em 0, clip,width=0.17 \textwidth]{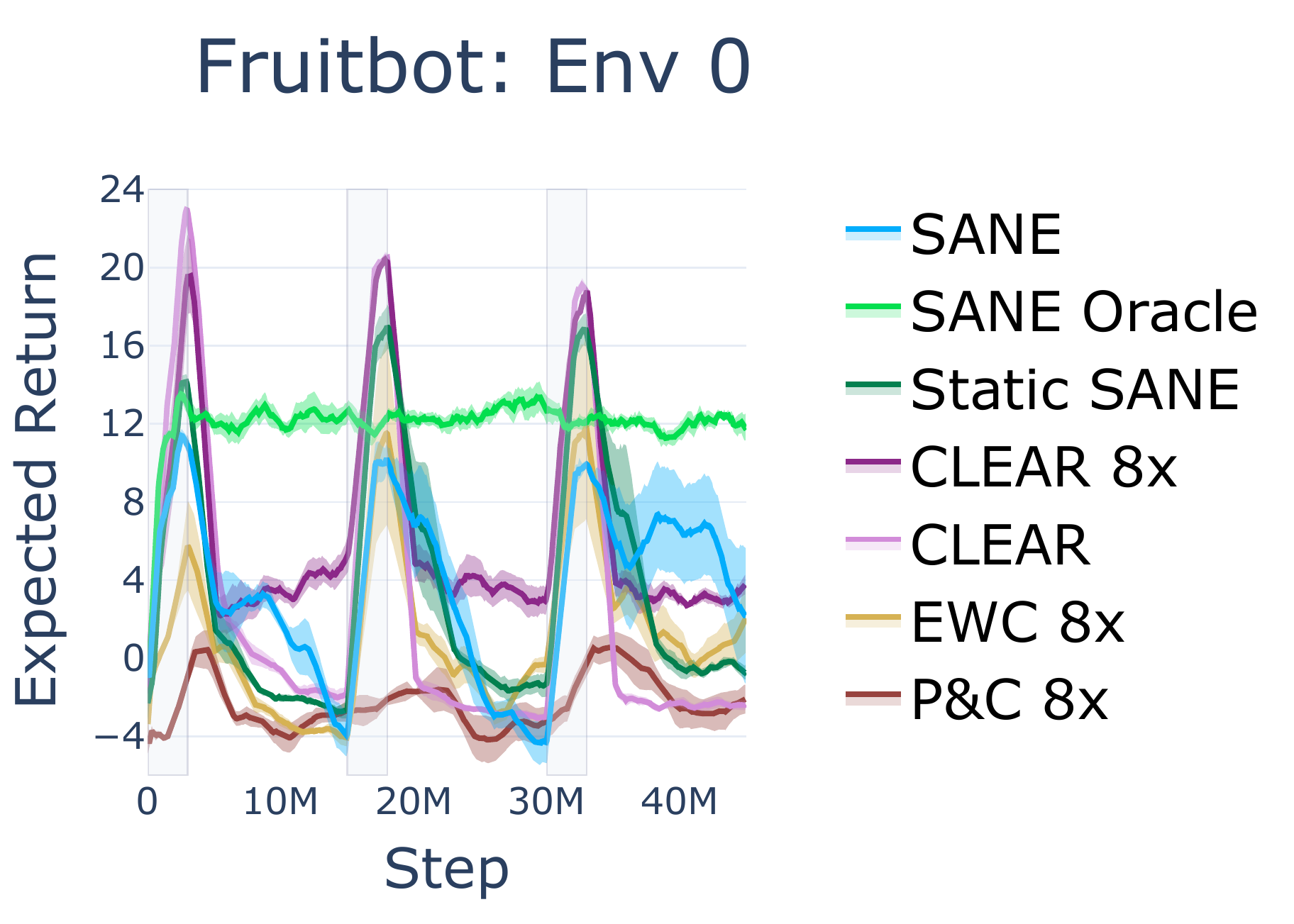}
    \includegraphics[trim=0 0em 20em 0, clip,width=0.17 \textwidth]{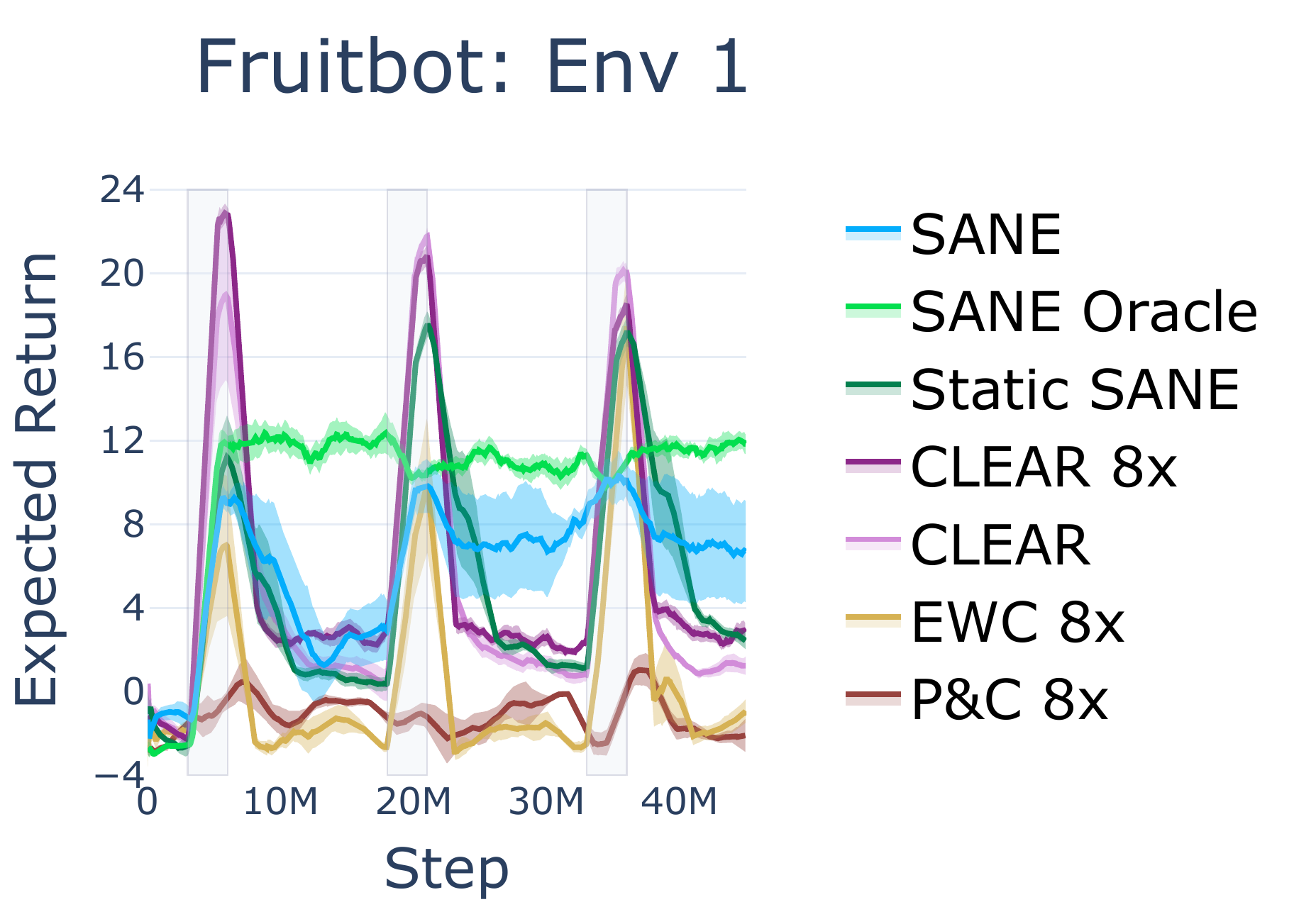}
    \includegraphics[trim=0 0em 20em 0, clip,width=0.17 \textwidth]{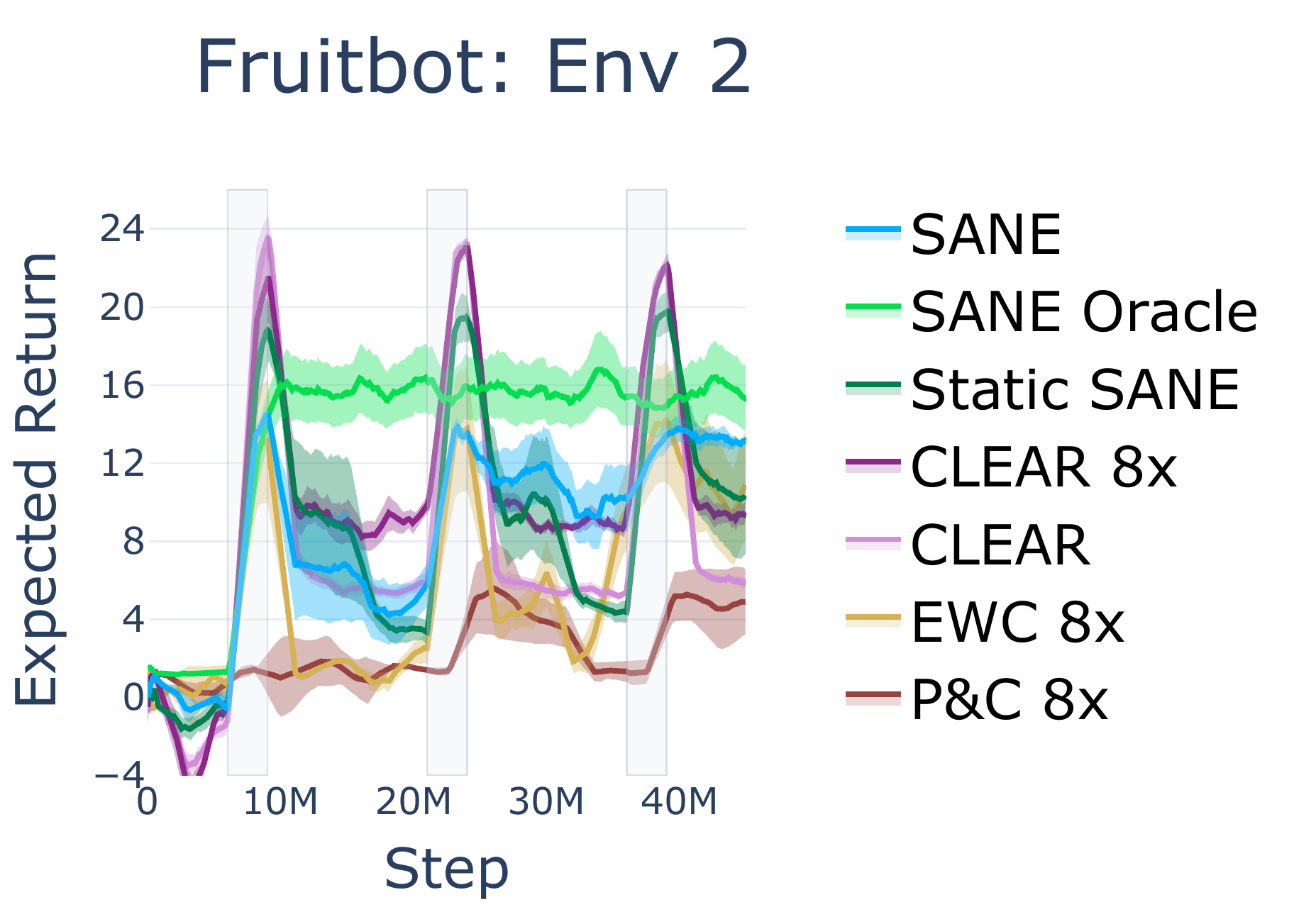}
    \includegraphics[trim=0 0em 20em 0, clip,width=0.17 \textwidth]{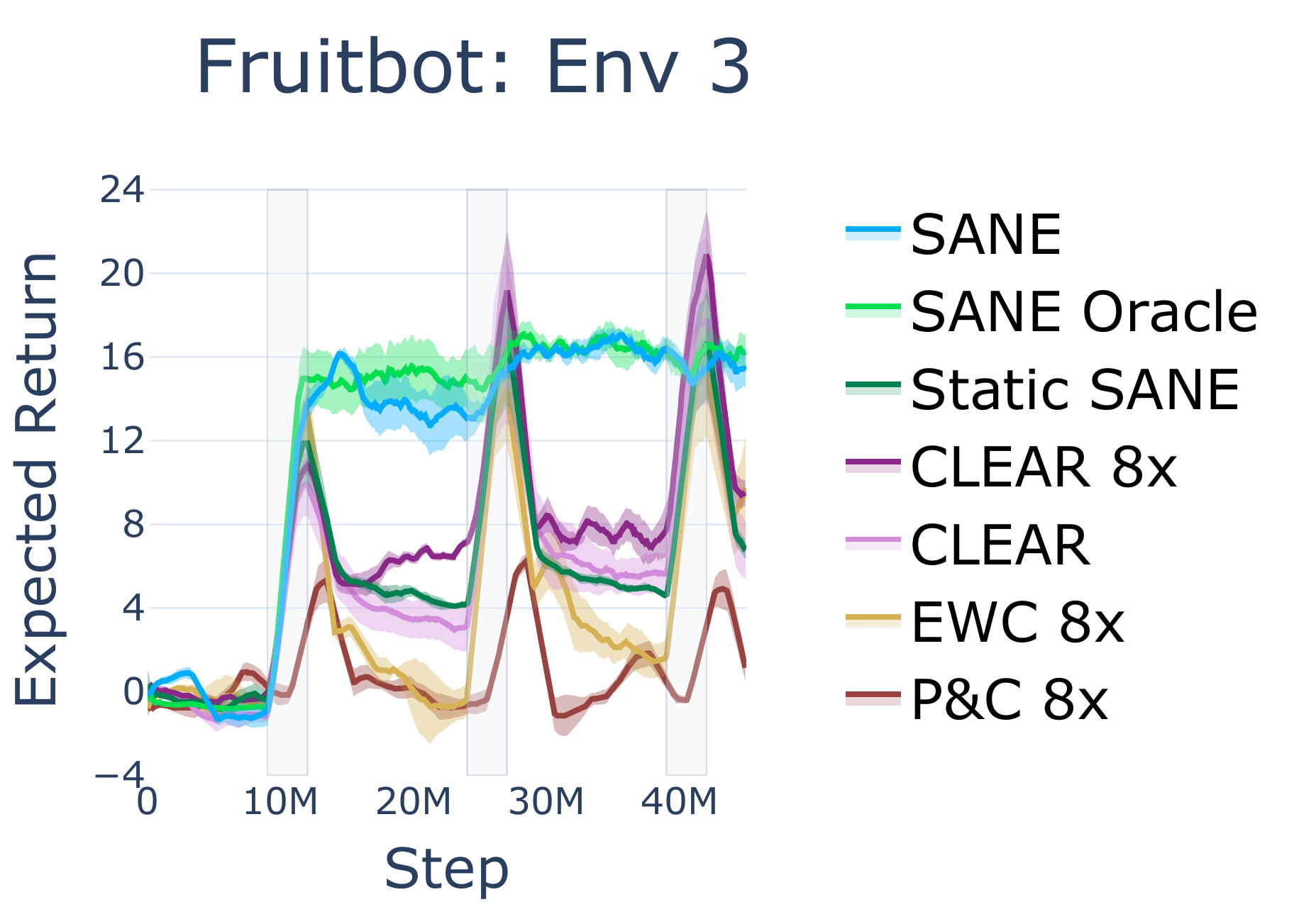}
    \includegraphics[trim=0 0em 0em 0, clip,width=0.275 \textwidth]{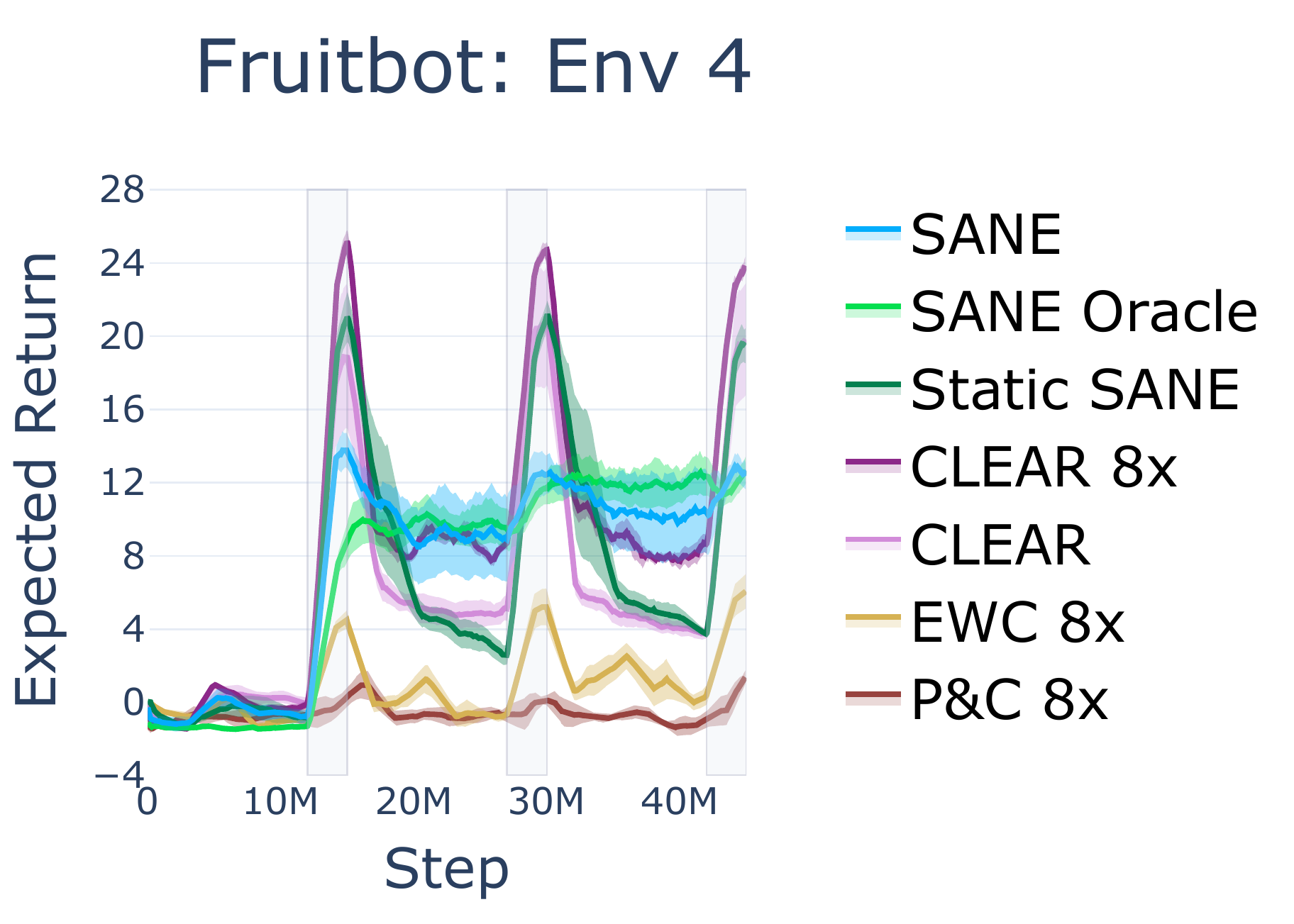}
    \caption{Results on the Fruitbot sequence. SANE performs particularly well on Envs 1 and 3, comparable to CLEAR 8x on Envs 2 and 4, and struggles some with Env 0. Gray shaded rectangles show when the agent trains on each task.}
    \label{fig:fruitbot_results}
\end{figure}

\noindent {\bf Fruitbot:} The final Procgen sequence we use is based on Fruitbot, where the environment continuously scrolls and the agent must move left and right to collect fruit, avoid vegetables, and make it through gaps in the wall. Continual Evaluation results, shown in Figure \ref{fig:fruitbot_results}, are less clear-cut than the previous two experiments. SANE clearly exceeds baselines on recall on Envs 1 and 3, but remains comparable to the CLEAR 8x baseline on Envs 2 and 4, and struggles on Env 0, only exceeding the baseline in the final cycle. Furthermore for the most part SANE receives a lower maximum score than the CLEAR baselines, with the exception of Env 3. Table \ref{table:forgetting_stats} shows that despite the mixed qualitative results, SANE again exceeds baselines quantitatively.


\section{Analysis of SANE}

We generate two additional figures to help us analyze our SANE ensembles. The first is a \textit{module ID} plot. On creation we assign every module a unique identifier: an integer that increases per module created. This ID is constant through the lifetime of the module, including when other modules get merged into it. We can plot what module is active by plotting its module ID. This allows us to see when there are periods of rapid creation (steep regions of the graph), when older modules are re-used, and when modules are being stably activated. 

The second plot is a \textit{lineage} plot, showing a graph that represents the history of the ensemble, with each node in the graph representing a module. A blue line indicates that one module spawned another, and a red line indicates that a module was merged into another. Light blue nodes represent modules that are current available to be activated at the current time. An example lineage plot\footnote{An interactive lineage plot can also be viewed at \github.} is shown in Appendix \ref{section:fruitbot_lineage}.

\subsection{Single Run: Climber}
\label{section:analysis_climber}
\begin{figure}[]
    \centering
    \includegraphics[trim=0 4em 13em 0em, clip,width=0.12 \textwidth]{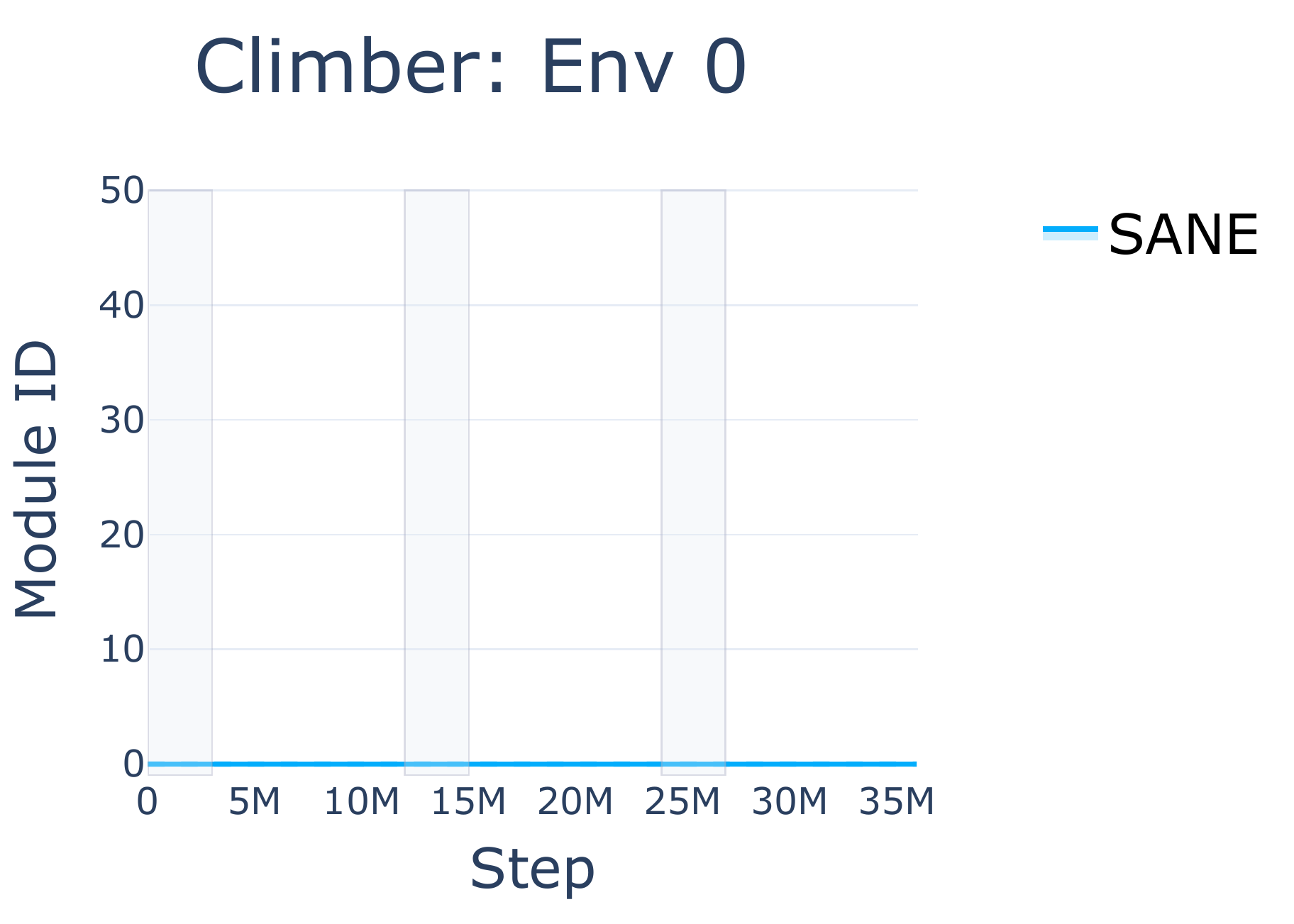}
    \includegraphics[trim=0 4em 13em 0, clip,width=0.12 \textwidth]{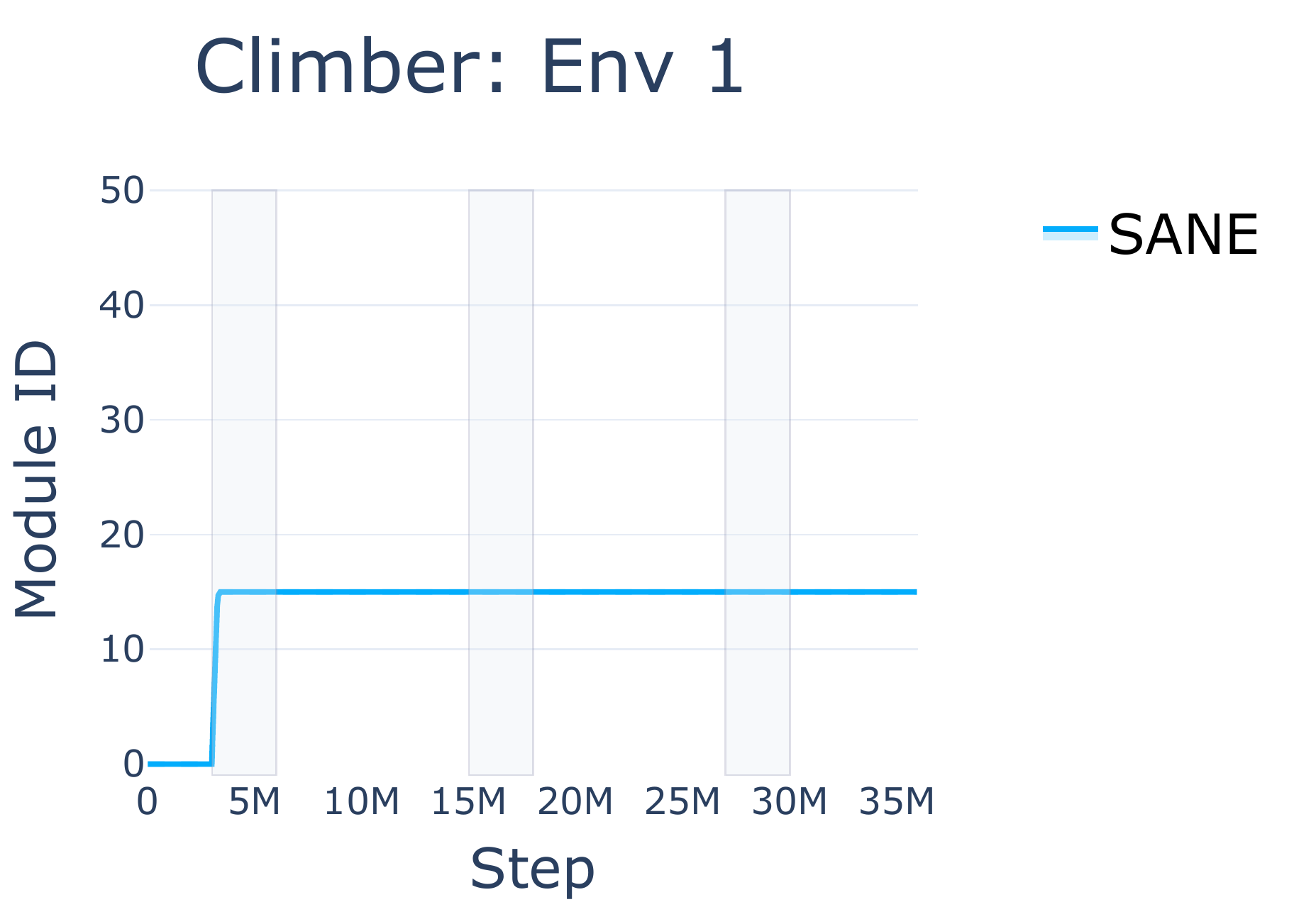}
    \includegraphics[trim=0 4em 13em 0, clip,width=0.12 \textwidth]{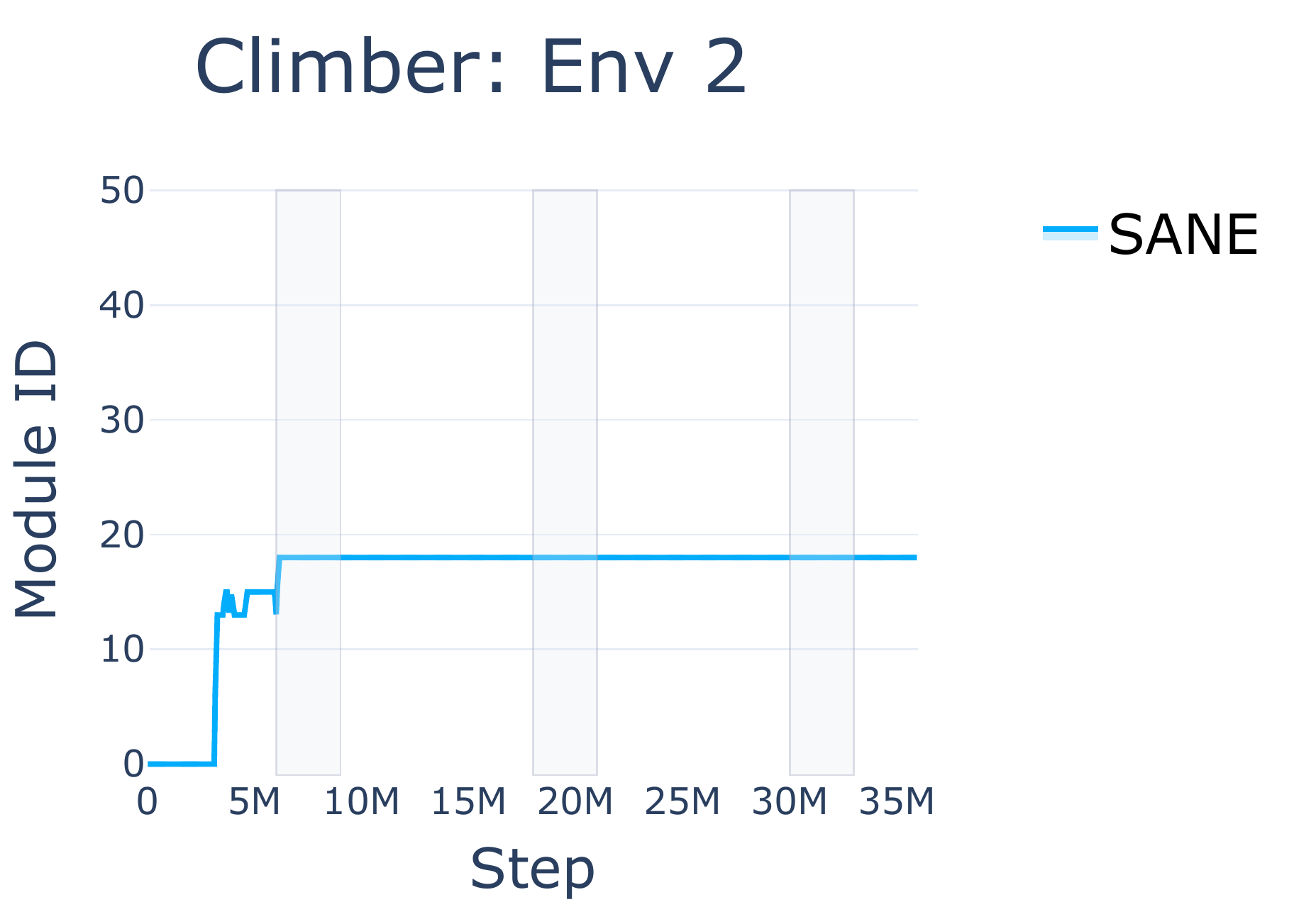}
    \includegraphics[trim=0 4em 0em 0, clip,width=0.16 \textwidth]{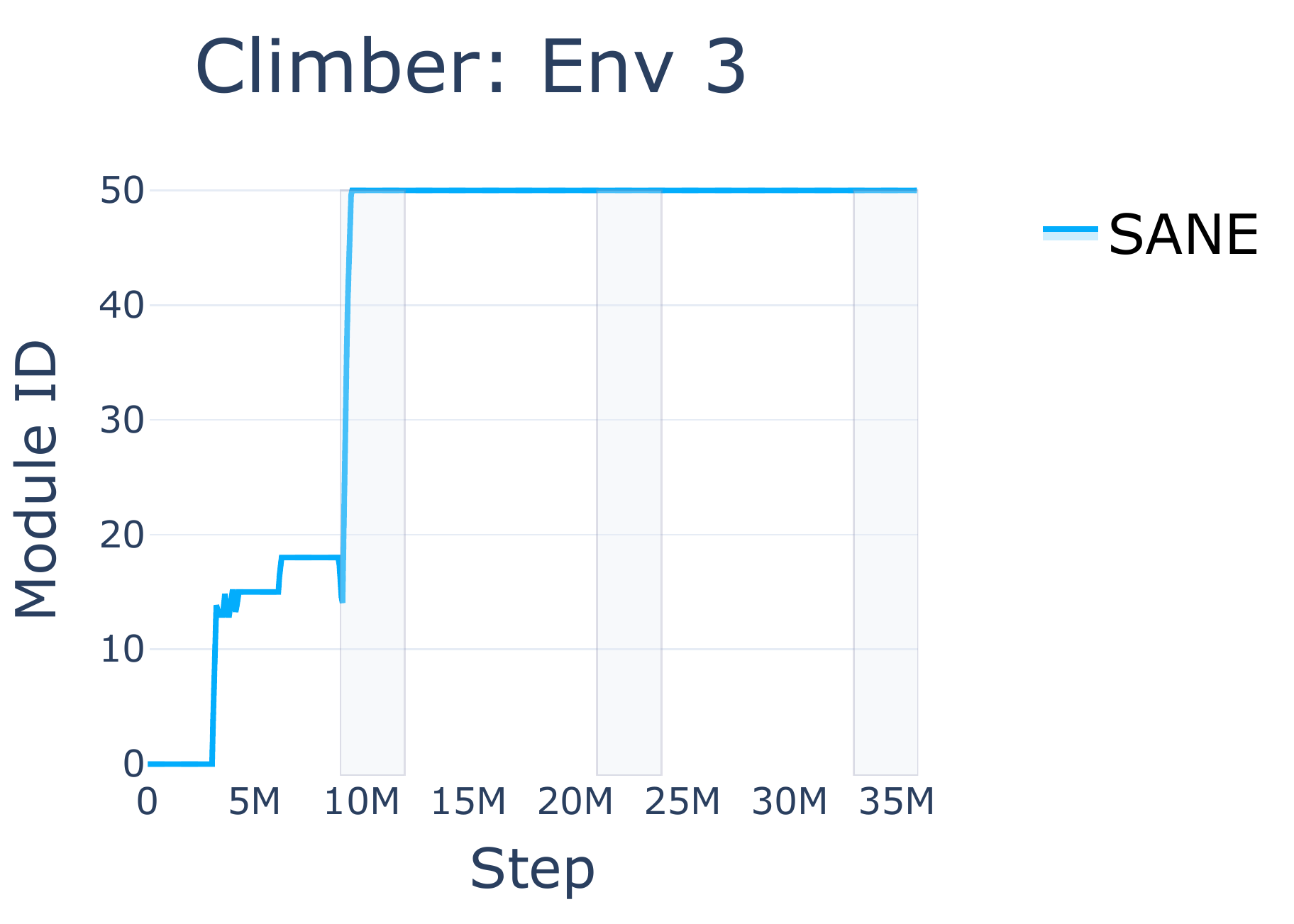}\\
    \includegraphics[trim=0 1em 13em 5em, clip,width=0.12 \textwidth]{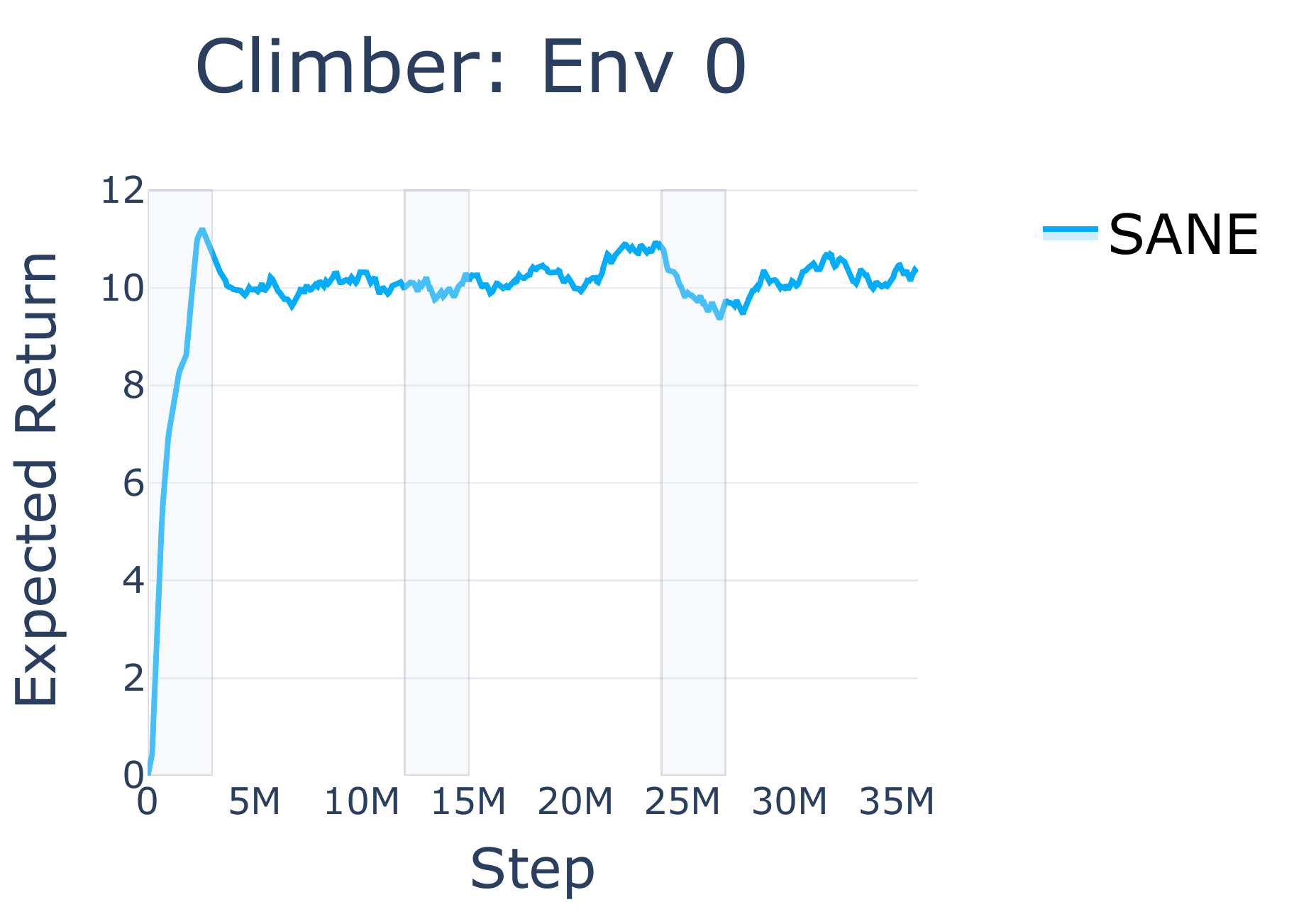}
    \includegraphics[trim=0 1em 13em 5em, clip,width=0.12 \textwidth]{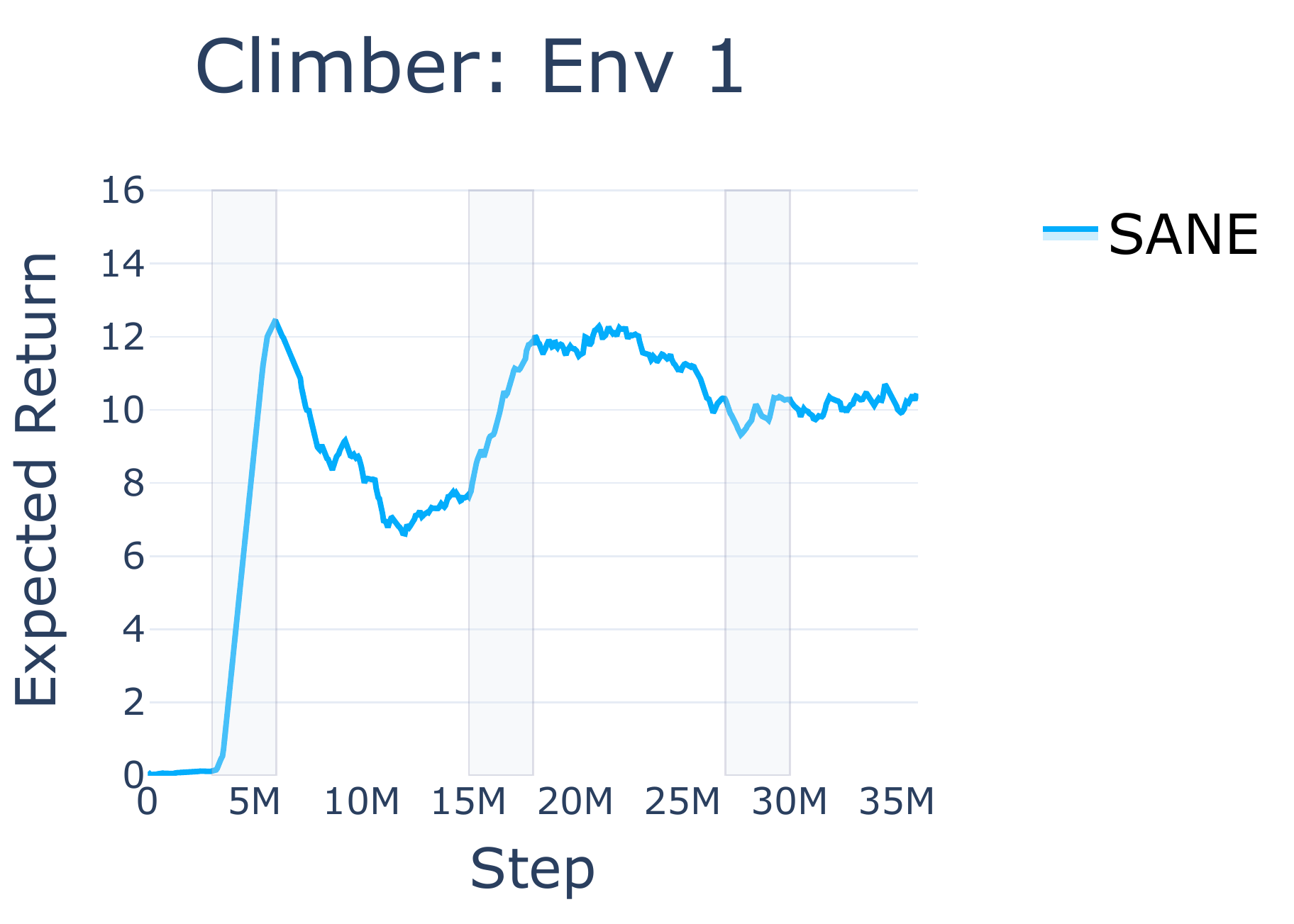}
    \includegraphics[trim=0 1em 13em 5em, clip,width=0.12 \textwidth]{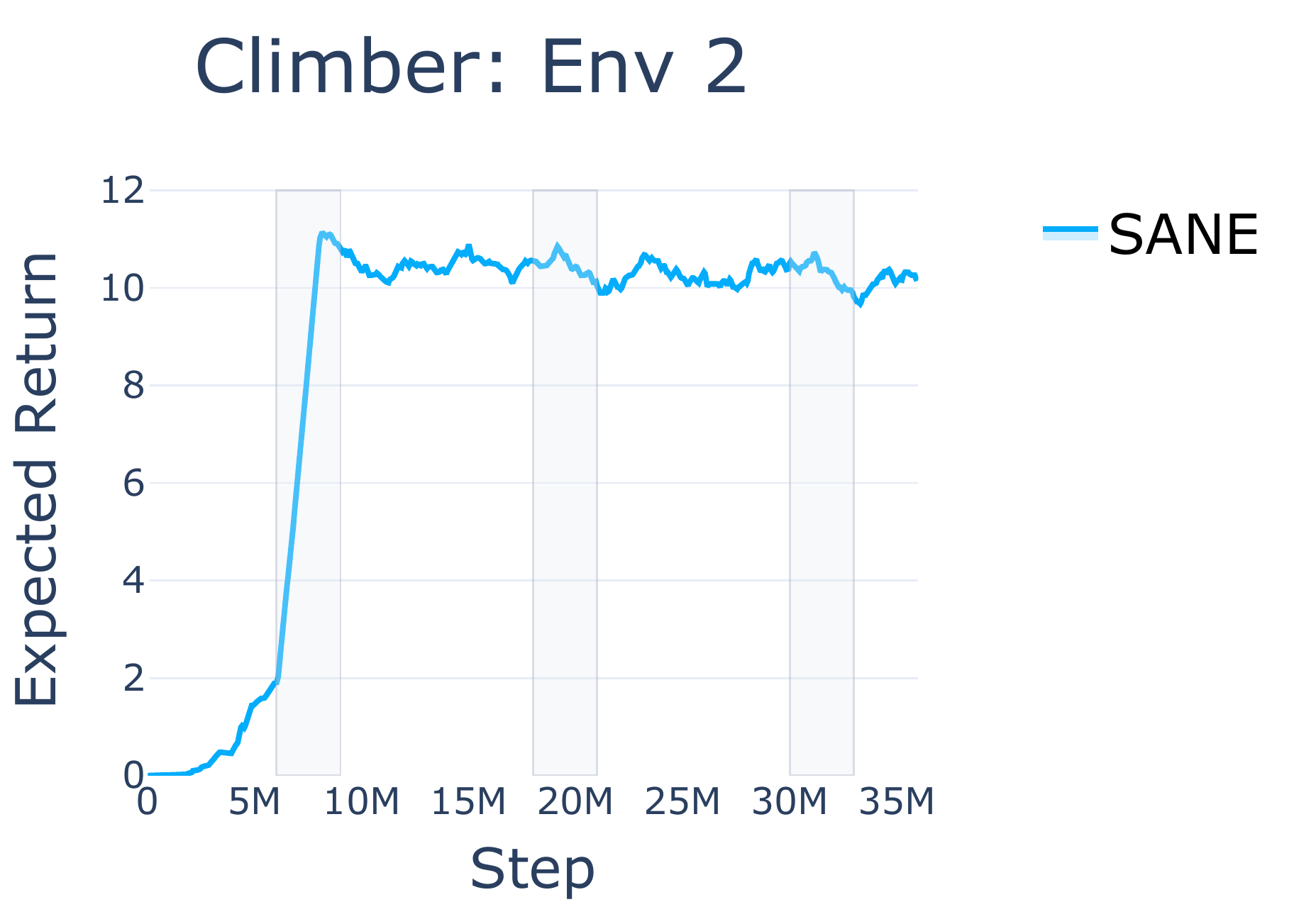}
    \includegraphics[trim=0 1em 0em 5em, clip,width=0.16 \textwidth]{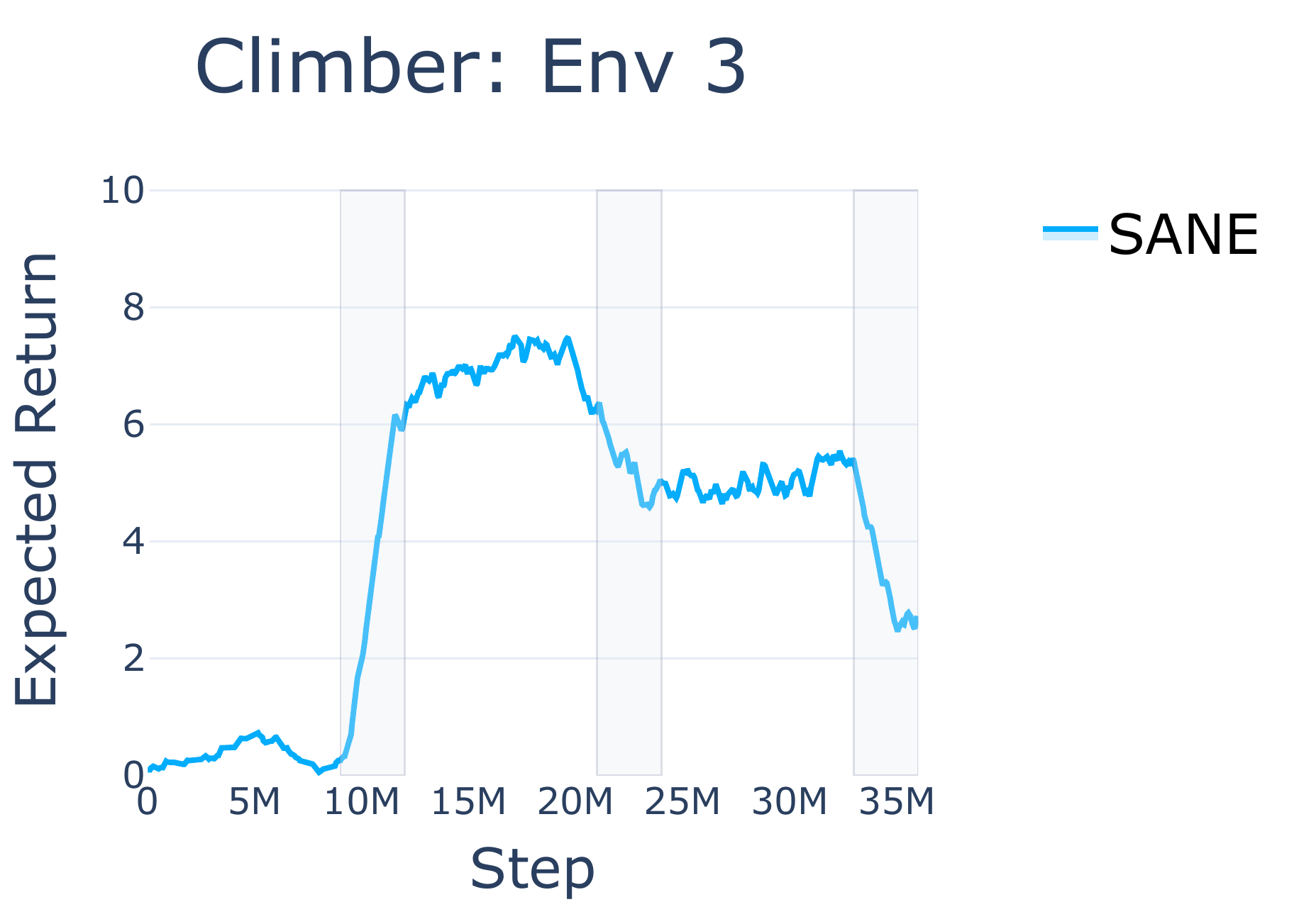}
    \caption{Module ID and expected return plots aligned by timestep, to show module activation during a single run of Climber. Gray shaded rectangles show when the agent trains on each task.}
    \label{fig:climber_single}
\end{figure}

We start by analyzing a single (non-hand-picked) run of SANE in Climber, to demonstrate the dynamics of learning in a simple environment where behaviors are readily separable. In Figure \ref{fig:climber_single}, we aligned a graph of the ID of the currently active module with the reward received at that time. 

We can see the desired behavior in this case: several new modules are created (a sharp increase in module ID is observed) as performance successively fails to meet expectation, until a suitable module is created. Additionally, we can observe that before a task is trained upon, it is likely to use the best module for the current task. E.g. Env 3 uses Env 0's active module (module 0) for the first 3M steps, then Env 1's module (15) for most of the next 3M, then Env 2's module (18) for the next 3M, until during its own training period drift is detected and a custom module is created. 

It is also worth remarking on the decline in Env 3's performance, which occurs particularly while the task is being trained. It occurs while a consistent module is being activated, so is not related to the ensembling behaviors of module creation or merging. Observing the behavior indicates that the agent is jumping into a bat rather than avoiding it, so it would seem it is overfitting to the jumping behavior, possibly as a result of the decreased replay buffer size.


\subsection{Fruitbot Analysis}

Fruitbot performed least well of our experiments, so we dive in further to understand the dynamics at play.

\begin{figure}[]
    \centering
    \includegraphics[trim=0 0em 20em 0, clip,width=0.16 \textwidth]{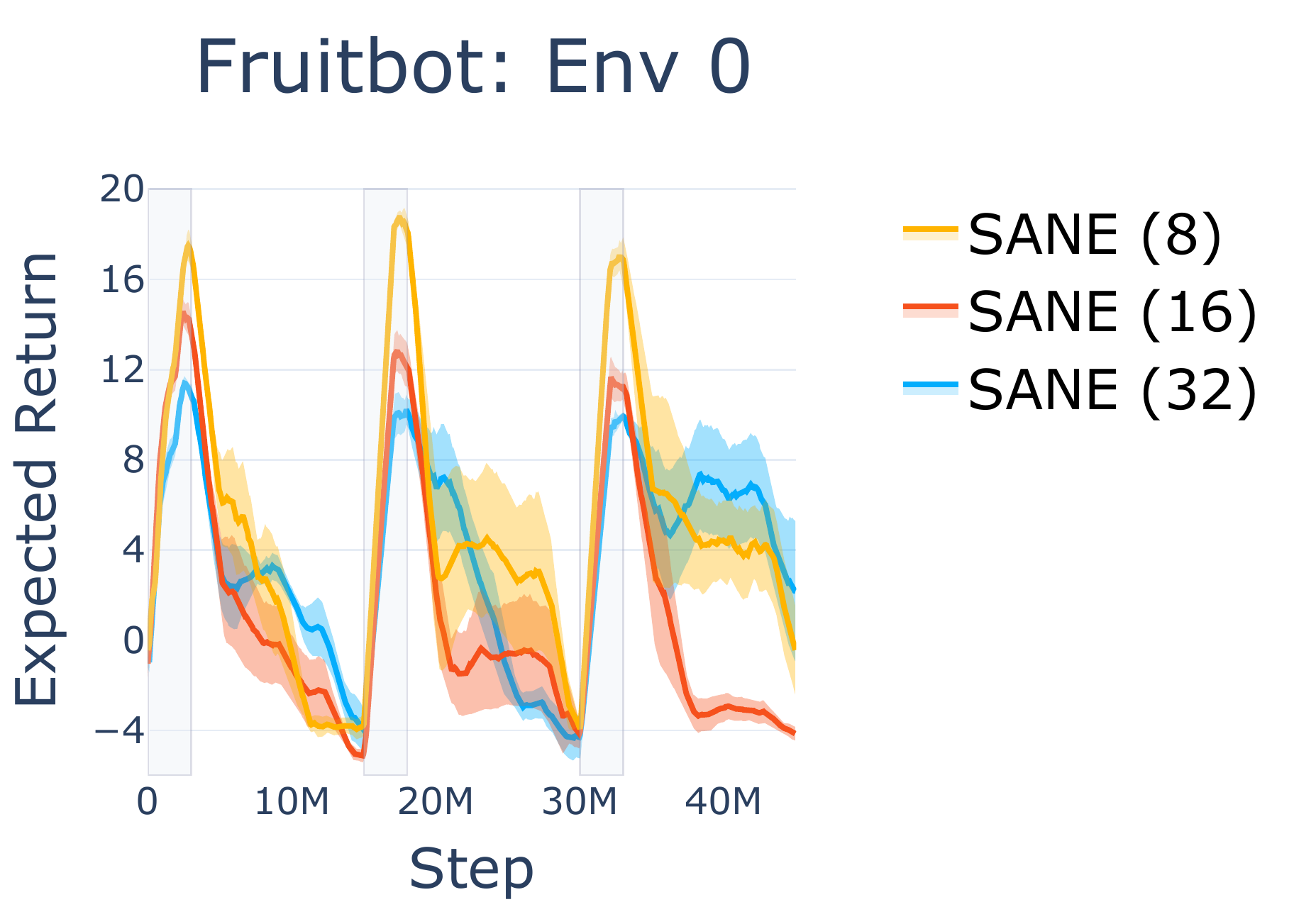}
    \includegraphics[trim=0 0em 20em 0, clip,width=0.16 \textwidth]{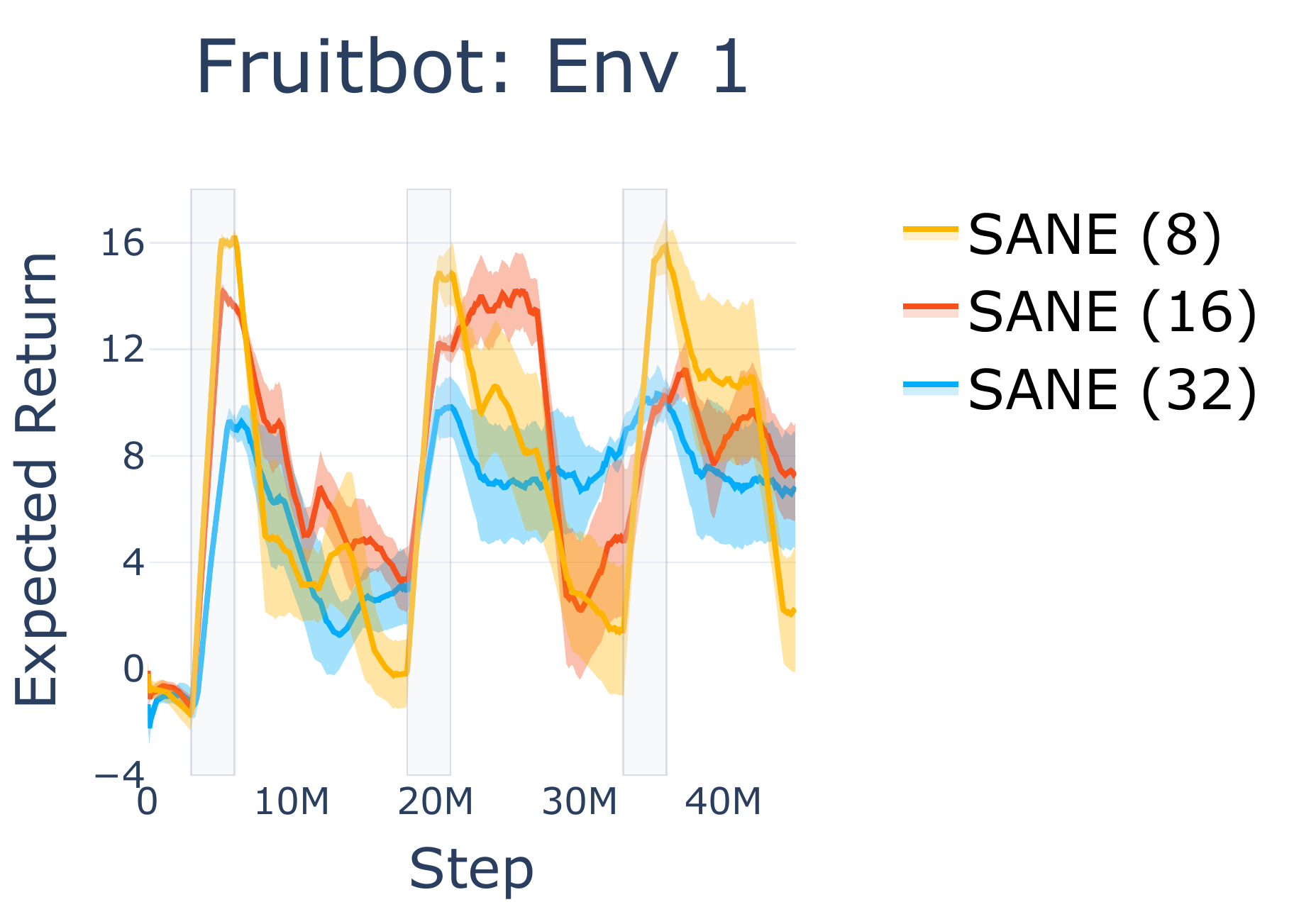}
    \includegraphics[trim=0 0em 20em 0, clip,width=0.16 \textwidth]{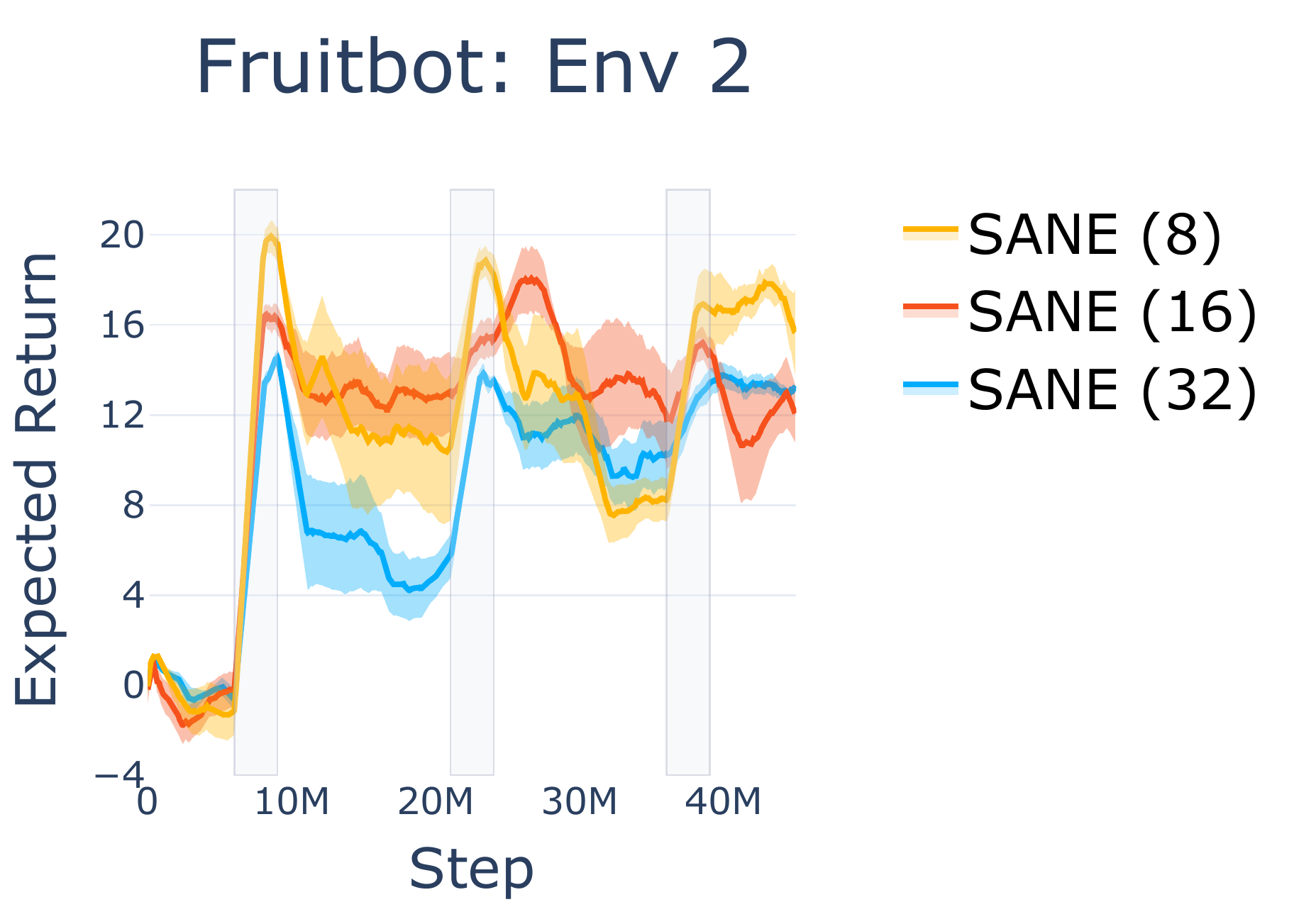}
    \includegraphics[trim=0 0em 20em 0, clip,width=0.16 \textwidth]{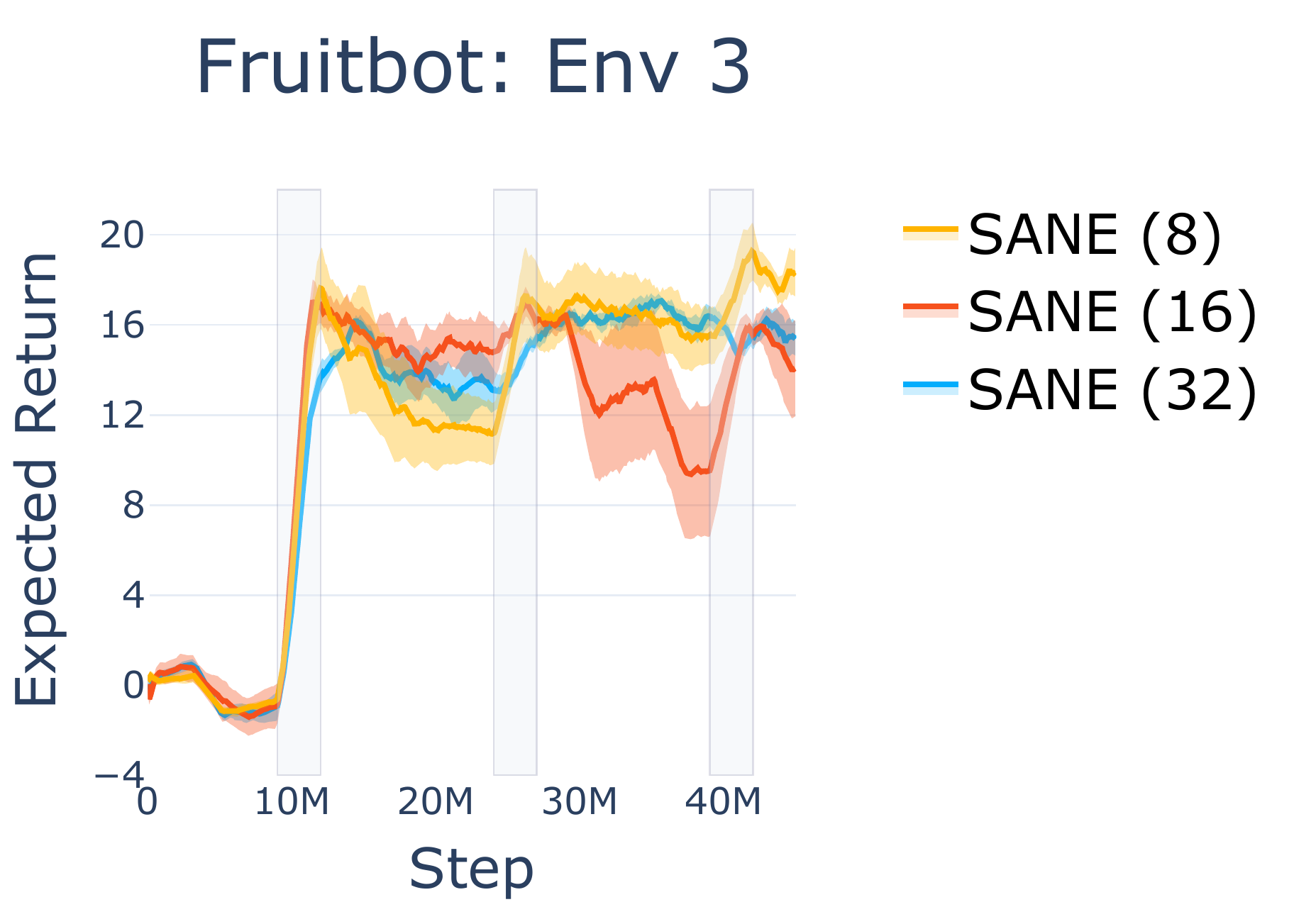}
    \includegraphics[width=0.26 \textwidth]{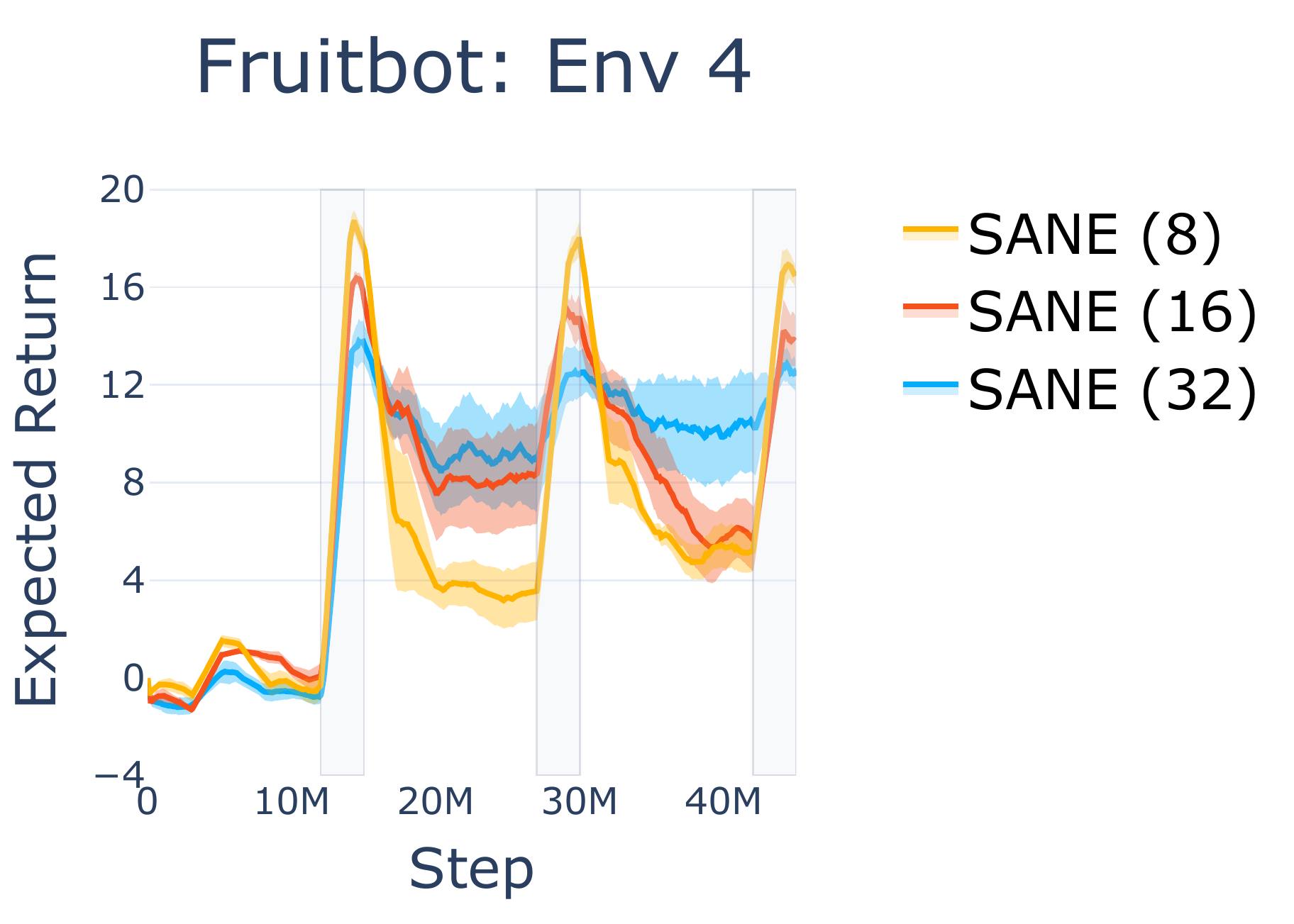}
    \caption{Comparison of SANE variations on the Fruitbot task sequence. The number in parentheses indicates the number of modules in the ensemble, ie. SANE (8) has 8 modules. Gray shaded rectangles show when the agent trains on each task.}
    \label{fig:sane_ablations}
\end{figure}

\noindent {\bf Module Count Ablation:} We first discuss the difference in expected return when we vary the number of modules for SANE, as visualized in Figure \ref{fig:sane_ablations}. Overall, we observe that the fewer the modules, the higher the maximum scores received. The one exception is Env 3, where 8 and 16 modules both receive comparable scores. However, in general the fewer the modules the more forgetting is observed as well. This is particularly noticeable on Envs 0, 1, and 4, with more ambiguity on Envs 2 and 3. 

As we observed in the analysis of Climber, when a new task is switched to, we don't create just a single module. Rather, the critic steadily learns to adapt to the new task; each time it passes the $v^{LCB}$ threshold, a new module is created. Once the maximum number of modules has been reached, this triggers a merge. Since a merge combines the replay buffers of the two modules, when two ``compatible'' modules are merged, the resulting policy is more robust than that of the individual modules. However when two modules representing conflicting behaviors are merged, we see a reduction in performance. Taken together, this means that as merging is occurring, more modules in the ensemble will generally be more stable over time, but might be slower to learn. We discuss a concrete example of this in Section \ref{section:fruitbot_single} below.

\subsubsection{Single Run}
\label{section:fruitbot_single}

\begin{figure}[]
    \centering
    \includegraphics[trim=0 4em 15em 0em, clip, width=0.15 \columnwidth]{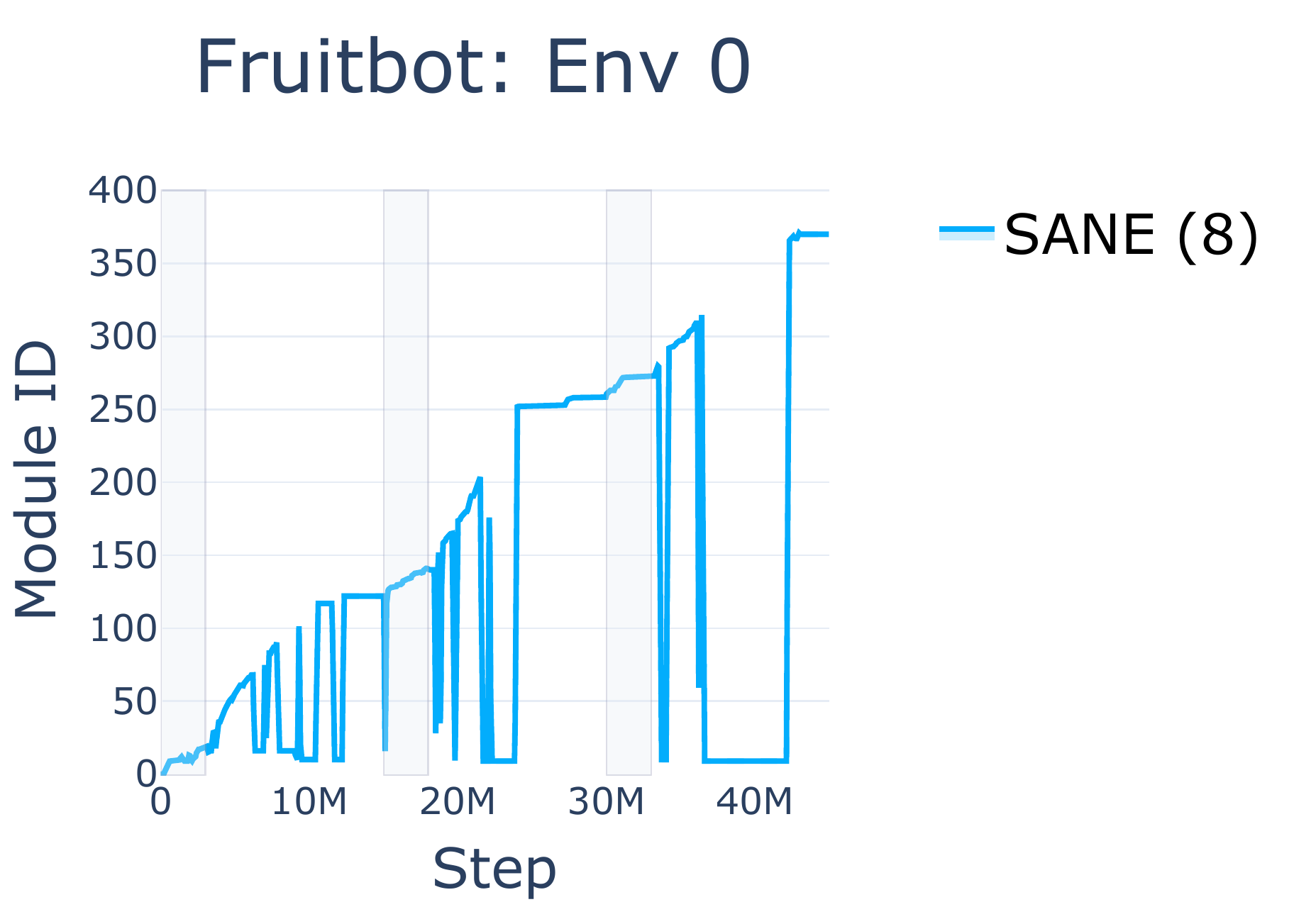}
    \begin{tikzpicture}
        \draw (0, 0) node[inner sep=0] {\includegraphics[trim=0 4em 15em 0, clip,width=0.15 \columnwidth]{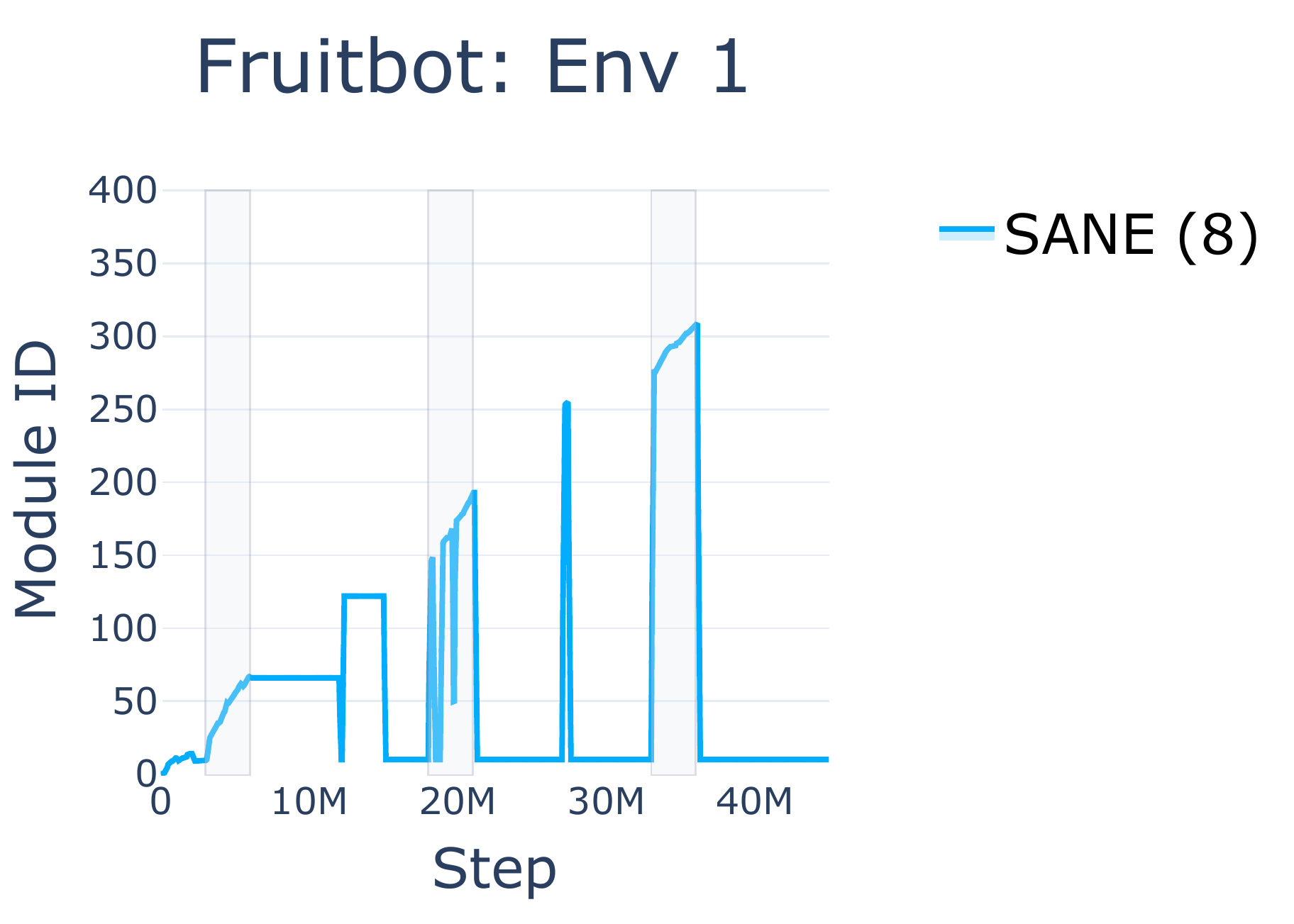}};
        \draw (0.1, 0.03) node {\tiny A};
    \end{tikzpicture}
    \begin{tikzpicture}
        \draw (0, 0) node[inner sep=0] {\includegraphics[trim=0 4em 15em 0, clip,width=0.15 \columnwidth]{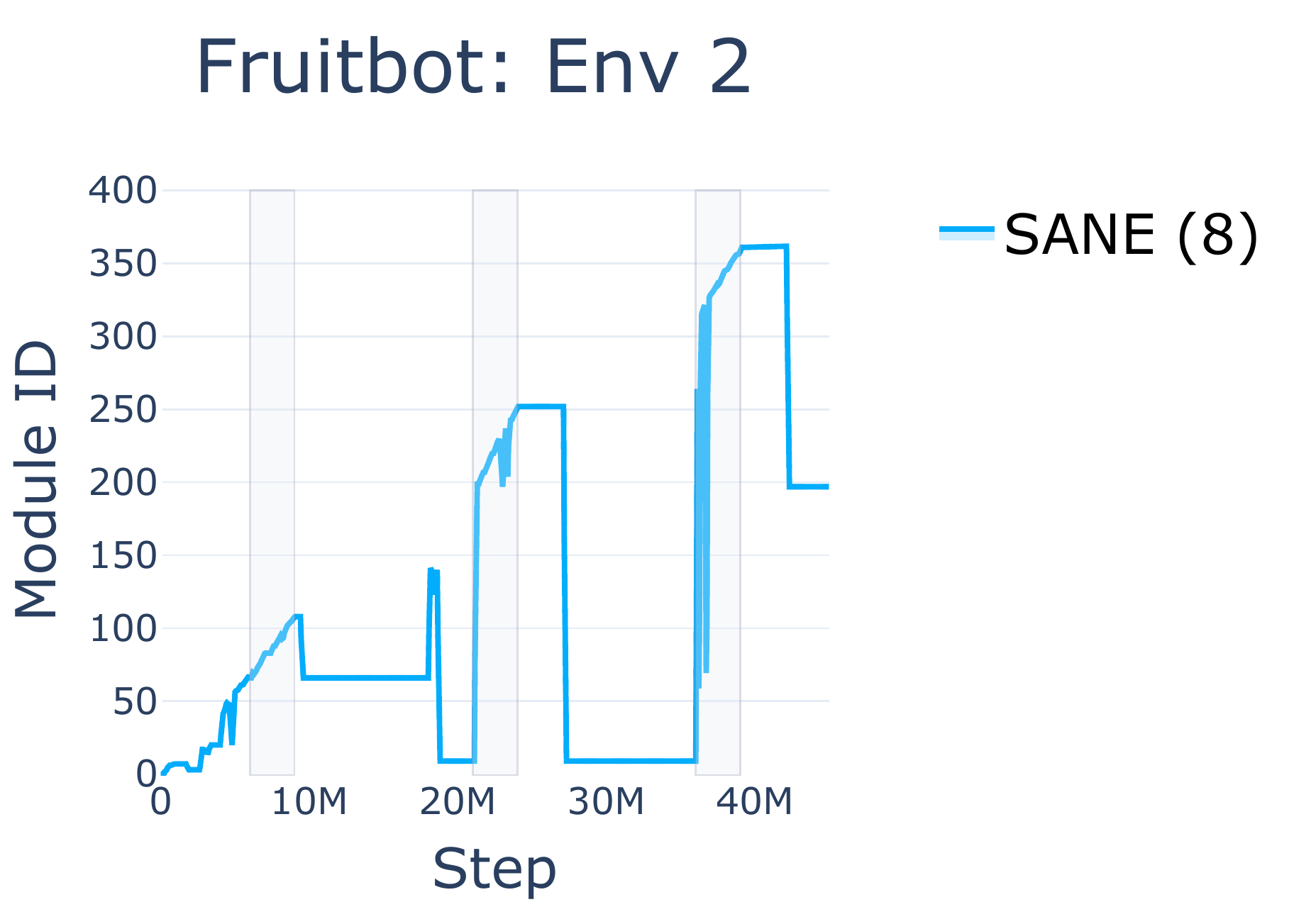}};
        \draw (0.35, 0.18) node {\tiny B};
    \end{tikzpicture}
    \includegraphics[trim=0 4em 15em 0, clip,width=0.15 \columnwidth]{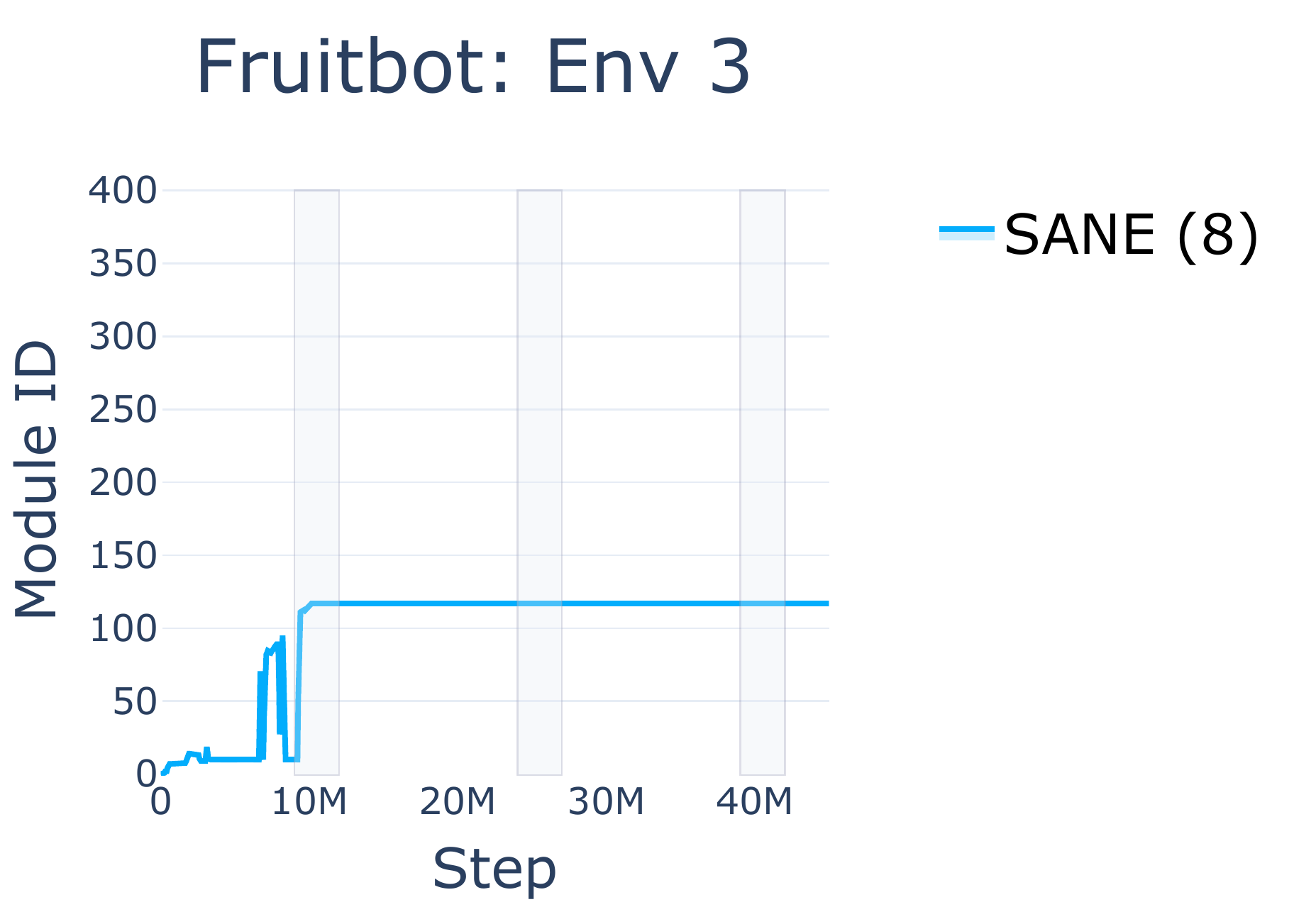}
    \begin{tikzpicture}
        \draw (0, 0) node[inner sep=0] {\includegraphics[trim=0 4em 0em 0, clip,width=0.21 \columnwidth]{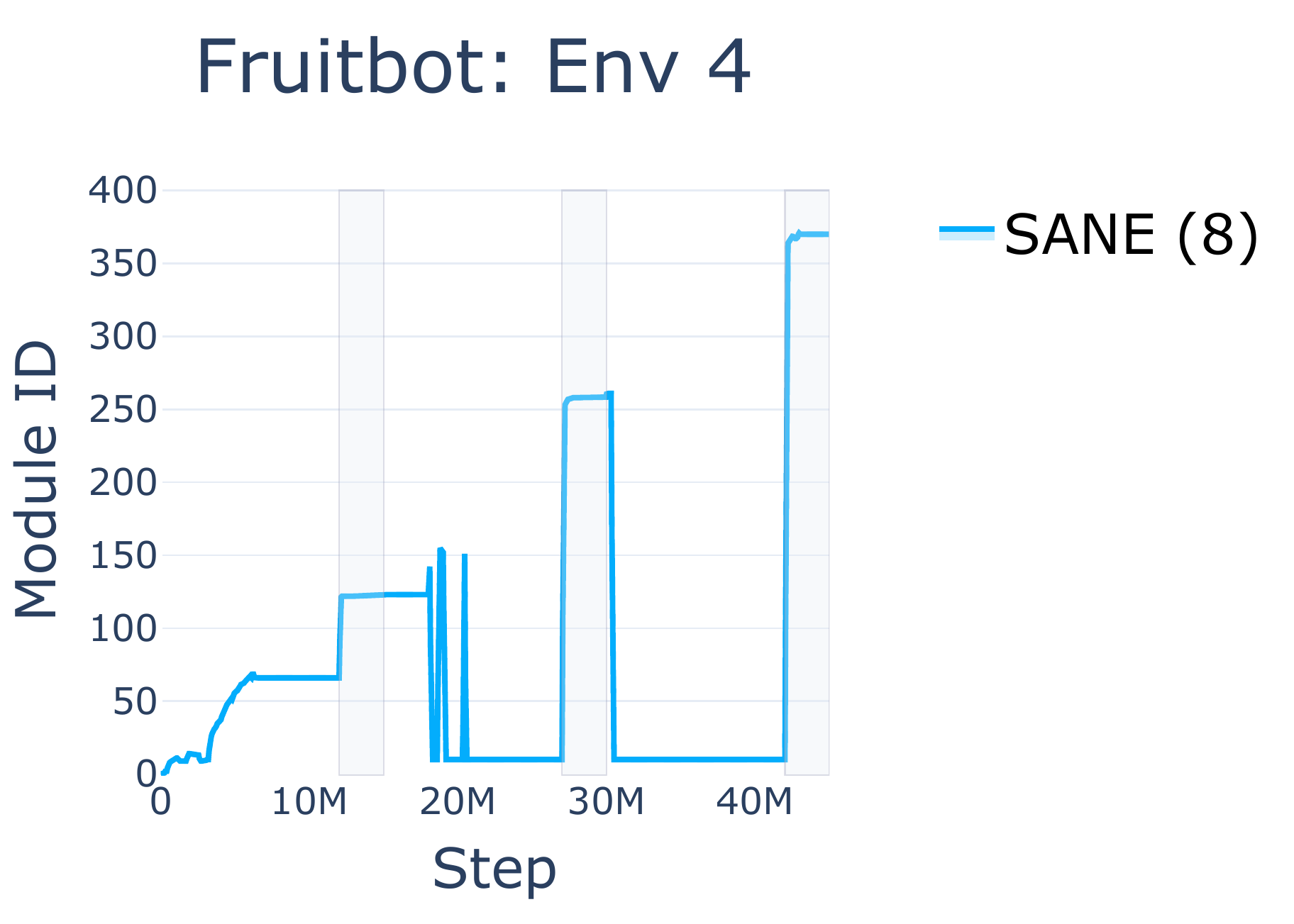}};
        \draw (-0.05, 0.20) node {\tiny C};
    \end{tikzpicture}\\
    \includegraphics[trim=0 4em 15em 5em, clip,width=0.15 \columnwidth]{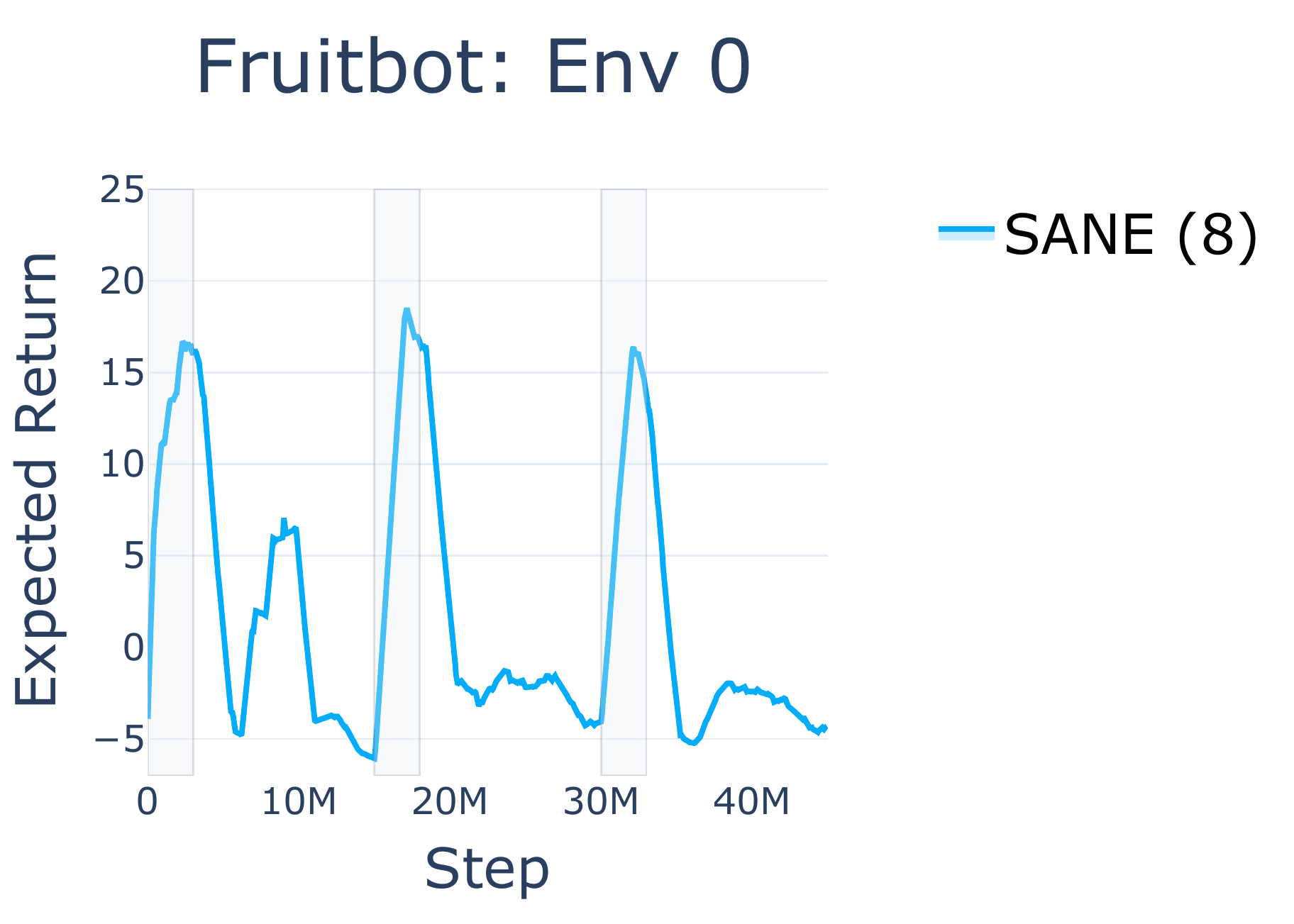}
    \includegraphics[trim=0 4em 15em 5em, clip,width=0.15 \columnwidth]{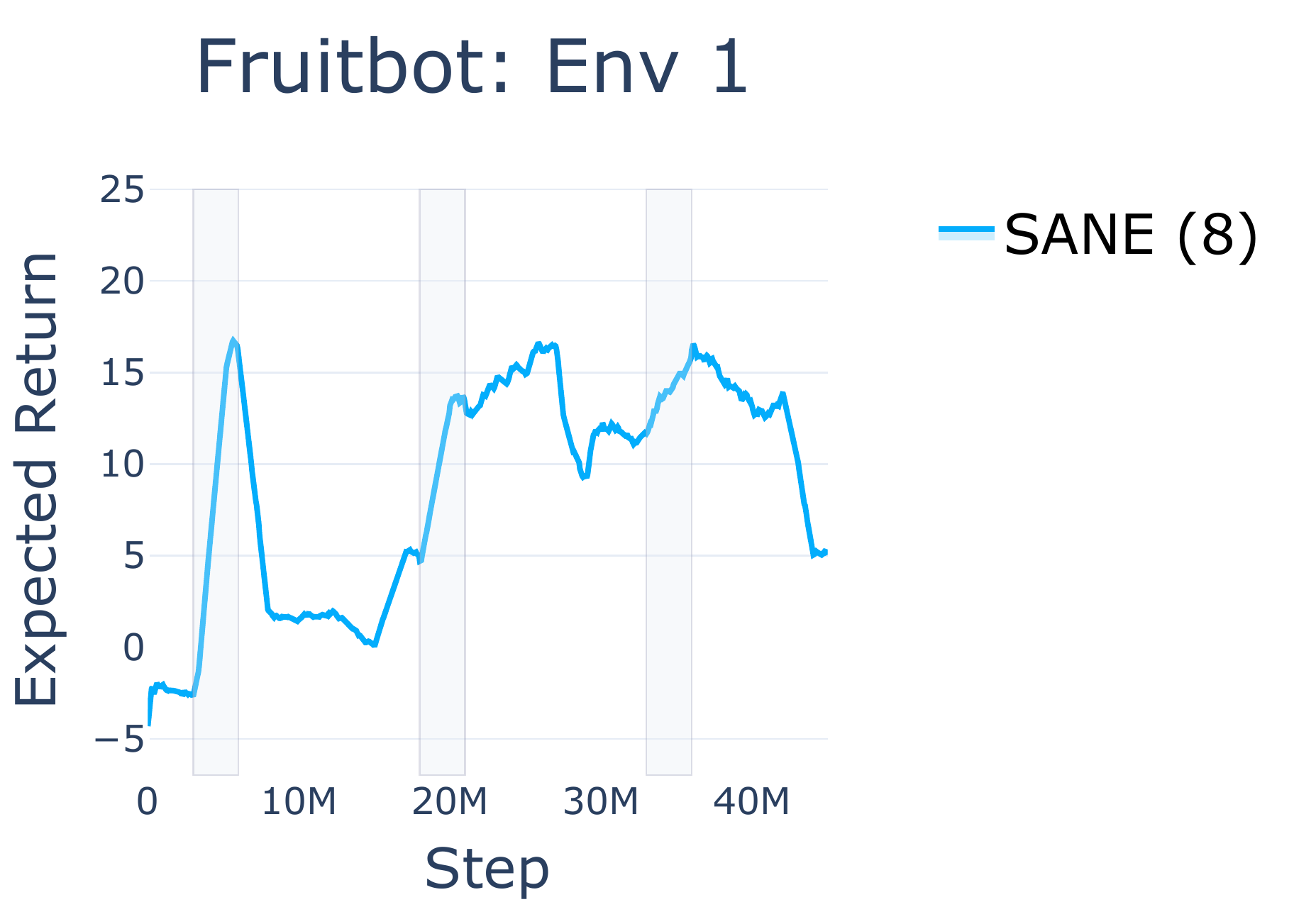}
    \includegraphics[trim=0 4em 15em 5em, clip,width=0.15 \columnwidth]{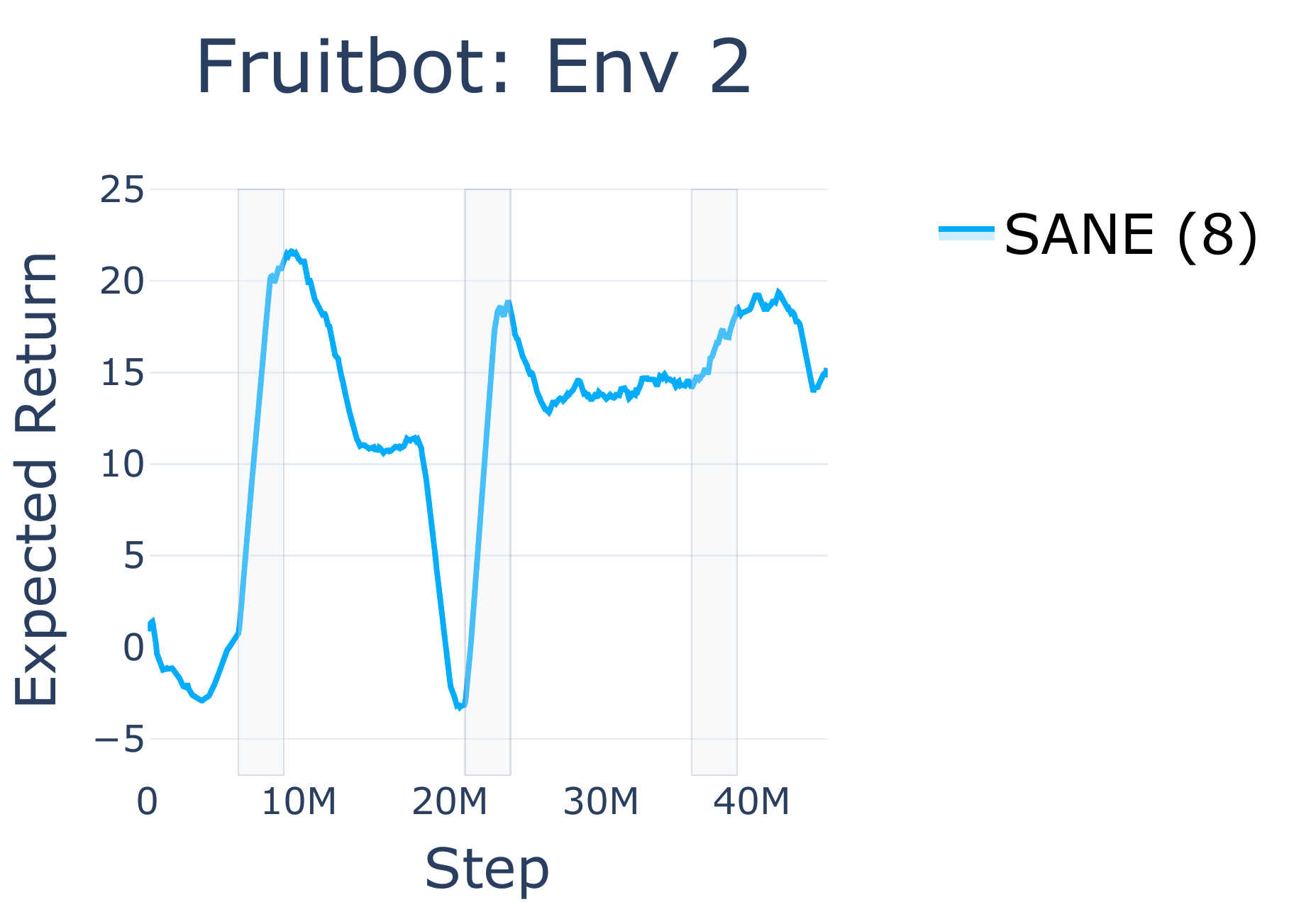}
    \includegraphics[trim=0 4em 15em 5em, clip,width=0.15 \columnwidth]{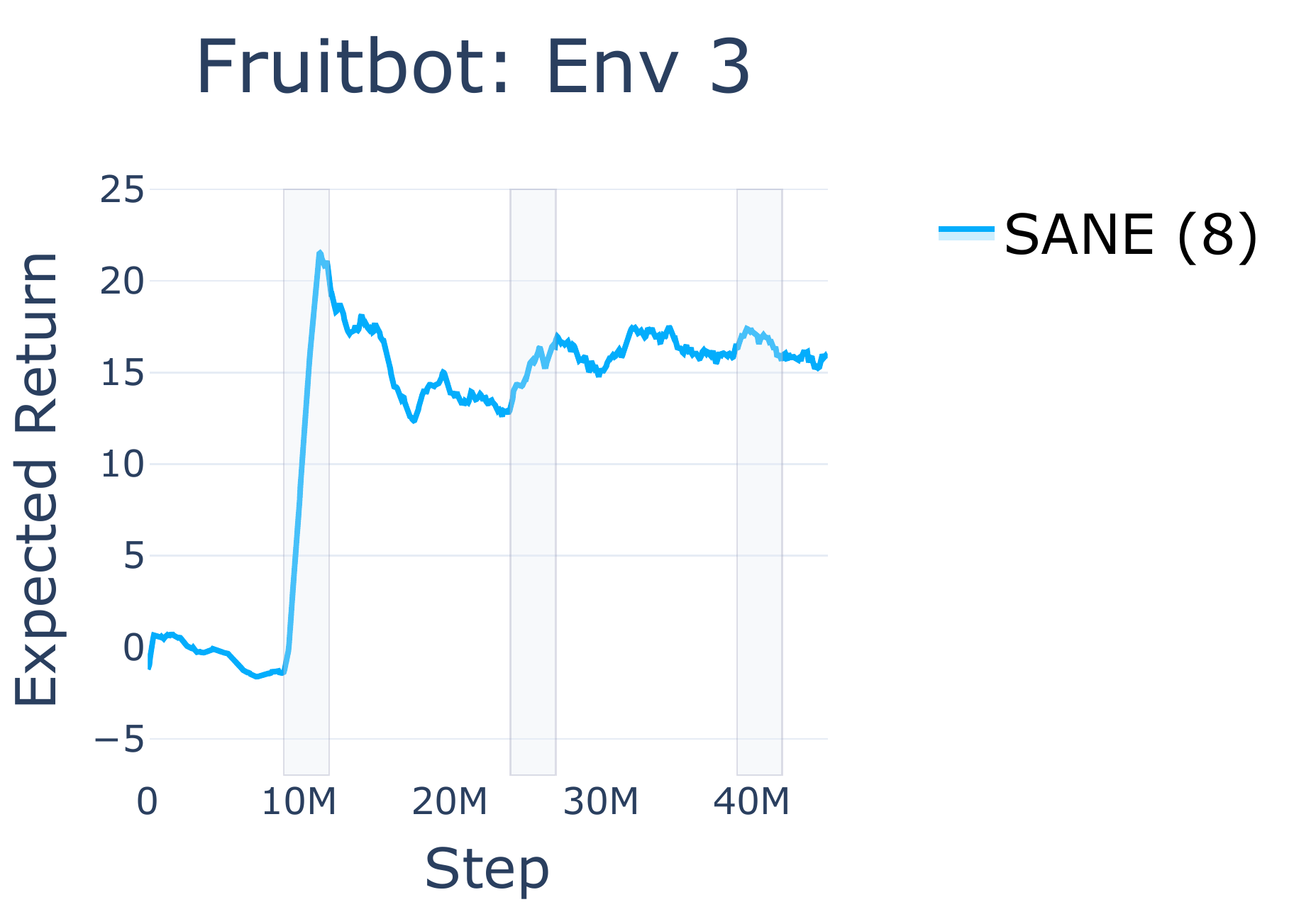}
    \includegraphics[trim=0 4em 0em 5em, clip,width=0.21 \columnwidth]{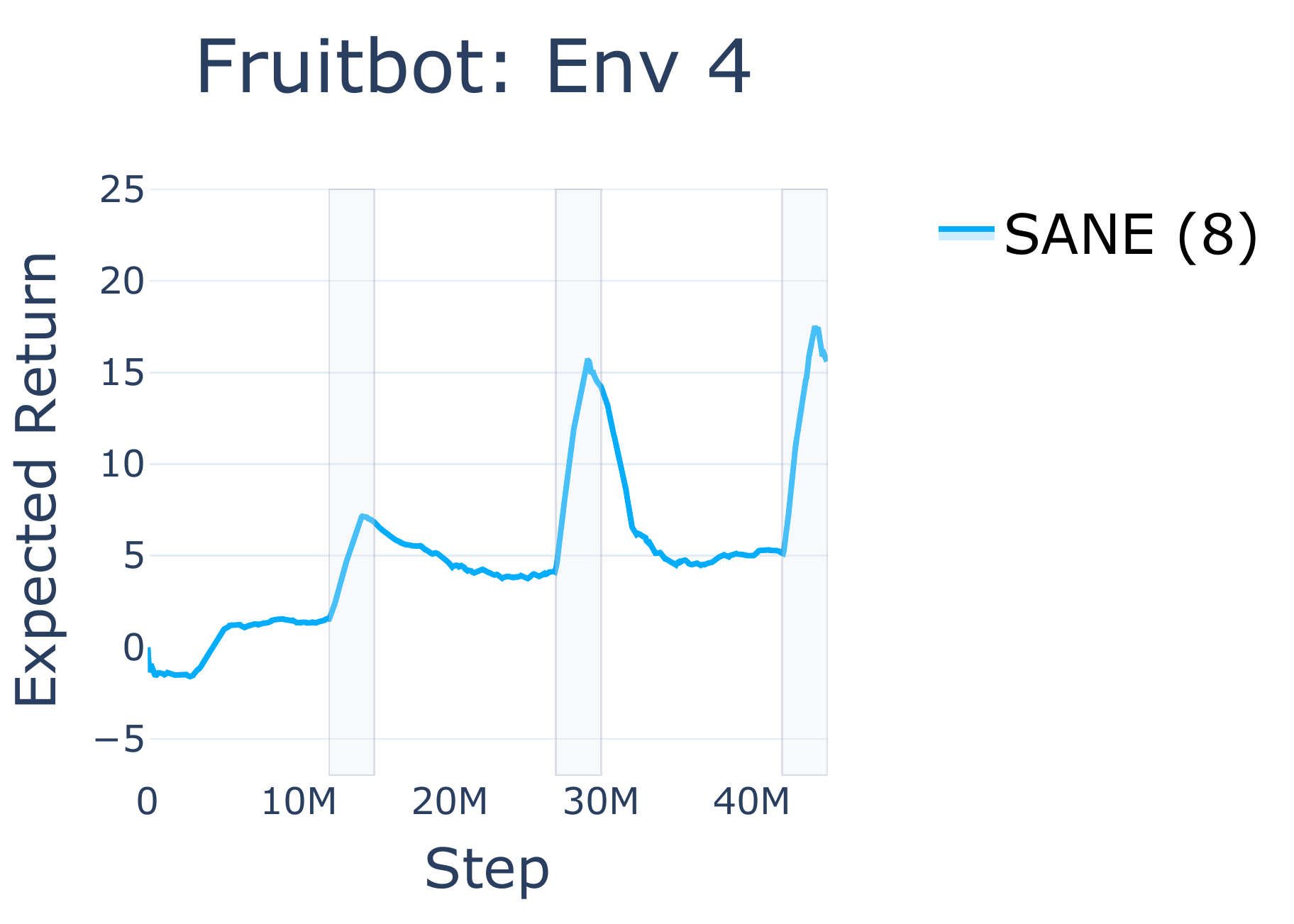}
    \caption{The Module ID and Expected Return plots for a single run of Fruitbot, aligned by timestep to see how modules are getting used and created while Fruitbot is training.}
    \label{fig:fruitbot_single}
\end{figure}
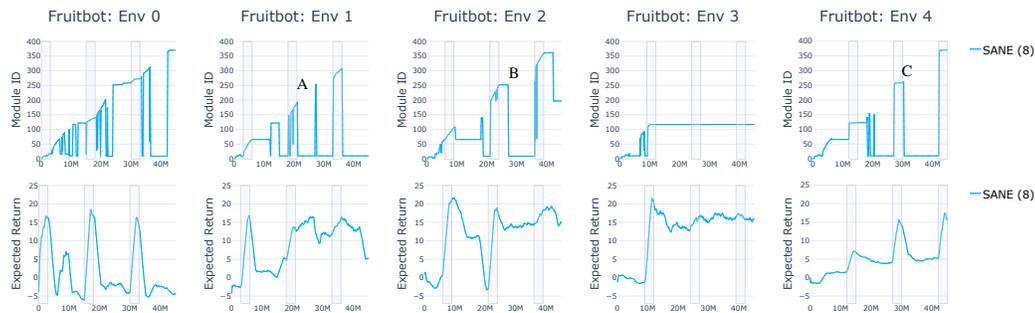

Here we analyze a single run of Fruitbot, which allows us to see in more detail the dynamics of SANE. We use an ensemble with 8 modules to simplify analysis. We focus on three important points, labeled A, B, and C in Figure \ref{fig:fruitbot_single}. In all three cases the module the Environment is using switches to an older module. 

At A (21M timesteps), Env 1 switches from using Module 193 to Module 10. By analyzing the Lineage plot (not shown here due to its large size) we see that 193 merged into Module 150, which then merged into Module 10. Thus the continued high performance on this task can be explained by a successful sequence of merges.

At B (27M timesteps), Env 2 switches from using Module 252 to Module 9. In this case, Module 252, which is a direct descendent of Module 9, merged into Module 197. Module 197 is at this point still a module available in the ensemble, meaning Env 2 began to activate a high-performing previous module (Module 9) instead of the result of the merge, implying that the critic value of 252 decayed as a result of its merger into Module 197. However, performance was rescued by return to a previous module and performance remains high.

At C (30M timesteps), Env 4 switches from using Module 261 to Module 10. Module 10, as we saw in case A, is a module that is well-suited for Env 1. In this case, Module 261 merged into Module 262, a descendent of Module 9, which as we saw in case B is well-suited for Env 2. Essentially, our 8 module ensemble lacks the capacity to adequately represent all of the behaviors necessary for this sequence of tasks, and start combining policies destructively, resulting in forgetting. This is mitigated by introducing more modules into our ensemble, as shown in Figure \ref{fig:sane_ablations}

\section{Conclusion}

Inspired by the fact that catastrophic forgetting is caused by updating all neurons in a network for all tasks, we propose the creation of self-activating modules to break up a network into modular components that only get updated when they are used. Our experimental results suggest that a dynamic ensemble, which creates modules as necessary and merges them to conserve resources, performs better than a static ensemble where all modules are created up-front. By combining these two novel features, we present SANE (Self-Activating Neural Ensembles) for continual reinforcement learning. We demonstrate SANE on sequences of Procgen levels that prove particularly challenging for the current state-of-the-art (CLEAR), showing that our method reliably improves the mitigation of catastrophic forgetting. Furthermore, we present a thorough analysis of SANE, showing how modules are created, used, and merged on individual runs of Climber and Fruitbot, to provide a more comprehensive view into the system.

\textbf{Limitations and future directions} In this paper, we present a straightforward and simple instantiation of SANE, which has some limitations. First, using the initial observation of the episode to choose which module to activate limits the current method to tasks that are distinguishable immediately. Second, the complete separation of modules precludes transfer, wherein improvement on one task benefits performance on another. Future work to address these issues may include choosing the active module every $n$ steps instead of once at the beginning of an episode, or making a hierarchical version of SANE where similar tasks activate similar paths through the tree while distinct tasks activate non-overlapping paths.

\clearpage

\section*{Acknowledgements}
This work was supported by ONR MURI, ONR Young Investigator Program, and DARPA MCS.

\bibliography{refs_new}
\bibliographystyle{collas2022_conference}

\appendix
\clearpage

\section{Appendix}

\subsection{Hyperparameters}
\label{section:hyperparams}

Here we give the hyperparameters for our methods; see the provided code for more details. For convenience, we put all parameters that vary between methods above the line, and those that are consistent below. 

\setlength\tabcolsep{1.5pt} 
\begin{table}[H]
    \small
    \centering
    \begin{tabular}{lcccccc}
        \toprule
        Hyperparameter &&&&&& Shared\\
        \midrule
        Num. actors &&&&&& 16 \\
        Learner threads &&&&&&  1 \\
        Unroll length &&&&&&  32 \\
        Grad clip &&&&&& 40 \\
        Reward clip &&&&&& $[-1, 1]$ \\
        Normalize rewards &&&&&& No \\
        Entropy cost &&&&&& 0.01 \\
        Discount factor &&&&&& 0.99 \\
        LSTM &&&&&& No \\
        Network arch. &&&&&& Nature CNN \\
        Learning rate &&&&&& 4e${-4}$ \\ 
        Optimizer &&&&&& RMSProp \\
        &&&&&& $\alpha$ = 0.99 \\
        &&&&&& $\epsilon$ = 0.01 \\
        &&&&&& $\mu$ = 0 \\
        \bottomrule
        \toprule
        & (Climber \& Miner) & (Fruitbot) & & & & \\
        Hyperparameter & SANE & SANE & CLEAR & P\&C & EWC & \\
        \midrule
        Batch size & 2 & 2 & 2 & 18 & 2 & \\
        Baseline cost & 5.0 & 0.5 & 0.5 & 0.5 & 0.5 & \\
        EWC $\lambda$ & & & & 30 & 300 & \\
        EWC, min task steps & & & & & 2e5 & \\
        Fisher samples & & & & 100 & 100 & \\
        Normalize Fisher & & & & Yes & No & \\
        Online EWC $\gamma$ & & & & 0.99 & \\
        KL cost & & & & 1.0 & & \\
        Policy cloning cost & & & 0.1 & & \\
        Value cloning cost & & & 0.005 & & \\
        Replay ratio & 8 & 8 & 8 & & \\
        Replay buffer size & 50k & 12.5k & 400k & & \\
        & (per module) & (per module) & & & \\
        \midrule
        Max num. modules & 8 & 32 & & & \\
        $\alpha_{u, inf}$ & 1.0 & 1.0 & & & \\
        $\alpha_{u, create}$ & 0.1 & 0.1 & & & \\
        $\alpha_{l, create}$ & 10.0 & 0.5 & & & \\
        Critic cadence $T$ & 1000 & 10000 & & & \\
        Critic target update rate $\tau_V$ & 0.9 & 0.9 & & & \\
        Uncertainty cost $\mu$ & 1.0 & 0.1 & & & \\
        \bottomrule
    \end{tabular}
    \caption{Hyperparameters for all methods. The network architecture is the ``Nature CNN'' model~\cite{mnih2015human}.}
    \label{tab:climb_miner_params}
\end{table}


Note that there are two different $\alpha$ parameters for SANE. One describes the parameter used for inference ($\alpha_{u, inf}$) and the other describes the parameters used during module creation ($\alpha_{x, create}$).

\begin{figure}[H]
    \centering
    \includegraphics[trim=0 0em 22em 0, clip,width=0.19 \textwidth]{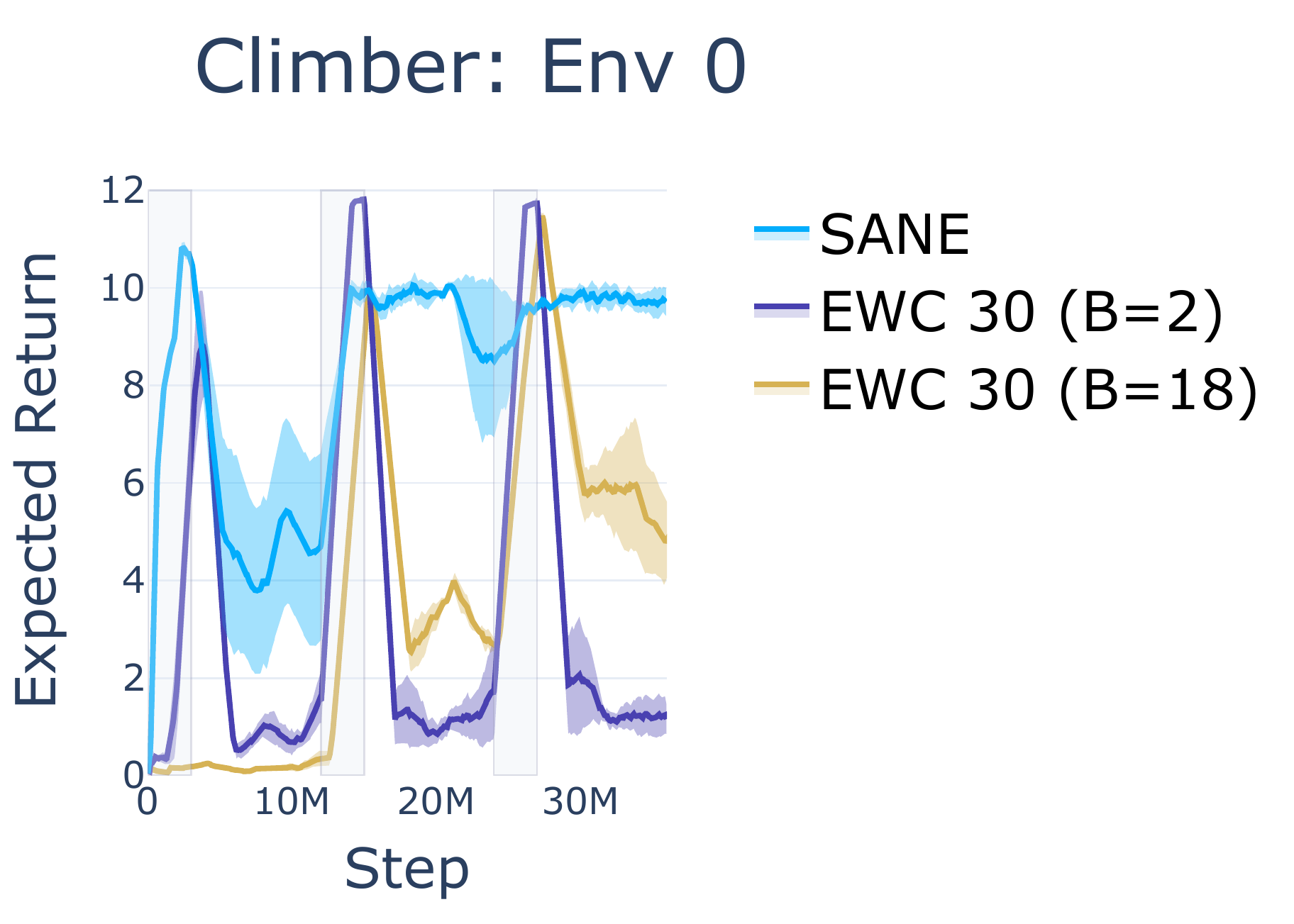}
    \includegraphics[trim=0 0em 22em 0, clip,width=0.19 \textwidth]{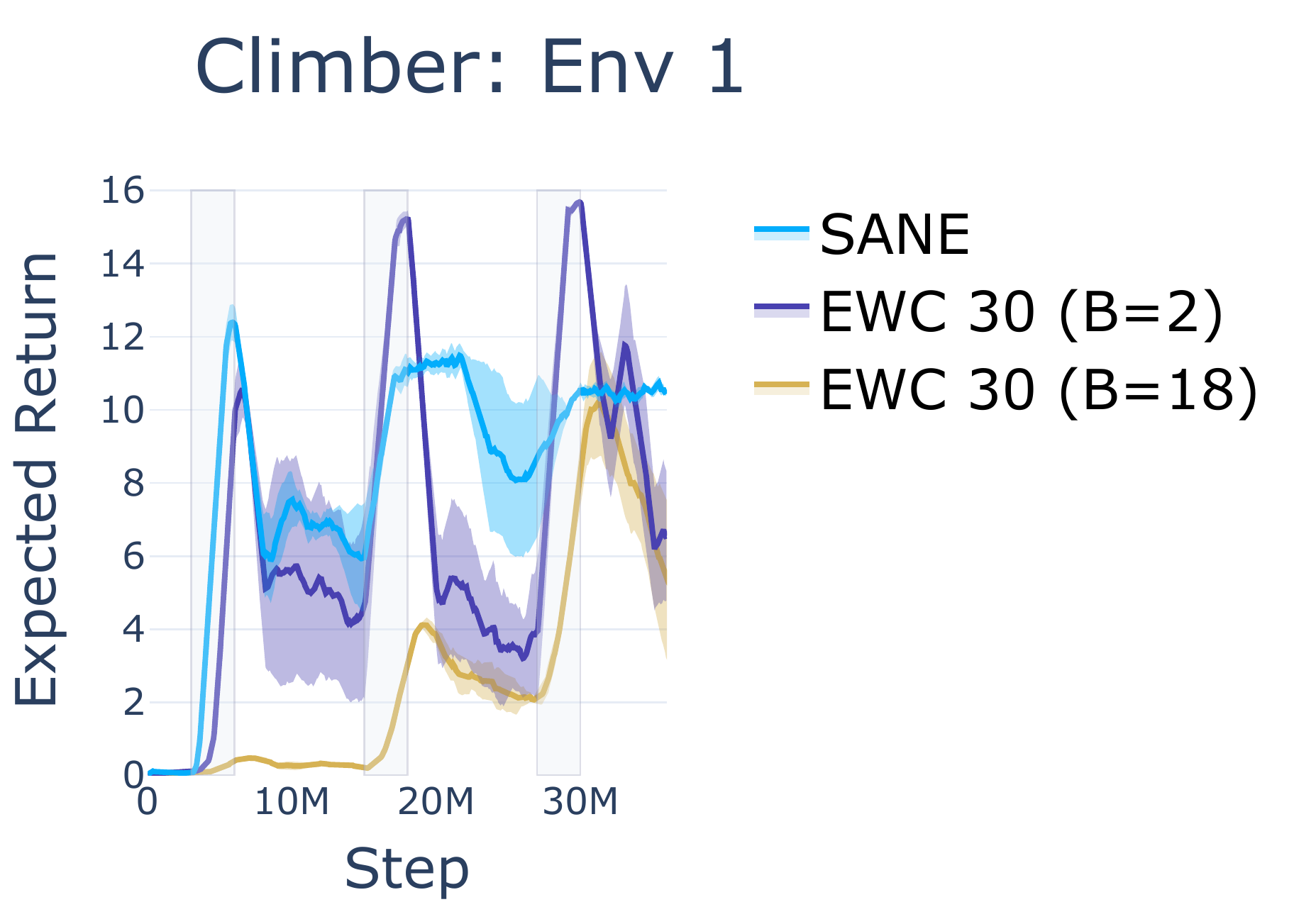}
    \includegraphics[trim=0 0em 22em 0, clip,width=0.19 \textwidth]{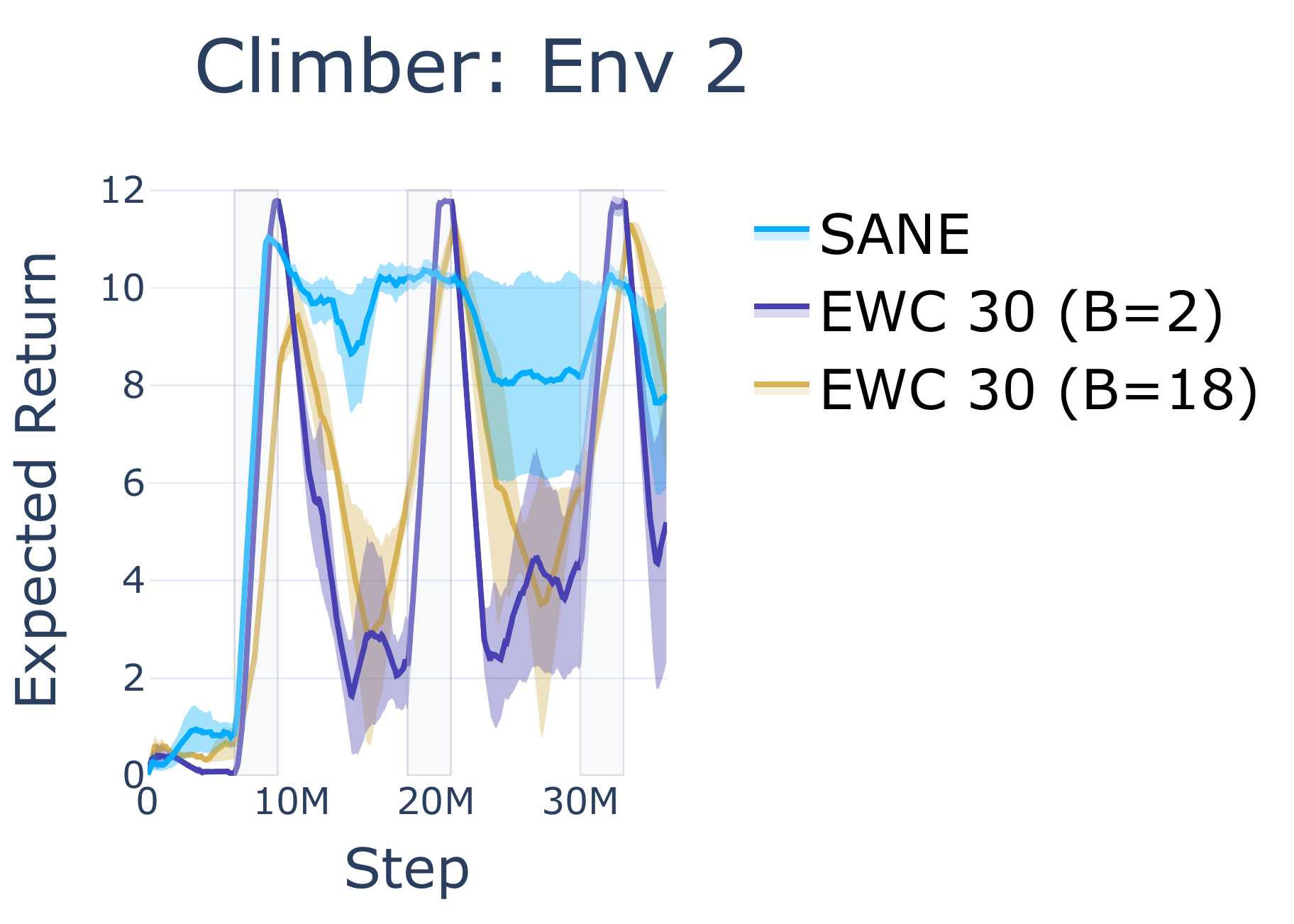}
    \includegraphics[width=0.325 \textwidth]{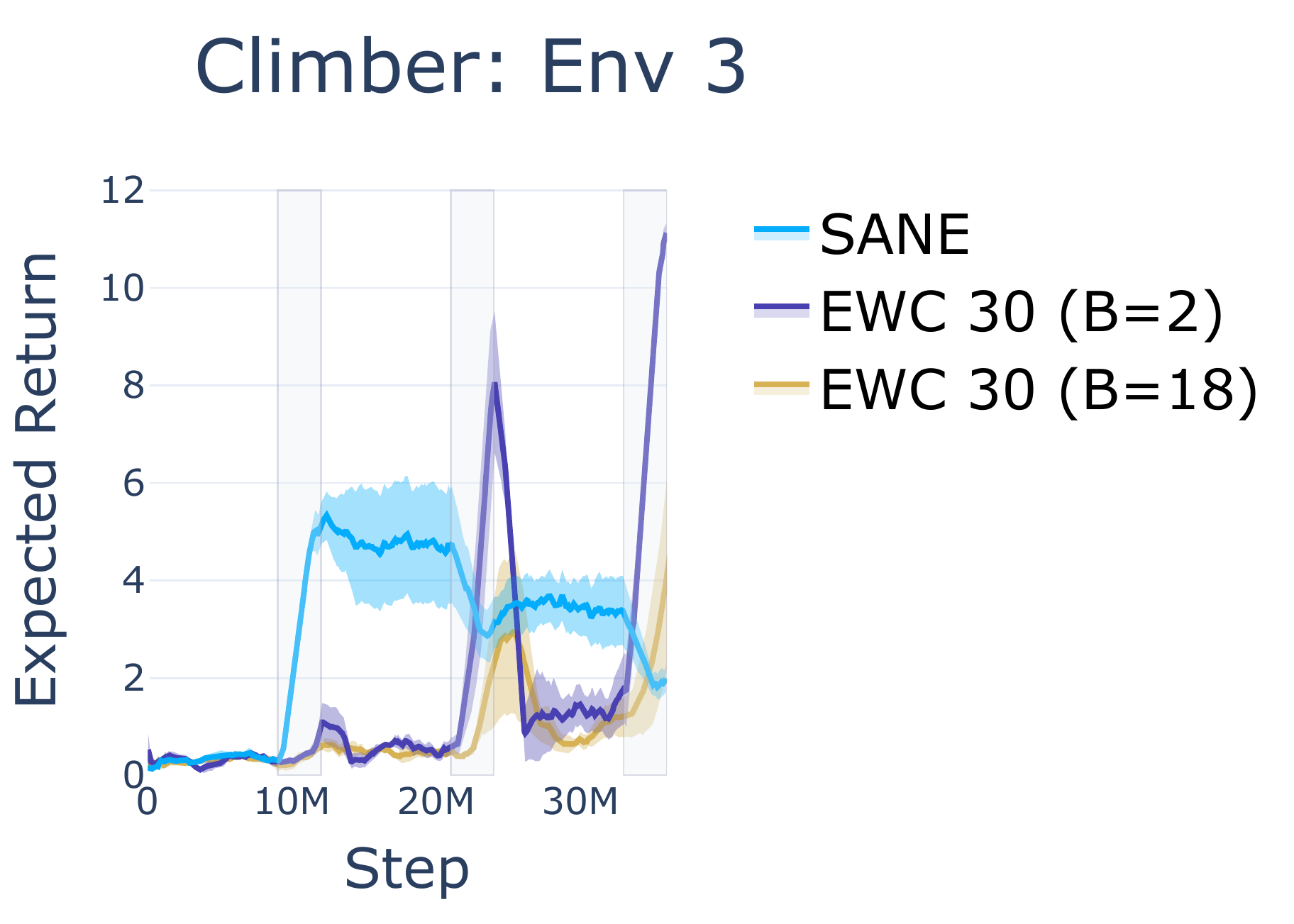}
    \caption{Comparison of running EWC with a batch size of 2 versus the default (18).}
    \label{fig:ewc_tuning_2batch}
\end{figure}

\noindent {\bf Batch size } Since two of the methods augment the batch with more data (SANE and CLEAR), there is the question of how to size the batches of the other baselines (EWC, P\&C) fairly. There are two ways to view it. The first is to equalize the total amount of data the optimizer sees per gradient step (i.e. since the augmented size is 18, collect a batch size of 18 for EWC and P\&C). The second is to equalize the amount of new data the optimizer sees (i.e. 2 new trajectories are collected so use a batch size of 2 for EWC and P\&C). We compare the differences in Figure \ref{fig:ewc_tuning_2batch}. We can see that a smaller batch (i.e. training more often) results in learning the tasks more quickly, but the impact on recall is more ambiguous: on Env 0 the smaller batch size recalls less well, but it's comparable on all other environments. Worth noting also is that the smaller batch size takes considerably longer to train. Given these facts we have chosen to use the larger batch size for our EWC and P\&C baselines.

\noindent {\bf Increasing network size} The base architecture used for all methods is the Nature CNN model~\citep{mnih2015human} with $\sim$6e5 parameters. To make the 8x version, we multiplied the number of filters at every convolutional layer by 6 (for a total of $\sim$5.7e6 parameters). We opted for making the network wider instead of deeper because: (i) conceptually this is more similar to the structure of the SANE ensemble (ii) it does not introduce the possibility of decreased performance due to gradient vanishing or exploding~\cite{srivastava2015_verydeep}.

\subsubsection{Architecture Overview}

Here we describe how SANE utilizes the highly parallel IMPALA method. Every SANE module is one instance of an IMPALA agent; all modules operate independently, with no shared parameters or data. Each IMPALA agent is composed of a set of actors that are constantly collecting data and populating a shared buffer with the results. When a separate learner thread detects that the minimum batch size of trajectories has been collected, it computes the losses and runs a gradient step. Not needing to pause the actors to update the model provides significant runtime improvements.

SANE augments this basic structure in one primary way. In order to switch modules, the currently running module is paused (all actors stop collecting data, and all learner threads stop updating the model) every $s_{yield}$ seconds. At this point an activation score is computed for every module and the highest-scoring module is activated. Activation means that all actors are restarted, and model updating resumes. This alternating of module activation and data collection continues until all tasks are complete.

\subsubsection{Hyperparameter Tuning}

\begin{figure}[H]
    \centering
    \includegraphics[trim=0 0em 18em 0, clip,width=0.19 \textwidth]{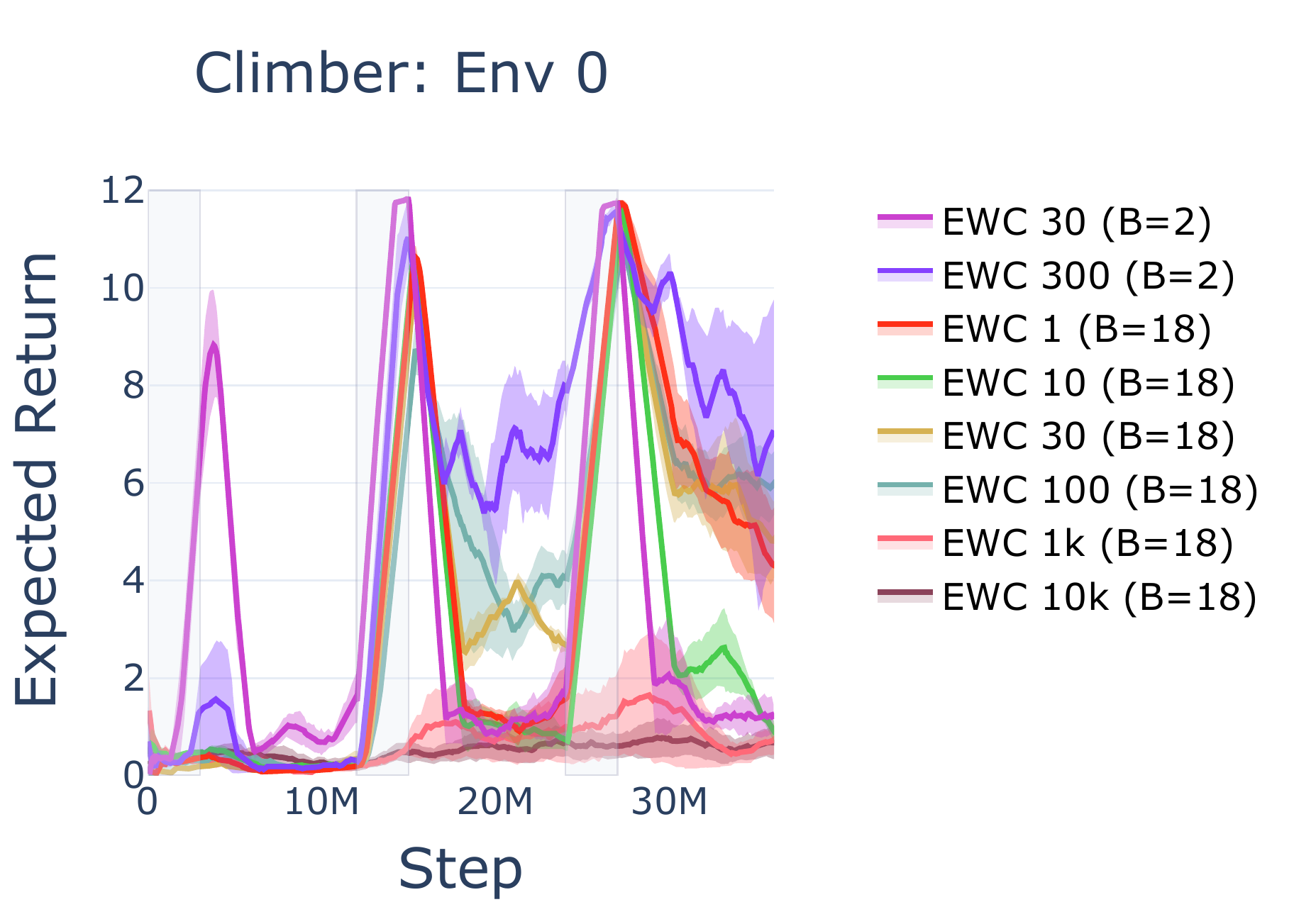}
    \includegraphics[trim=0 0em 18em 0, clip,width=0.19 \textwidth]{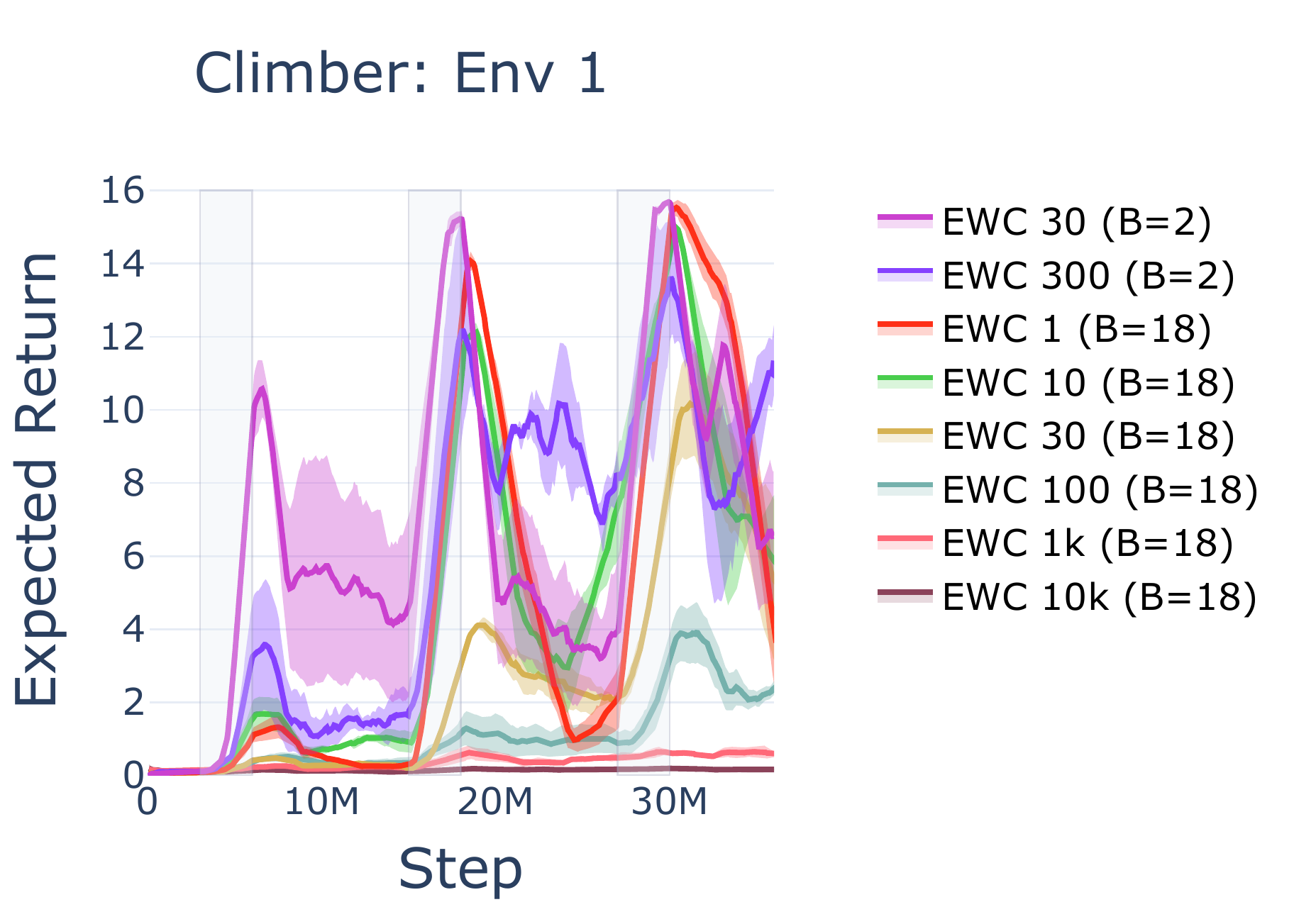}
    \includegraphics[trim=0 0em 18em 0, clip,width=0.19 \textwidth]{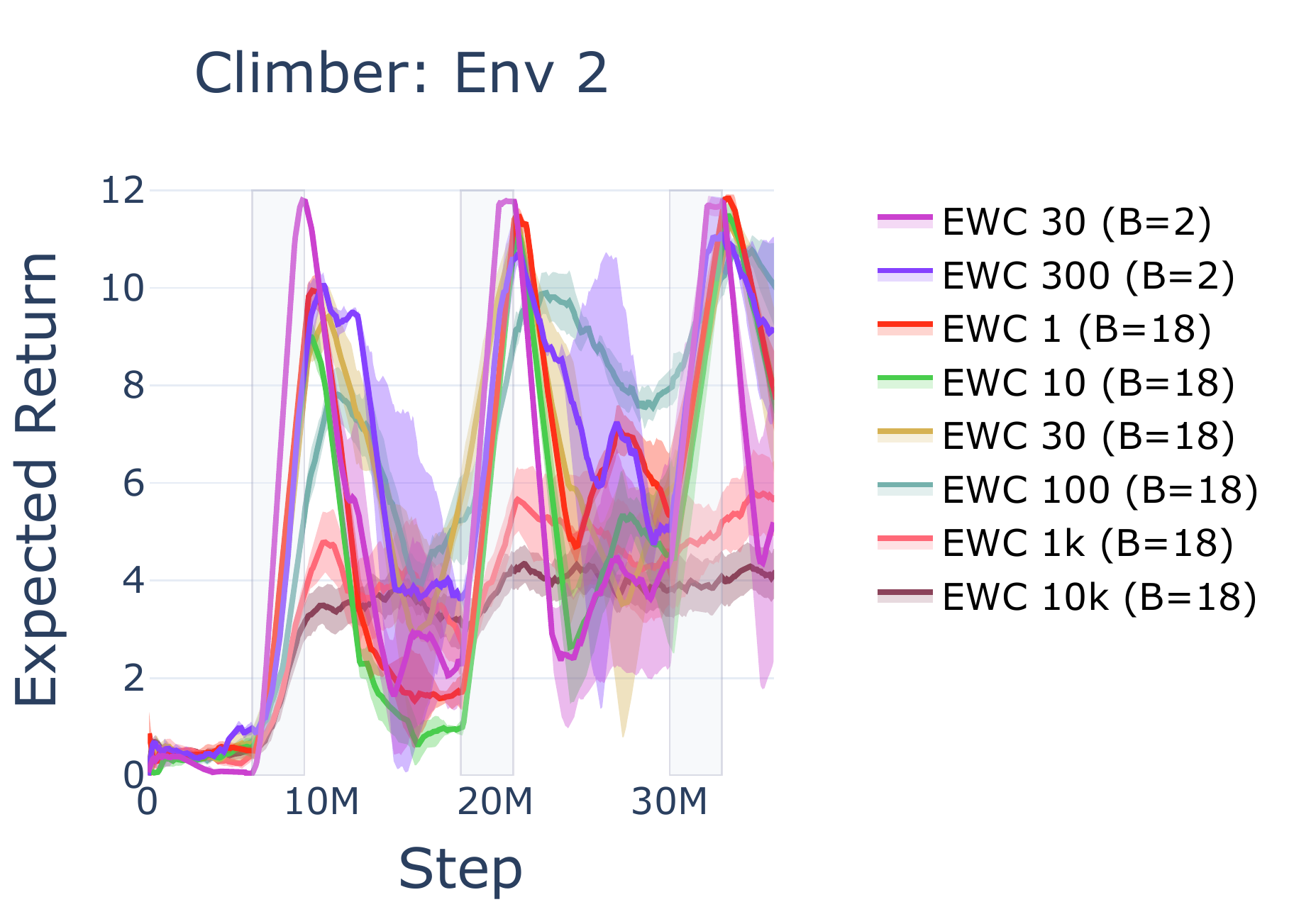}
    \includegraphics[width=0.29 \textwidth]{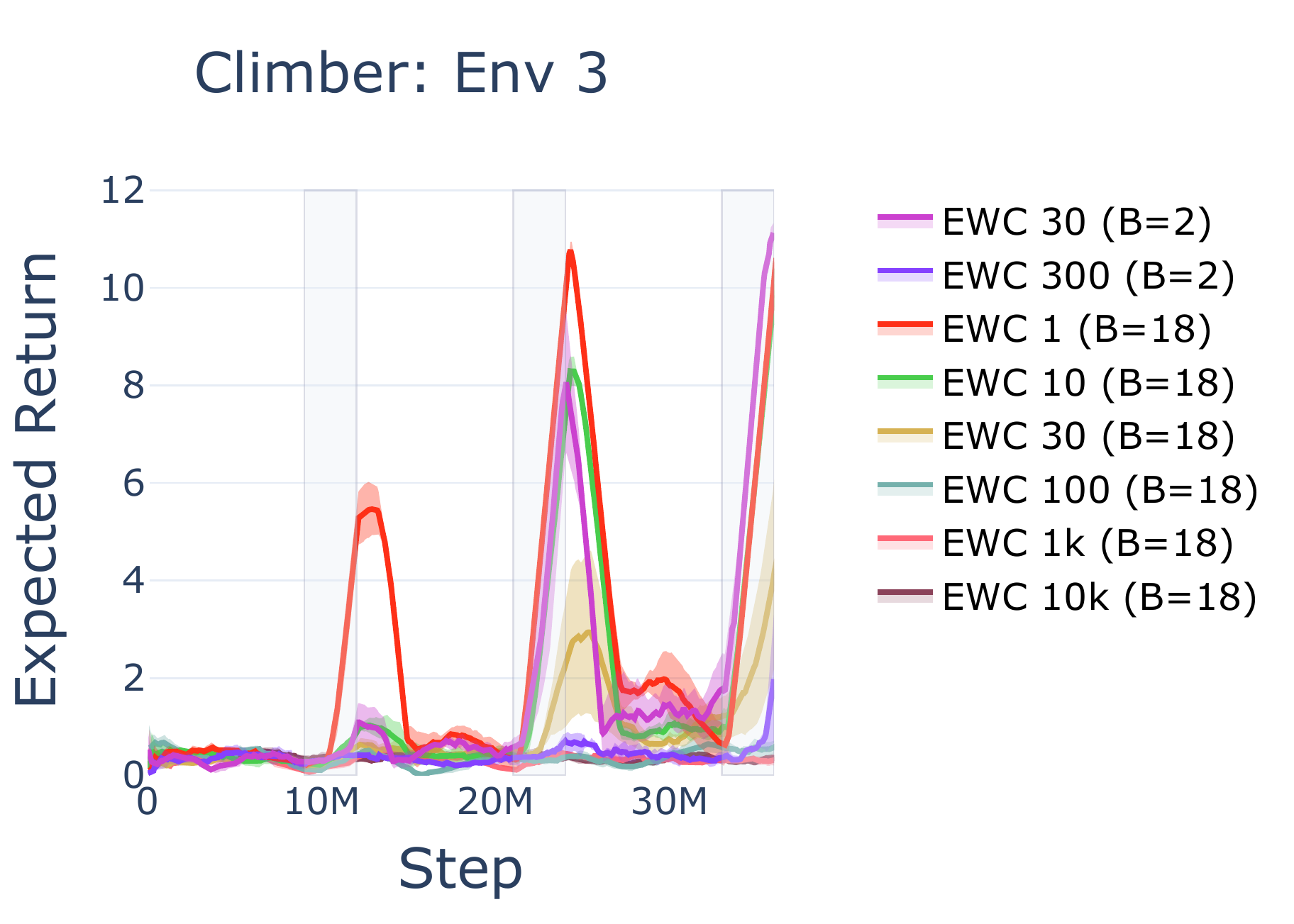}
    \caption{Comparison of EWC $\lambda$ variations on the Climber task sequence. The number represents $\lambda$.}
    \label{fig:ewc_tuning}
\end{figure}

First, we present the $\lambda$ tuning graph for EWC in Figure \ref{fig:ewc_tuning}. We ran using $\lambda$ values in the range [1, 1e5] with a batch size of 18 (B=18) on the Climber task sequence with 3 seeds. We also ran two experiments with (B=2).  We can see that while the lower $\lambda$ values (1 and 10) learn the tasks better, they are also more inclined to forget, as can be seen particularly on Envs 0 and the first cycle of Env 2. Based on these results we chose $\lambda=300$ and a batch size of 2 for the experiments presented in the paper as the best balance of learning and remembering.

\begin{figure}[H]
    \centering
    \includegraphics[trim=0 0em 18em 0, clip,width=0.19 \textwidth]{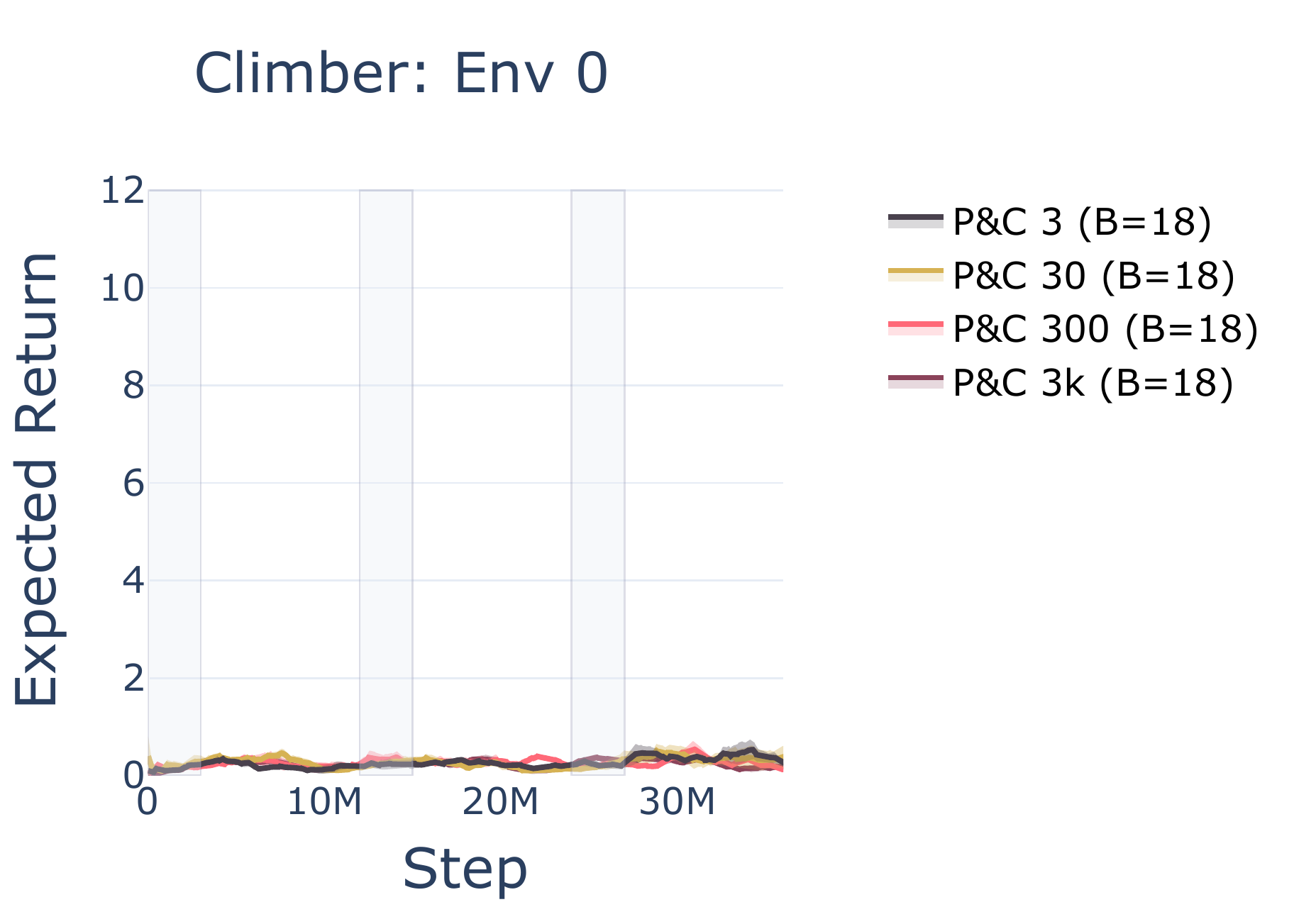}
    \includegraphics[trim=0 0em 18em 0, clip,width=0.19 \textwidth]{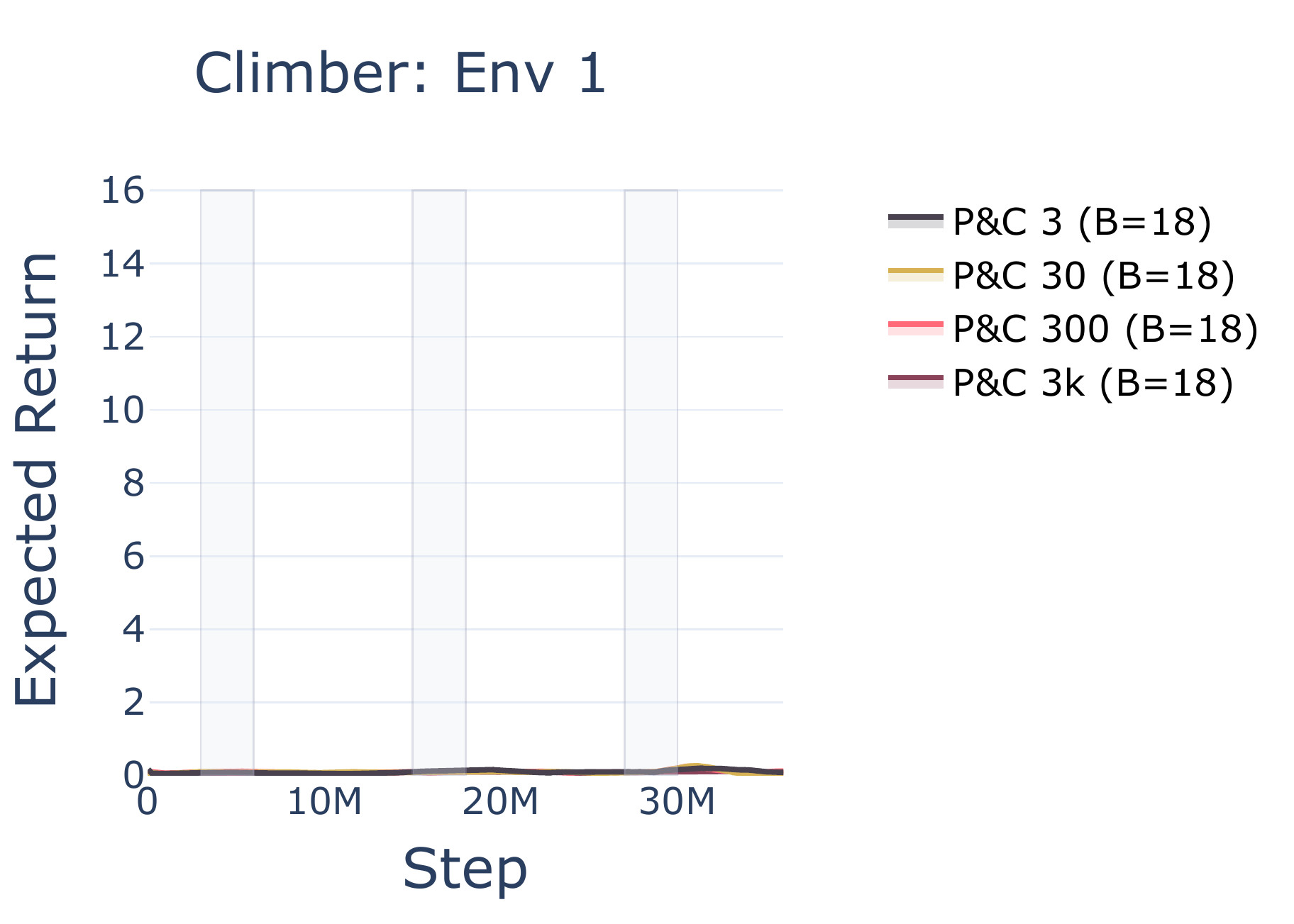}
    \includegraphics[trim=0 0em 18em 0, clip,width=0.19 \textwidth]{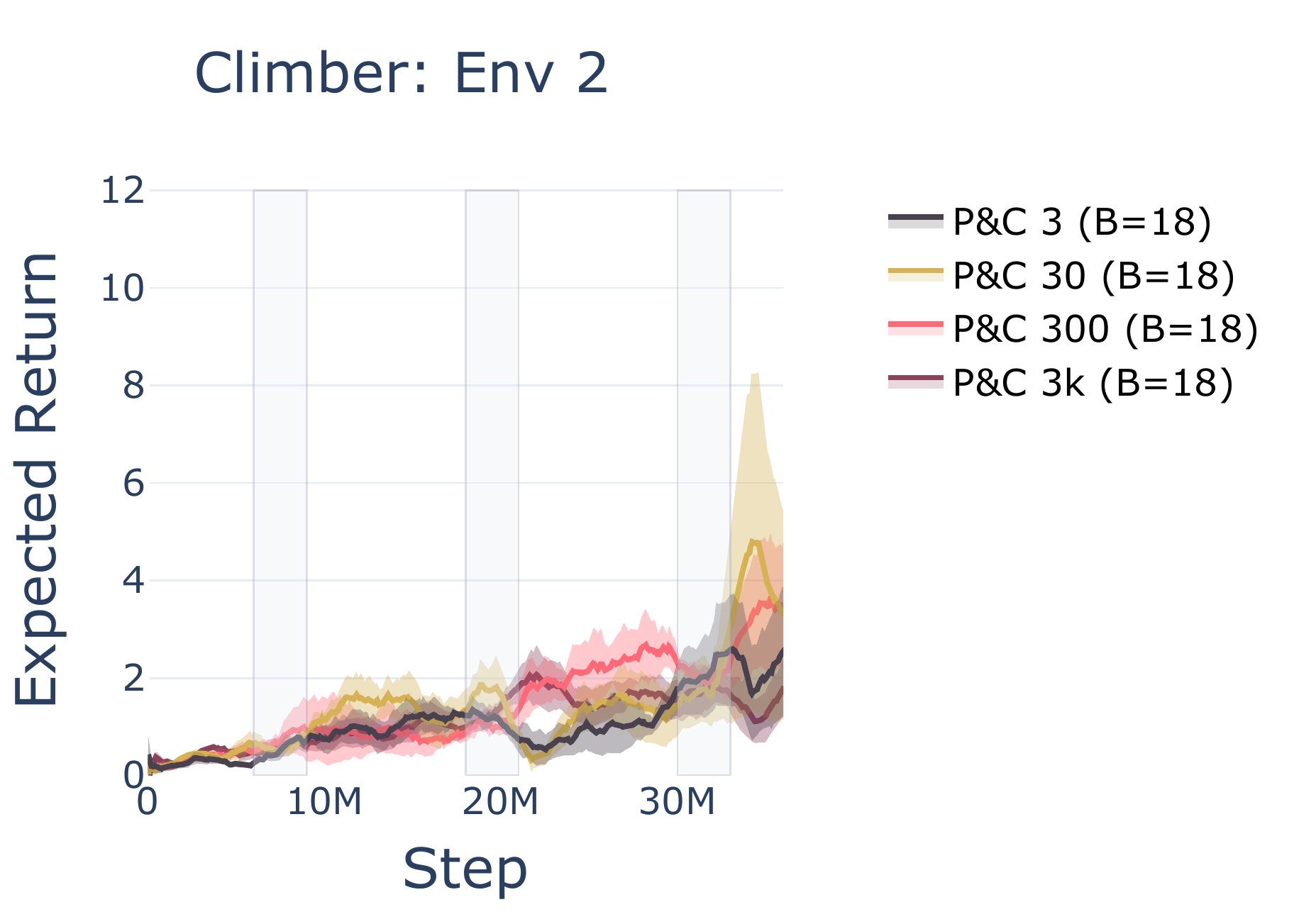}
    \includegraphics[width=0.29 \textwidth]{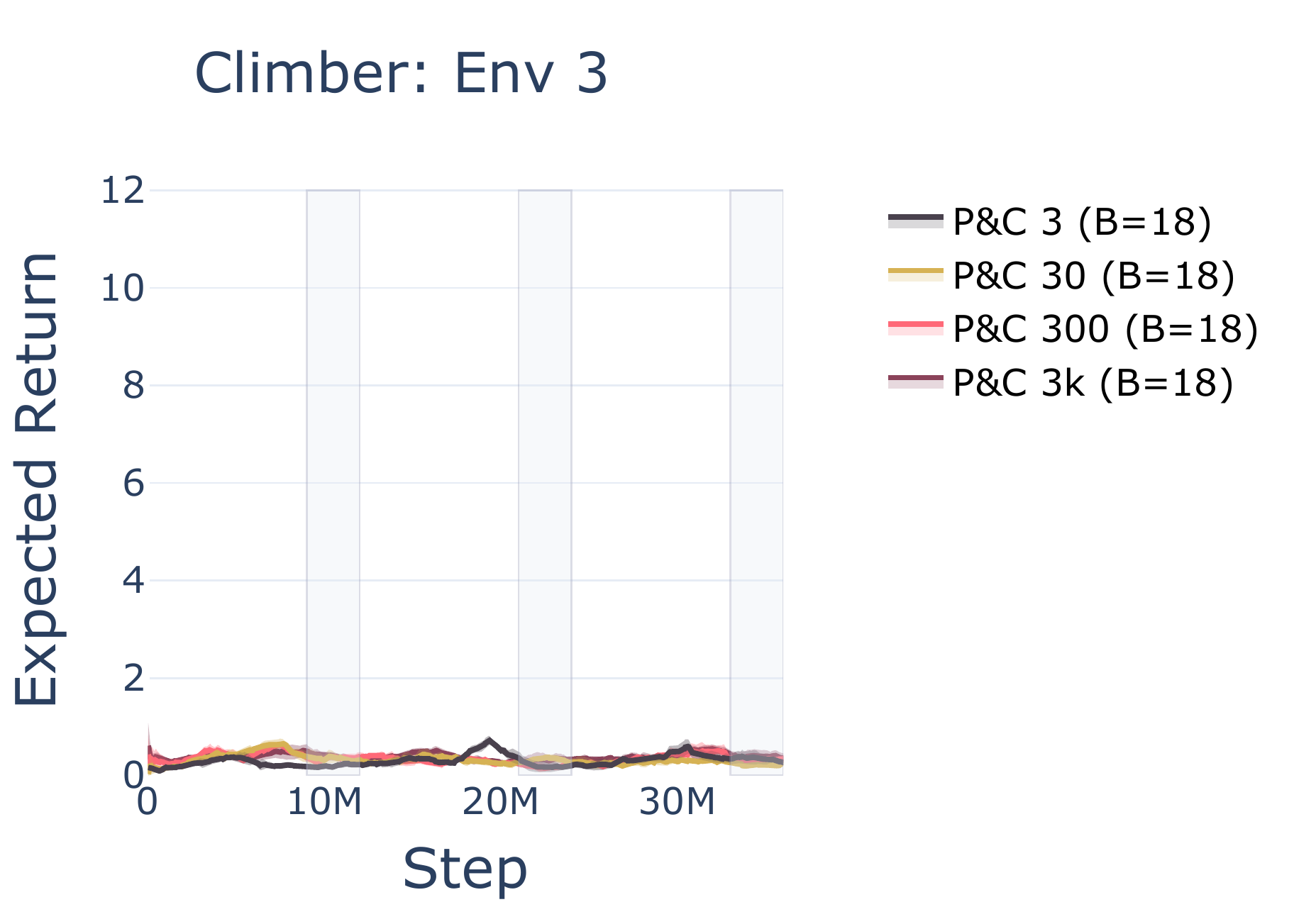}
    \caption{Comparison of P\&C $\lambda$ variations on the Climber task sequence. The number represents $\lambda$.}
    \label{fig:pnc_tuning}
\end{figure}

Second, we present the $\lambda$ tuning graph for P\&C in Figure \ref{fig:pnc_tuning}, representing values in the range [3, 3e3], over 3 seeds for a batch size of 18. We thus chose $\lambda=30$ for our experiments 

\subsection{Parameter Ablation on Fruitbot}
\label{section:fruitbot_params}

As shown in Table \ref{tab:climb_miner_params} we use different parameters for Climber/Miner vs Fruitbot. For simplicity we refer to the former set of parameters as SANE v1 and the latter as SANE v2.

\begin{figure}[H]
    \centering
    \includegraphics[trim=0 0em 20em 0, clip,width=0.15 \columnwidth]{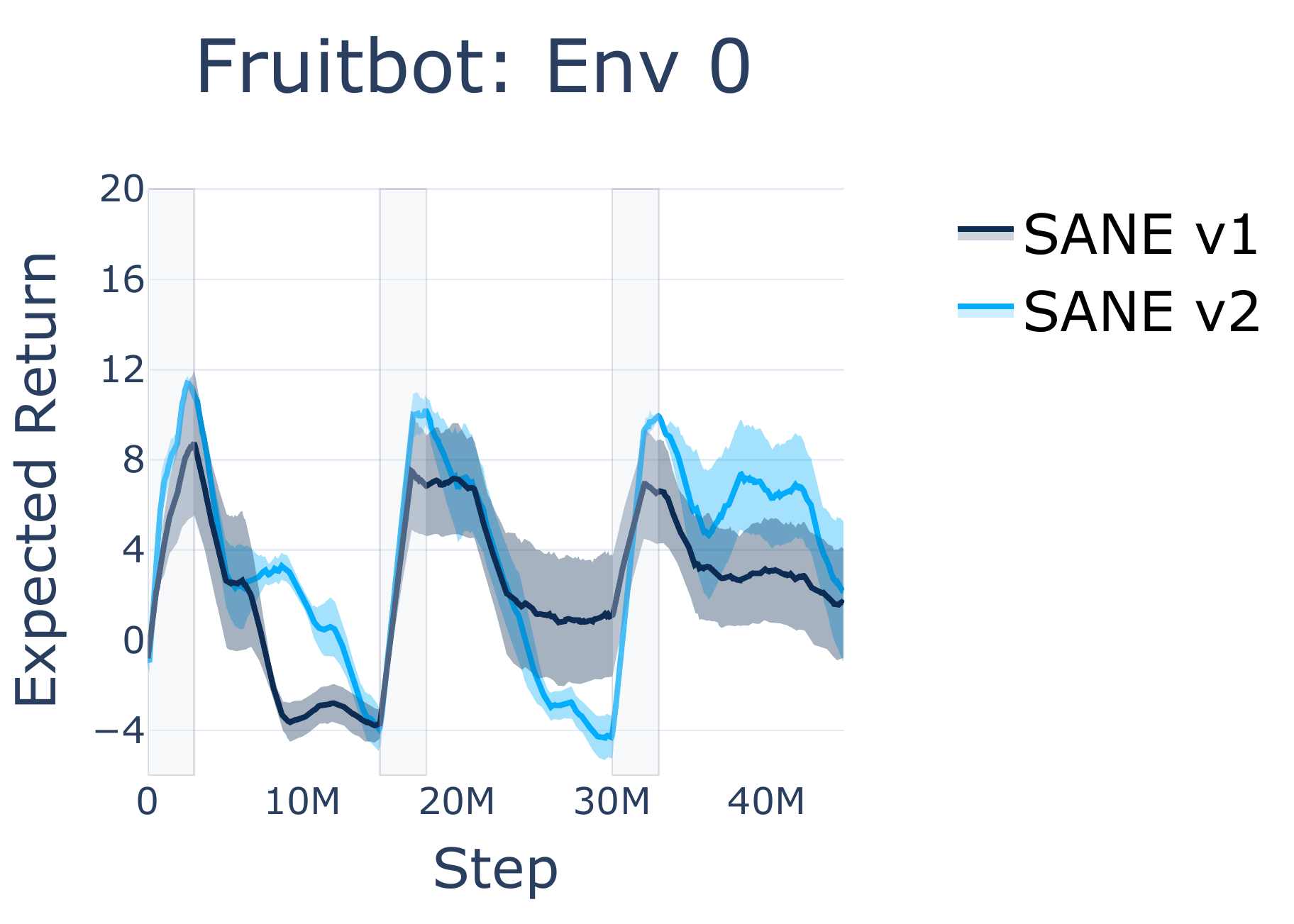}
    \includegraphics[trim=0 0em 20em 0, clip,width=0.15 \columnwidth]{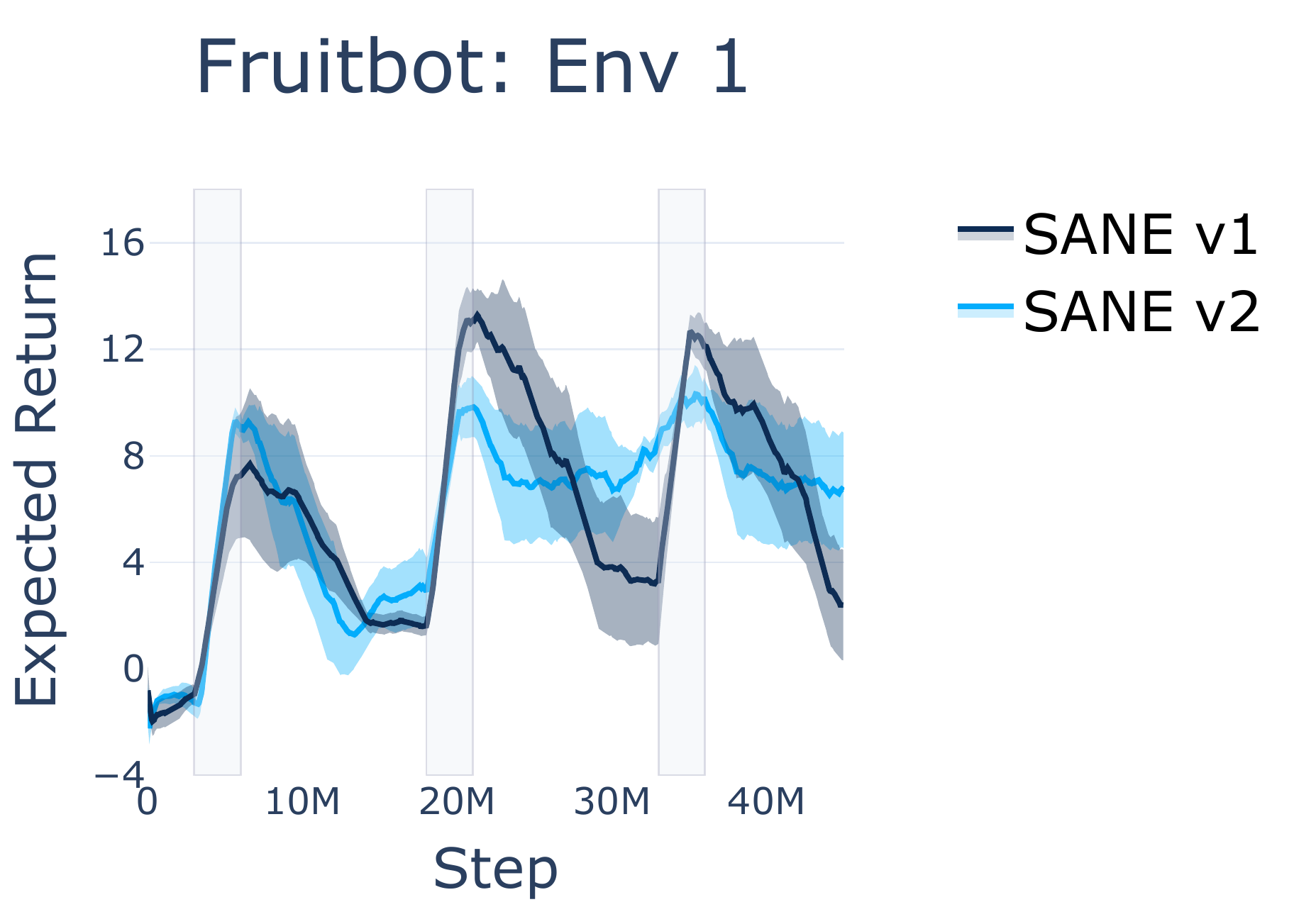}
    \includegraphics[trim=0 0em 20em 0, clip,width=0.15 \columnwidth]{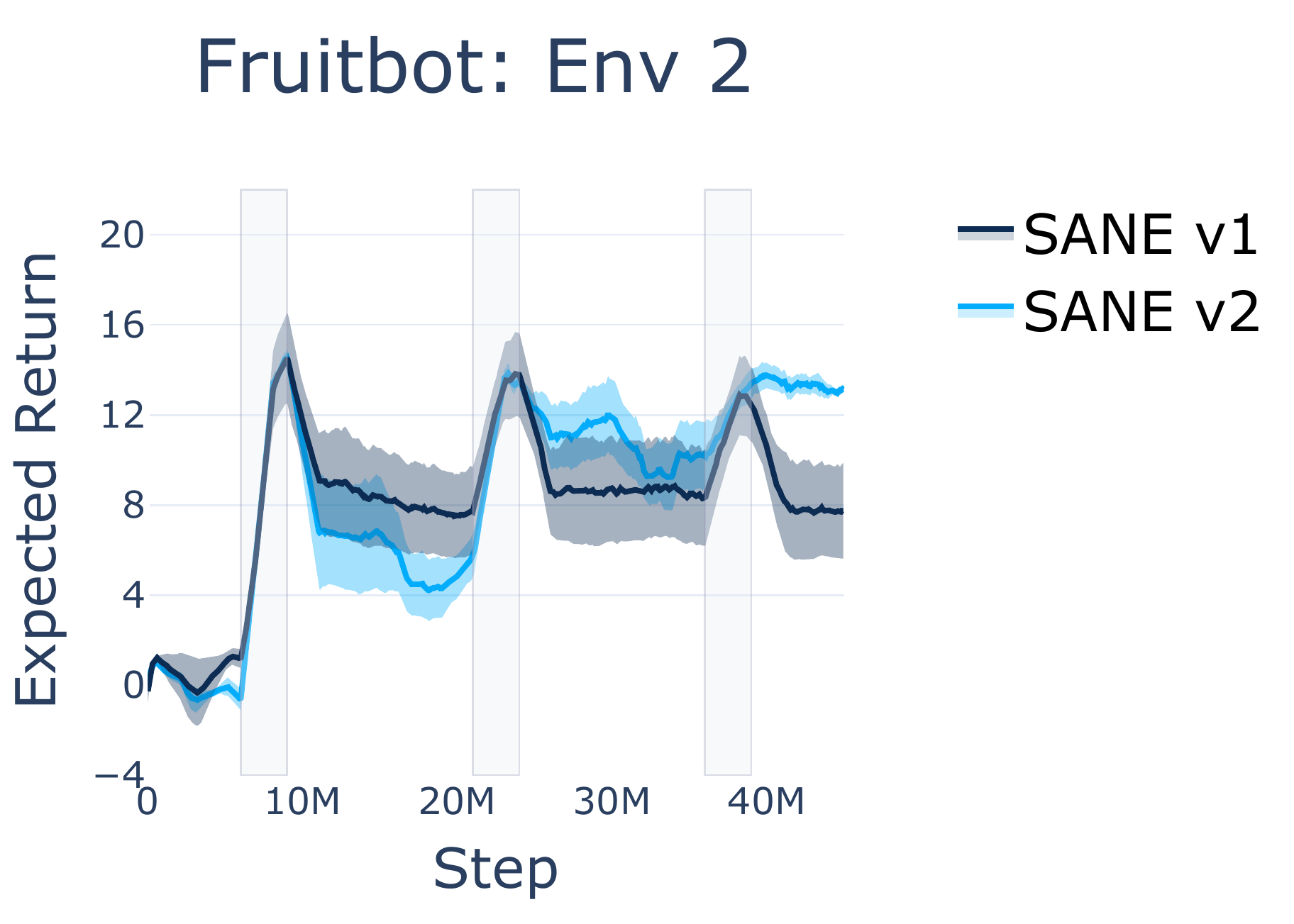}
    \includegraphics[trim=0 0em 20em 0, clip,width=0.15 \columnwidth]{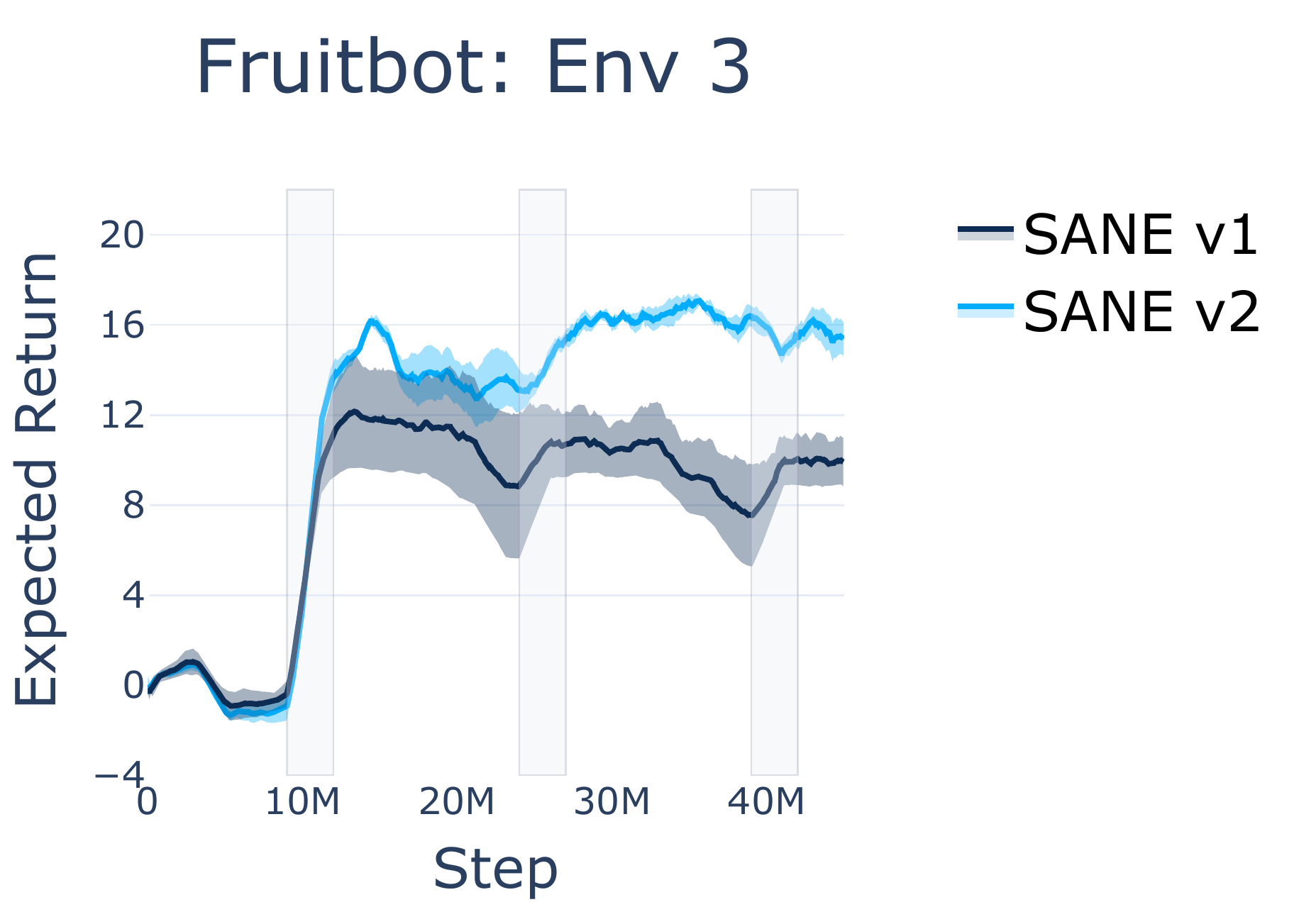}
    \includegraphics[width=0.24 \columnwidth]{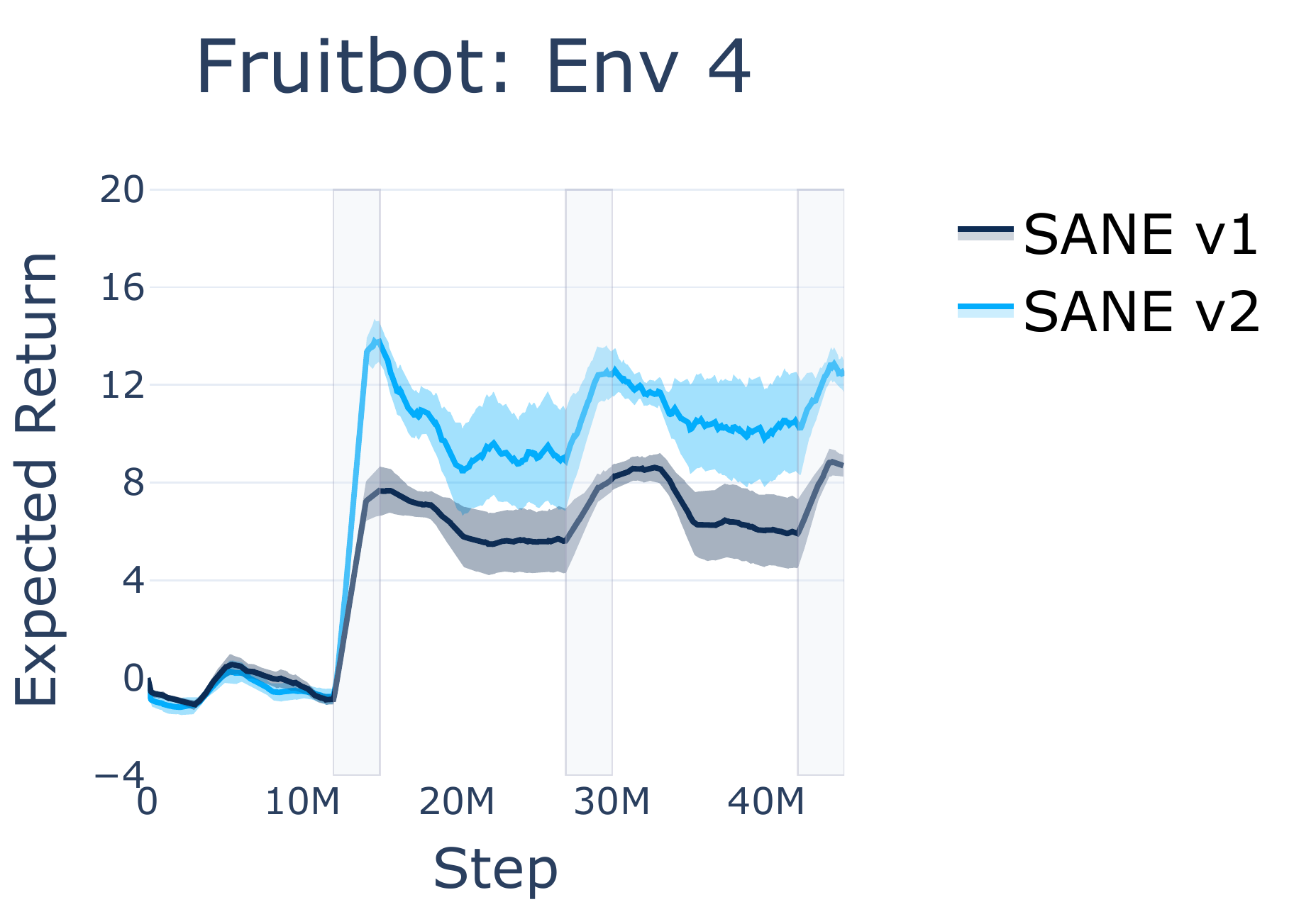}
    \caption{Comparison of SANE variations on the Fruitbot task sequence.}
    \label{fig:sane_ablations_fruitbot}
\end{figure}

Conceptually, the difference between SANE v1 and SANE v2 is that the former trains the critic more aggressively (higher coefficients and fewer frames between target network updates), but also has a much more conservative $v_{LCB}$, meaning more drift must be observed before a node is created. We observe that the former performed well on Climber and Miner, but on Fruitbot we obtained higher performance using the smoother critic training of SANE v2, as shown in Figure \ref{fig:sane_ablations_fruitbot}. However, it is worth noting that the ensemble learns and remembers in both cases, even when the parameters vary widely (e.g. $\alpha_{l,create}$ differs by a factor of 20.). However, optimal parameter selection remains an area of future work.

\subsection{Lineage plot, Fruitbot}
\label{section:fruitbot_lineage}

\begin{figure}[H]
    \centering
    \includegraphics[width=0.2 \columnwidth]{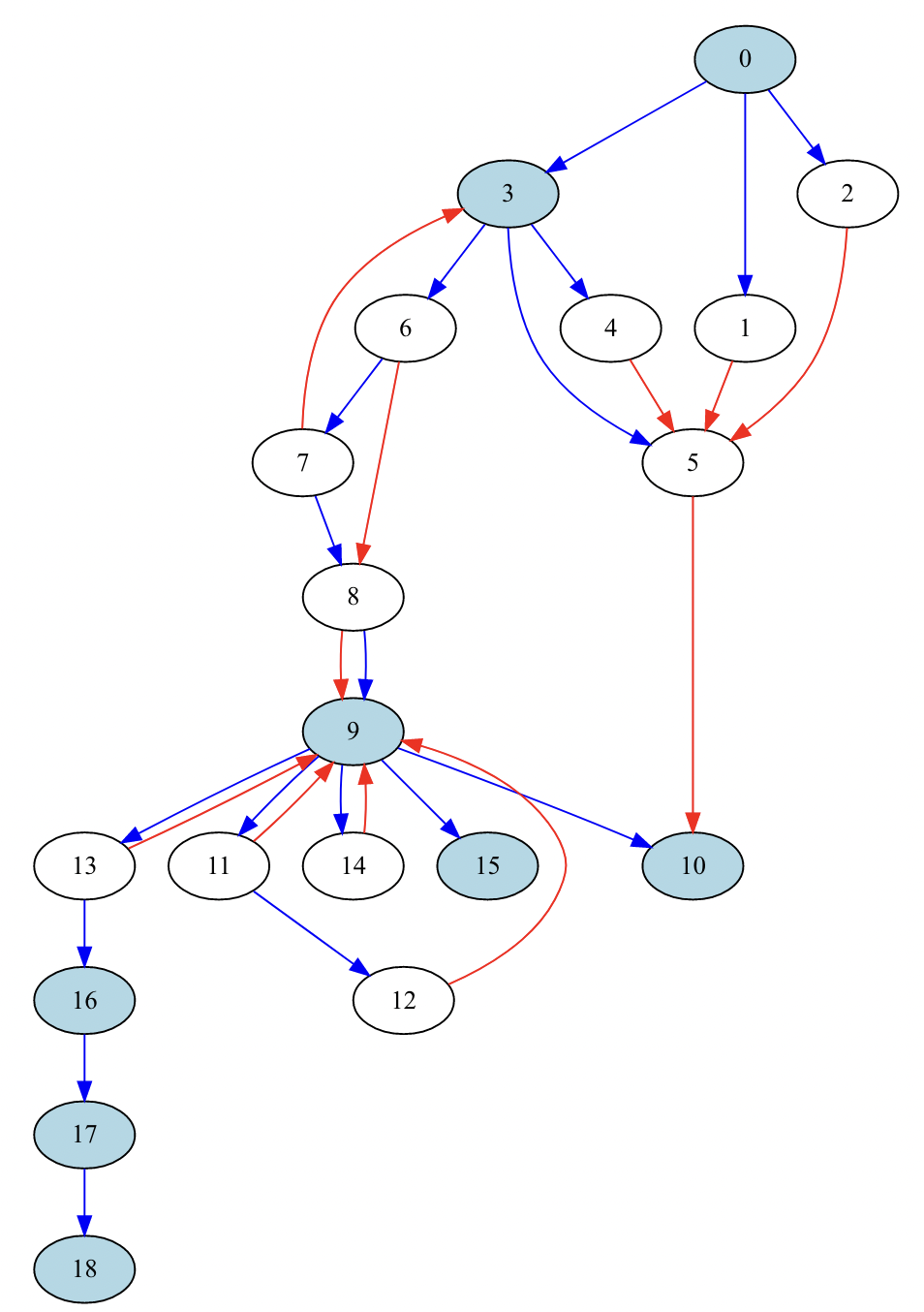}
    \includegraphics[trim=0 0em 0em 0, clip,width=0.3 \columnwidth]{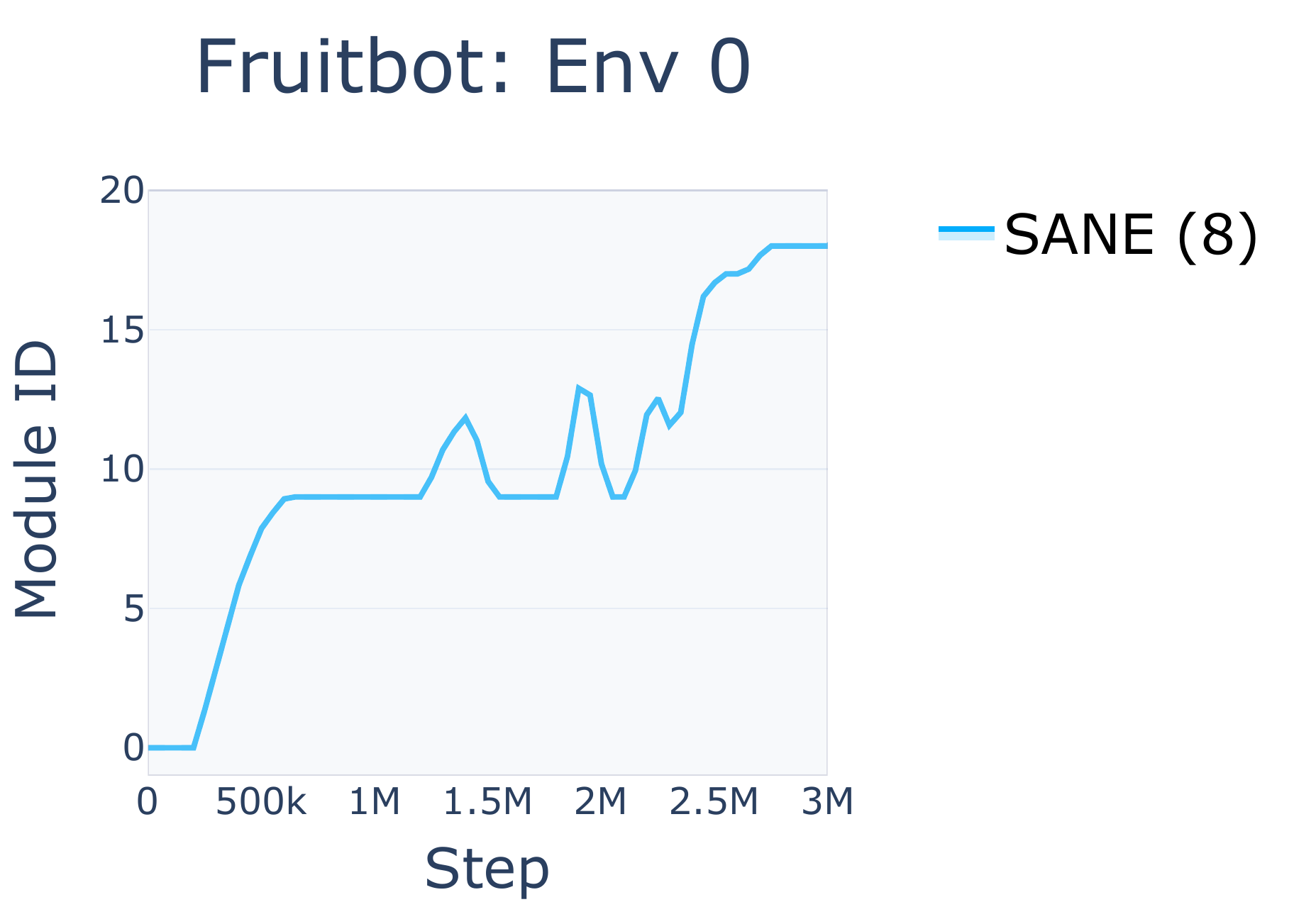}
    \caption{An example Lineage plot, showing how nodes are created and merged while training on Env 0 of Fruitbot.}
    \label{fig:fruitbot_lineage}
\end{figure}

While we are not able to provide the full Lineage plots used in the analysis above (a graph with hundreds of nodes displays poorly in pdf form), we show an example of what one looks like for the first task of Fruitbot. Node 0 is used initially, but soon spawns a cascading sequence of Nodes that settles for some time on Node 9. There is some churn as nodes are created but merged back into 9, until another burst of creation settles finally on Node 18, though we see in Figure \ref{fig:fruitbot_single} that this does not last either. 

If the policies and critics improved monotonically, we would not see such a chaotic plot, but instead one more like what we see for Climber, in Figure \ref{fig:climber_single}. Essentially, since creation only happens when we detect that a node is worse than its anchor, this pattern of creation represents much noisier learning.




\subsection{Algorithm pseudocode for SANE}
\label{sec:pseudocode}


\begin{algorithm}
\caption{SANE}\label{alg:sane}
\begin{algorithmic}[1]
\Require timestep $t$, total task timesteps $T$, state at episode start $s_0$, max allowed module count $N$, modules $\mathcal{M} = \{M_1, \ldots, M_k\}$ where $M_i$ contains actor $\pi_i$, critic $V_i$, static anchor critic $a_i$, and replay buffer $R_i$.

\While{$t < T$} 

     \State $M_{max} = \argmax_i(v_{UCB, i}(s_0))$ \Comment{select active module}

    \State{Collect $t_{new}$ timesteps of data using $M_{max}$}
    
    \State{$t := t + t_{new}$}

\\
\If{$v_{UCB, M_{max}}(s) < v_{a}(s)$} \Comment{negative drift detected, so add module} \\

     \State{$M_{new} := clone(M_{max})$}
     \State{$a_{new} := clone(V_{max})$} 
     \State{$\mathcal{M} := \{M1,...,M_k,M_{new}\}$}
     
\Else

\If{$v_{LCB, m}(s) > v_{a}(s)$} \Comment{positive drift detected, so update module}

 \State{$ a_{max} :=  clone(V_{max})$}

\EndIf
\EndIf

\If{$|M| > N$} \Comment{merge modules}

\State{$g_i := avg(batch(R_i)['observation'])$} \Comment{compute an avg observation for each module}

\State{$M_i$, $M_j$ = $\argmin_{i,j} ||g_i, g_j||_2$} 

\State{$M_{keep}$, $M_{remove}$ = which of $M_i$ or $M_j$ has been used more and fewer times, respectively}

\State{$R_{keep}$ = \{$R_i$, $R_j$\}}

\State{$\mathcal{M} = \mathcal{M} - M_{remove}$}

\State{}

\EndIf

\EndWhile
\end{algorithmic}
\end{algorithm}


\subsection{Final Average Performance Tables}

\begin{table}[H]
    \centering
    \begin{tabular}{c|c|c|c|c|c|c|c}
        \toprule
        Task & SANE & Static SANE & CLEAR 8x & CLEAR & EWC 8x & PnC 8x & SANE Oracle\\
        \hline
        Env 0 & $\bm{9.72 \pm 0.29}$    & $2.54 \pm 1.85$ & $3.02 \pm 0.14$ & $1.26 \pm 0.08$ & $7.46 \pm 0.88$ & $0.24 \pm 0.06$ &                       $9.52 \pm 0.14$ \\
        Env 1 & $10.43 \pm 0.04$   & $9.80 \pm 0.49$ & $2.34 \pm 0.32$ & $1.96 \pm 0.21$ & $\bm{14.37 \pm 1.09}$ & $0.22 \pm 0.05$ &                $10.02 \pm 0.22$ \\
        Env 2 & $7.82 \pm 1.92$         & $6.95 \pm 1.77$ & $1.08 \pm 0.05$ & $1.57 \pm 0.43$ & $\bm{9.33 \pm 0.91}$ & $3.70 \pm 0.95$ &               $10.66 \pm 0.08$ \\
        Env 3 & $1.99 \pm 0.28$         & $2.22 \pm 0.34$ & $7.46 \pm 0.20$ & $8.02 \pm 0.26$ & $\bm{8.40 \pm 0.42}$ & $0.34 \pm 0.05$  &                   $2.71 \pm 0.59$ \\
        \bottomrule
    \end{tabular}
    \caption{Climber final average performance. The Oracle is not considered for the purposes of highlighting the average performance, as it is an idealized model. }
    \label{tab:climber_final}
\end{table}

\begin{table}[H]
    \centering
    \tablemarginhack
    \begin{tabular}{c|c|c|c|c|c|c|c}
        \toprule
        Task & SANE & Static SANE & CLEAR 8x & CLEAR & EWC 8x & PnC 8x & SANE Oracle\\
        \hline
        Env 0 & $\bm{10.25 \pm 0.35}$ & $6.61 \pm 2.18$ & $2.89 \pm 0.24$ & $2.35 \pm 0.12$ & $2.71 \pm 0.26$ & $2.03 \pm 0.55$ & $8.01 \pm 1.41$ \\ 
        Env 1 & $\bm{12.43 \pm 0.03}$ & $8.25 \pm 2.07$ & $9.24 \pm 0.71$ & $4.90 \pm 0.58$ & $11.06 \pm 0.68$ & $3.46 \pm 0.66$ &  $12.56 \pm 0.06$ \\ 
        Env 2 & $9.64 \pm 2.09$ & $9.46 \pm 2.11$ & $4.93 \pm 0.42$ & $3.90 \pm 0.39$ & $\bm{10.65 \pm 0.62}$ & $2.11 \pm 0.54$ &  $11.49 \pm 0.22$\\ 
        Env 3 & $2.97 \pm 0.05$ & $3.06 \pm 0.05$ & $7.06 \pm 2.31$ & $10.74 \pm 1.87$ & $\bm{12.96 \pm 0.02}$ & $1.93 \pm 0.26$ &  $3.15 \pm 0.06$\\
        \bottomrule
    \end{tabular}
    \caption{Miner final average performance. The Oracle is not considered for the purposes of highlighting the average performance, as it is an idealized model.}
    \label{tab:miner_final}
\end{table}

\begin{table}[H]
    \centering
    \tablemarginhack
    \begin{tabular}{c|c|c|c|c|c|c|c}
        \toprule
        Task & SANE & Static SANE & CLEAR 8x & CLEAR & EWC 8x & PnC 8x & SANE Oracle\\
        \hline
        Env 0 & $2.18 \pm 3.46$         & $-0.70 \pm 0.37$  & $\bm{3.77 \pm 0.50}$  & $-2.36 \pm 0.24$ & $2.05 \pm 1.77$ & $-2.09 \pm 0.74$ & $ 11.84 \pm 0.68$ \\ 
        Env 1 & $\bm{6.71 \pm 2.43}$    & $2.45 \pm 0.40$   & $2.85 \pm 0.45$       & $1.26 \pm 0.44$ & $-0.91 \pm 0.56$ & $-2.08 \pm 0.78$ & $11.85 \pm 0.47$\\ 
        Env 2 & $\bm{13.13 \pm 0.21}$   & $10.34 \pm 3.01$  & $9.50 \pm 0.52$       & $5.87 \pm 0.26$ & $10.87 \pm 2.73$ & $4.89 \pm 1.68$ & $15.39 \pm 1.71$\\ 
        Env 3 & $\bm{15.38 \pm 0.84}$   & $6.80 \pm 0.43$   & $9.52 \pm 0.66$       & $6.73 \pm 1.36$ & $9.74 \pm 2.51$ & $1.16 \pm 0.65$ & $16.25 \pm 0.88$\\ 
        Env 4 & $12.36 \pm 0.72$ & $19.48 \pm 0.95$ & $\bm{23.86 \pm 0.47}$         & $19.84 \pm 3.06$ & $6.07 \pm 0.94$ & $1.37 \pm 0.37$ & $12.66 \pm 0.75$\\ 
        \bottomrule
    \end{tabular}
    \caption{Fruitbot final average performance. The Oracle is not considered for the purposes of highlighting the average performance, as it is an idealized model.}
    \label{tab:fruitbot_final}
\end{table}

\subsection{2 Module Experiments}
Here we present training SANE using only 2 modules (each with 2e5 replay buffer entries) on Climber. The intent of this experiment is to look at how having fewer modules than tasks impacts performance. We can see that while peak performance is generally higher, recall is poor overall. Perhaps unsurprisingly, the curve resembles that of CLEAR: reaching similar maximum values, and observing the same forgetting. 

\begin{figure}[H]
    \centering
    \includegraphics[trim=0 0em 22em 0, clip,width=0.2 \columnwidth]{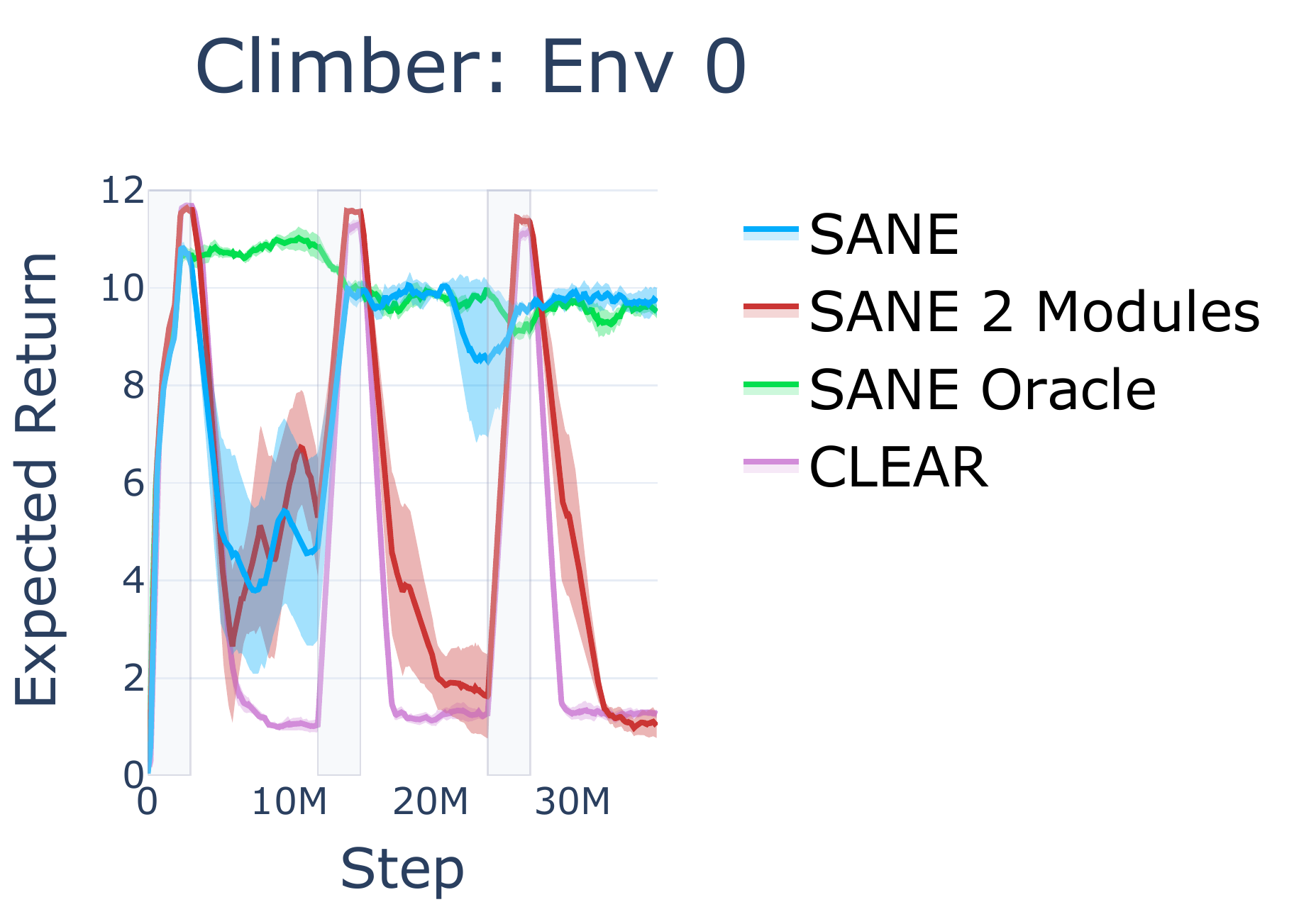}
    \includegraphics[trim=0 0em 22em 0, clip,width=0.2 \columnwidth]{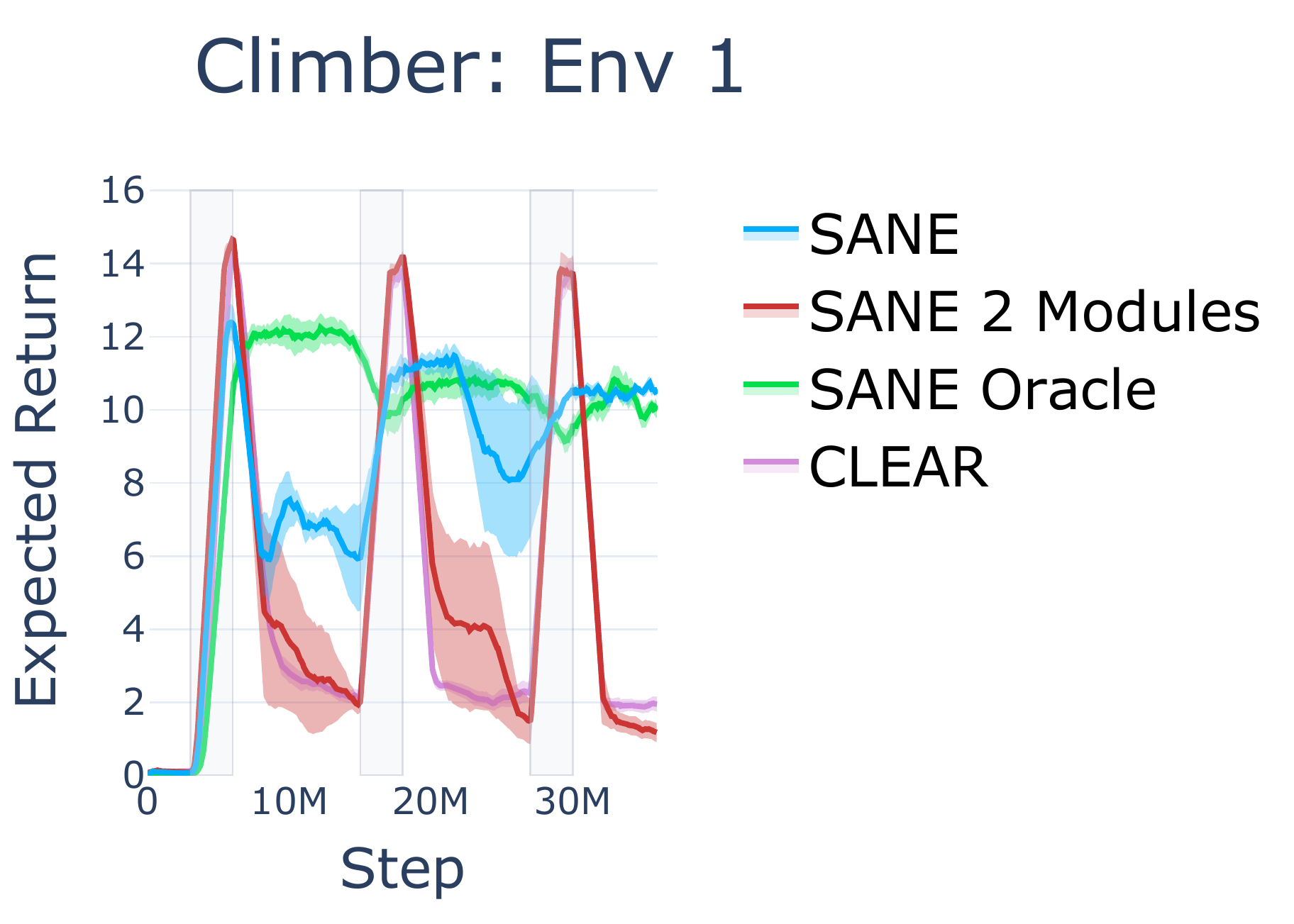}
    \includegraphics[trim=0 0em 22em 0, clip,width=0.2 \columnwidth]{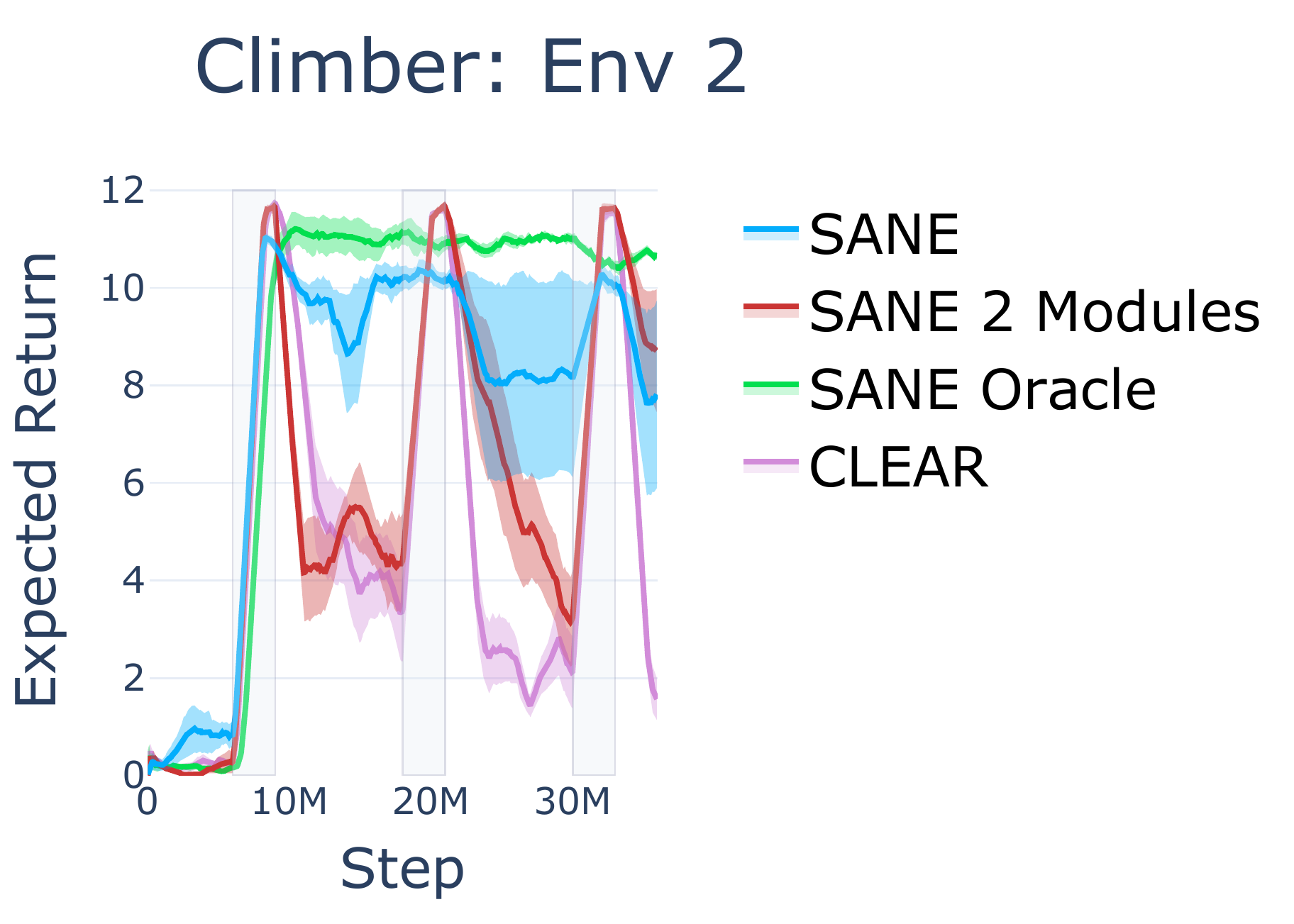}
    \includegraphics[width=0.345 \columnwidth]{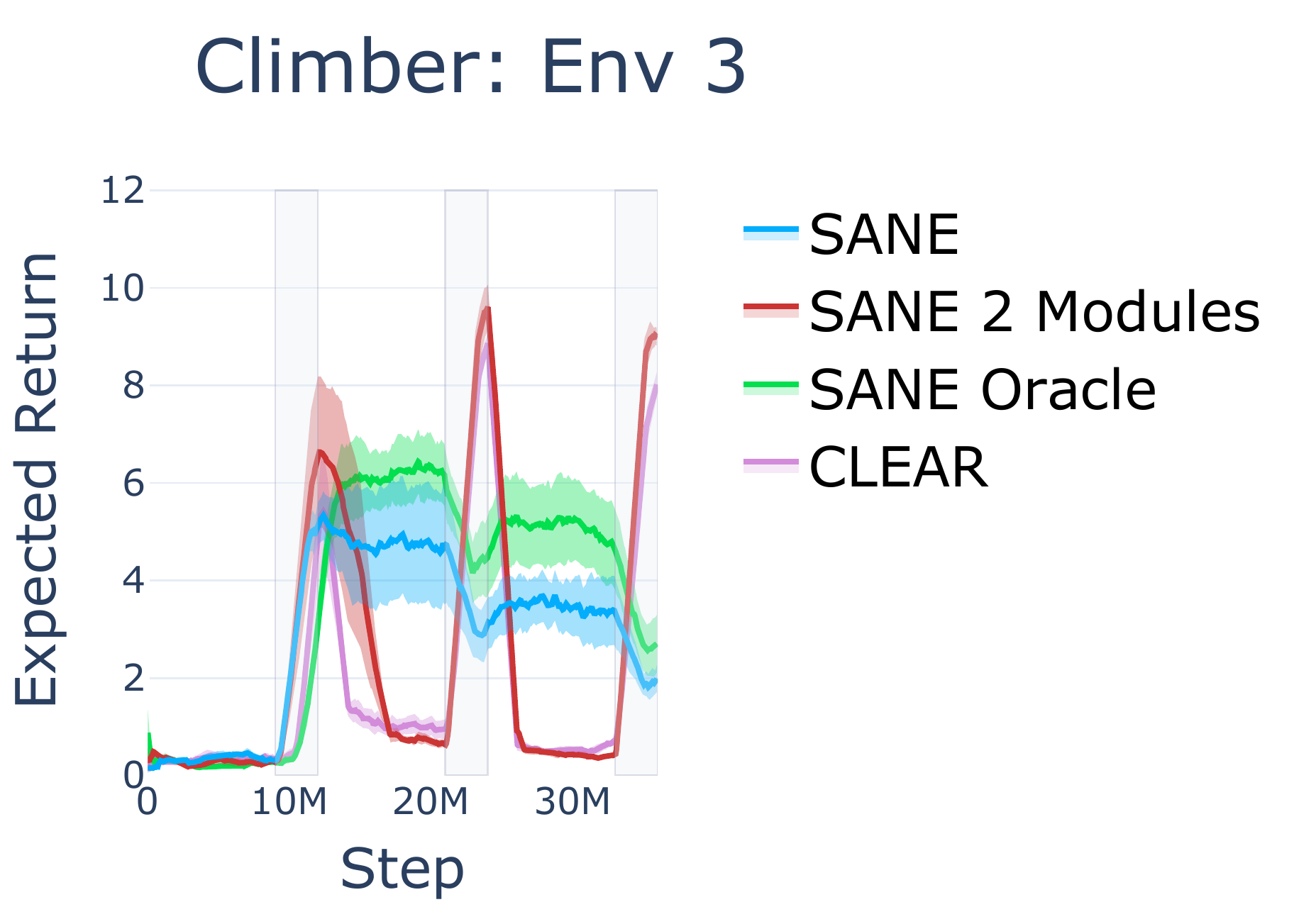}
    \caption{Comparison for using SANE with 2 modules to other SANE variants and CLEAR.}
    \label{fig:climber_results_2mod}
\end{figure}

\subsection{Extended Background}
\label{appendix:extended_background}
In the standard, single-task reinforcement learning scenario, we consider the task $\mathcal{T}$ as a discrete-time Markov Decision Process (MDP) consisting of a tuple $\brackets{S, A, T, r, \gamma, \rho_0,}$, with state space $S$, action space $A$, state transition probability function $T$, reward function $r$, discount factor $\gamma$, and probability distribution $\rho_0$ on the initial states $S_0 \subset S$. The goal is to learn a policy $\pi(a \vert s)$ which maximizes the expected return, where the return $R_t$ of a state $s$ at timestep $t$ is the discounted sum of rewards over an infinite-horizon trajectory from state $s$. 

For continual RL~\cite{kirkpatrick2017ewc, schwarz2018progress, rolnick2019experience}, we extend this setting by considering a sequence of $N$ tasks, $\mathcal{S}_N:=(\mathcal{T}_1 \ldots \mathcal{T}_N)$, presented to the agent. This induces a non-stationary learning process as any component of the MDP may change on task switch. A capable continual RL agent should continue to learn new skills (maintain plasticity), recall prior learned behavior (mitigate catastrophic forgetting), and transfer old abilities to new domains (demonstrate forward transfer). We assume the agent trains on task $\mathcal{T}_i$ at timesteps in the interval $[A_i, B_i)$, for $k_i = B_i - A_i$ timesteps. Here $A_i$ and $B_i$ are the task boundaries denoting the start and end, respectively, of task $\mathcal{T}_i$. Additionally, we may cycle through the tasks $M$ times, which yields full task sequence $\mathcal{S}_{NM}$ that has length $N\cdot M$.

\subsection{Analysis of Estimated Returns}

\begin{figure}[H]
    \centering
    \includegraphics[width=0.9 \columnwidth]{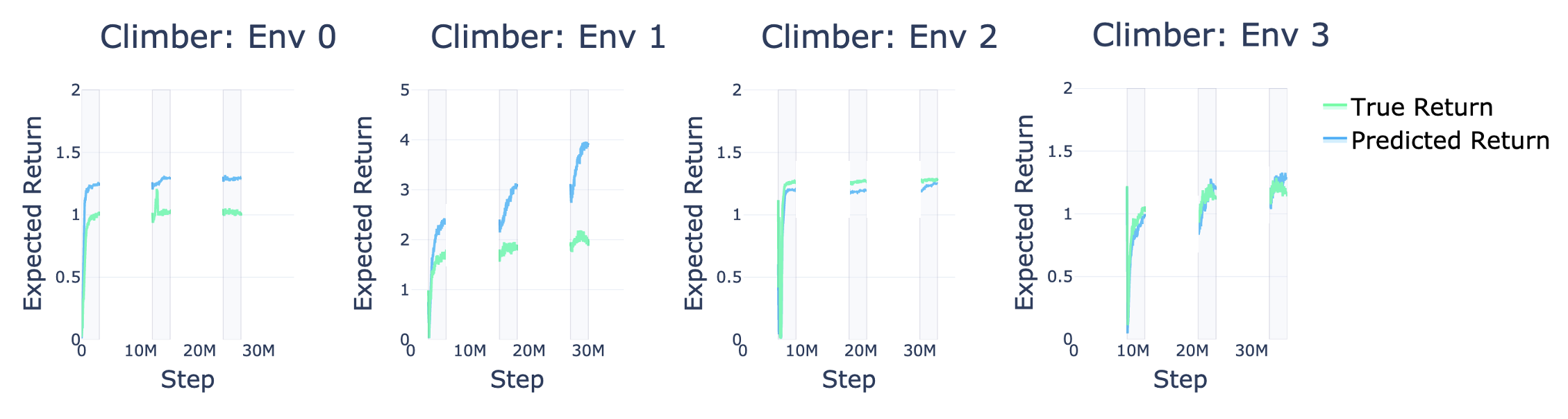}
    \caption{Comparison of the true and predicted vtrace returns.}
    \label{fig:climber_results_vtrace}
\end{figure}

In Figure~\ref{fig:climber_results_vtrace} we present SANE's estimated vtrace return in comparison to the observed vtrace return for the training periods for each task for a single run. There is a bit of overestimation in Env 0, more significant overestimation in Env 1, but almost none in Envs 2 and 3. Env 1 is also where we see the most uncertainty. Note that the first time a transition to a new task occurs, the estimated return is near zero.

We believe these estimates to be close enough that using our estimated vtrace returns is preferred over using a history of recent returns. An example of when this is the case is if every episode is different; then any window size to average over poses problems: the best module will not be activated and creation will not occur properly. Furthermore, we hope to extend SANE such that activation occurs many times during an episode as well, so modules can capture even more fine-grained behaviors. This fine-grained behavior becomes impossible if we are reliant on final returns.


\begin{figure}[H]
    \centering
    \includegraphics[width=0.9 \columnwidth]{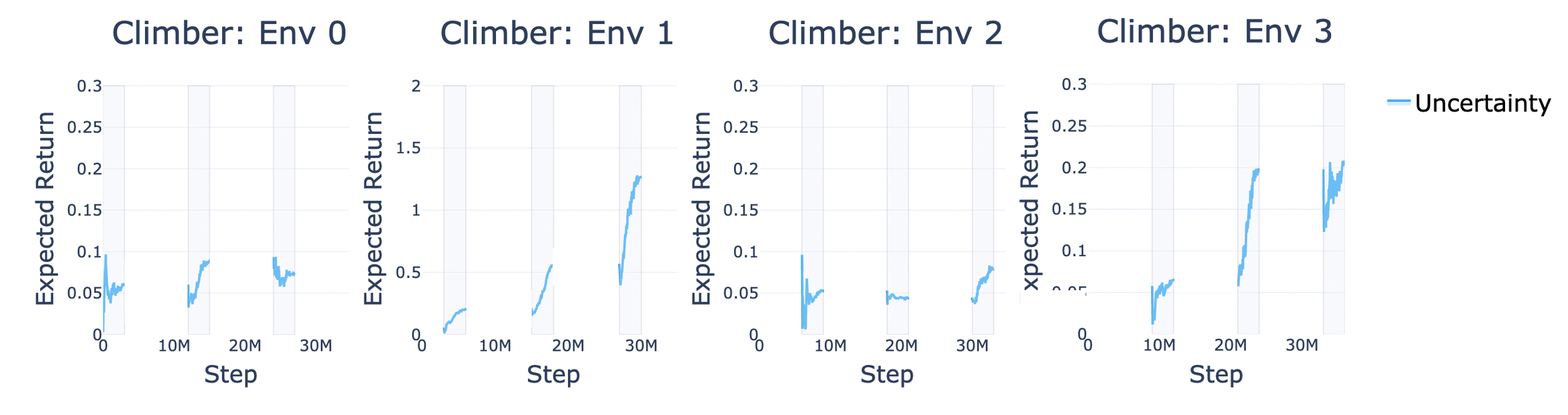}
    \caption{Graph of the uncertainty of SANE's prediction of our vtrace returns.}
    \label{fig:climber_results_vtrace_unc}
\end{figure}


\end{document}